%% file: colt2018.tex
\documentclass[final]{colt2018} 

\title[Exponential Convergence of Testing Error for Stochastic Gradient Methods]{Exponential Convergence of Testing Error\\
 for Stochastic Gradient Methods}
\usepackage{times}
\usepackage{color}
\usepackage{stmaryrd}
\usepackage{enumitem} 
\usepackage{hyperref}
\usepackage{graphicx}
\usepackage{stmaryrd}
\usepackage{stackengine}
\SetSymbolFont{stmry}{bold}{U}{stmry}{m}{n}
\usepackage[mathcal]{eucal}

  \coltauthor{\Name{Loucas Pillaud-Vivien} \Email{loucas.pillaud-vivien@inria.fr}\\
   \Name{Alessandro Rudi} \Email{alessandro.rudi@inria.fr}\\
   \Name{Francis Bach} \Email{francis.bach@inria.fr}\\
   \addr INRIA - D\'epartement d'Informatique de l'ENS \\
  Ecole Normale Sup\'erieure, CNRS, INRIA, PSL Research University
  }

\include{additional_defs_colt}

\begin{document}

\maketitle

\begin{abstract}
We consider binary classification problems with positive definite kernels and square loss, and study the convergence rates of stochastic gradient methods. We show that while the excess testing \emph{loss} (squared loss) converges slowly to zero as the number of observations (and thus iterations) goes to infinity, the testing \emph{error} (classification error) converges exponentially fast if low-noise conditions are assumed. To achieve these rates of convergence we show sharper high-probability bounds with respect to the number of observations for stochastic gradient descent.
\end{abstract}


\section{Introduction}

Stochastic gradient methods are now ubiquitous in machine learning, both from the practical side, as a simple algorithm that can learn from a single or a few passes over the data~\citep{bottou2005line}, and from the theoretical side, as it leads to optimal rates for estimation problems in a variety of situations~\citep{nemirovsky1983problem,polyak1992acceleration}.

They follow a simple principle~\citep{robbins:monro:1951}: to find a minimizer of a function~$F$ defined on a vector space from noisy gradients, simply follow the negative stochastic gradient and the algorithm will converge to a stationary point, local minimum or global minimum of $F$ (depending on the properties of the function $F$), with a rate of convergence that decays with the number of gradient steps  $n$ typically as  $O(1/\sqrt{n})$, or $O(1/n)$ depending on the assumptions which are made on the problem \citep[see, e.g.,][]{polyak1992acceleration,nesterov2008confidence,nemirovski2009robust,shalev2007pegasos,xiao2010dual,gradsto,newsto,daft}.

 On the one hand, these rates are optimal for the estimation of the minimizer of a function given access to noisy gradients~\citep{nemirovsky1983problem}, which is essentially the usual machine learning set-up where the function~$F$ is the expected \emph{loss}, e.g., logistic or hinge for classification, or least-squares for regression, and the noisy gradients are obtained from sampling a single pair of observations.

On the other hand, although these rates as $O(1/\sqrt{n})$ or $O(1/n)$ are optimal, there are a variety of extra assumptions that allow for faster rates, even exponential rates.

 First, for stochastic gradient from a finite pool, that is for $F= \frac{1}{k} \sum_{i=1}^k F_i$,   a sequence of  works starting from SAG~\citep{roux2012stochastic}, SVRG~\citep{johnson2013accelerating}, SAGA~\citep{defazio2014saga}, have shown explicit exponential convergence. However, these results, once applied to machine learning where the function $F_i$ is the loss function associated with the $i$-th observation of a finite training data set of size $k$, say nothing about the loss on unseen data (test loss). The rates we present in this paper are on \emph{unseen} data.

 Second, assuming that at the optimum all stochastic gradients are equal to zero, then for strongly-convex problems (e.g., linear predictions with low-correlated features), linear convergence rates can be obtained for test losses~\citep{solodov1998incremental,schmidt2013fast}. However, for supervised machine learning, this has limited relevance as having zero gradients for all stochastic gradients at the optimum essentially implies prediction problems with no uncertainty (that is, the output is a deterministic function of the input). Moreover, we can only get an exponential rate for strongly-convex problems and thus this imposes a parametric noiseless problem, which limits the applicability (even if the problem was noiseless, this can only reasonably be in a non-parametric way with neural networks or positive definite kernels). Our rates are   on noisy problems and on infinite-dimensional problems where we can hope that we approach the optimal prediction function with large numbers of observations. For prediction functions described by a reproducing kernel Hilbert space, and for the square loss, the excess testing loss  (equal to testing loss minus the minimal testing loss over all measurable prediction functions) is known to converge to zero at a subexponential rate typically greater than $O(1/n)$~\citep{dieuleveut2016nonparametric,daft}, these rates being optimal for the estimation of testing losses.

 Going back to the origins of supervised machine learning with binary labels, we will not consider getting to the optimal  testing \emph{loss} (using a  convex surrogate such as logistic, hinge or least-squares) but the testing \emph{error} (number of mistakes in predictions), also referred to as the 0-1 loss.

 It is known that the excess testing error (testing error minus the minimal testing error over all measurable prediction functions) is upper bounded by a function of the excess testing loss~\citep{zhang2004statistical,bartlett2006convexity}, but always with a loss in the convergence rate (e.g., no difference or taking square roots). Thus a slow rate in $O(1/n)$ or $O(1/\sqrt{n})$ on the excess loss leads to a slow(er) rate on the excess testing error.

 Such general relationships between excess loss and excess error have been refined with the use of \emph{margin conditions}, which characterize  how hard the prediction problems are~\citep[see, e.g.,][]{mammen1999smooth}. Simplest input points are points where the label is deterministic (i.e., conditional probabilities of the label are equal to zero or one), while hardest points are the ones where the conditional probabilities are equal to $1/2$. Margin conditions quantify the mass of input points which are hardest to predict, and lead to improved transfer functions from testing losses to testing errors, but still no exponential convergence rates~\citep{bartlett2006convexity}.
 
In this paper, we consider the strongest margin condition, that is conditional probabilities are bounded away from $1/2$, but not necessarily equal to $0$ or $1$. This assumption on the learning problem has been used in the past to show that regularized empirical (convex) risk minimization leads to exponential convergence rates~\citep{audibert2007fast,koltchinskii2005exponential}. Our main contribution is to show that stochastic gradient descent also achieves similar rates (see an empirical illustration in Figure \ref{fig:plots} in the Appendix \ref{ap:experiments}). This requires several side contributions that are interesting on their own, that is, a new and simple formalization of the learning problem that allows exponential rates of estimation (regardless of the algorithms used to find the estimator) and a new  concentration result for averaged stochastic gradient descent (SGD) applied to least-squares, which is finer than existing work~\citep{newsto}.
 
The paper is organized as follows: in \mysec{setup}, we present the learning set-up, namely binary classification with positive definite kernels, with a particular focus on the relationship between errors and losses. Our main results rely on a generic condition for which we give concrete examples in \mysec{examples}. In \mysec{sgd}, we present our version of stochastic gradient descent, with the use of tail averaging~\citep{jain2016parallelizing}, and provide new  deviation inequalities, which we apply in \mysec{expo} to our learning problem, leading to exponential convergence rates for the testing errors. We conclude in \mysec{conc} by providing several avenues for future work. Finally, synthetic experiments illustrating our results can be found in Section \ref{ap:experiments} of the Appendix.

\paragraph{Main contributions of the paper.} We would like to underline that our main contributions are in the two following results; (a) we show in Theorem \ref{th:errortail} the exponential convergence of stochastic gradient descent on the testing error, and (b) this result strongly rests on a new deviation inequality stated in Corollary \ref{co:SGDtailaveraged} for stochastic gradient descent for least-square problems. This last result is interesting on its own and gives an improved high-probability result which does not depend on the dimension of the problem and has a tighter dependence on the strongly convex parameter --through the effective dimension of the problem, see \citet{caponnetto2007optimal,dieuleveut2016nonparametric}.

\section{Problem Set-up}
\label{sec:setup}

In this section, we present the general machine learning set-up, from generic assumptions to more specific assumptions.

\subsection{Generic assumptions}

We consider a measurable set $\X$  and a probability distribution~$\rho$ on data $(x,y) \in \X \times \{-1,1\}$; we denote by $\rhox$ the marginal probability on $x$, and by $\rho(\pm 1|x)$ the conditional probability that $y=\pm1$ given $x$. We have $\E(y | x) = \rho(1|x) - \rho(-1|x)$. Our main margin condition is the following (and independent of the learning framework):

\bas\label{asm:separability} 
$ | \E (y | x) | \geqslant \delta$ almost surely for some $\delta \in (0,1]$.
\eas

This margin condition (often referred to as a low-noise condition) is commonly used in the theoretical study of binary classification~\citep{mammen1999smooth,audibert2007fast,koltchinskii2005exponential}, and usually takes the following form:  
$\forall \delta>0,\ \P(  |\E(y|x) | < \delta ) = O( \delta^\alpha)$ for $\alpha >0$. Here, however, $\delta$ is a fixed constant. Our stronger margin condition \asm{asm:separability} is necessary to show exponential convergence rates but we give also explicit rates in the case of the latter low-noise condition. This extension is derived in Appendix~\ref{ap:weakmargin} and more precisely in Corollary~\ref{co:weakmargin}. 
Note that the smaller the $\alpha$, the larger the mass of inputs with hard-to-predict labels. Our condition corresponds to $\alpha = +\infty$, and simply states that for all inputs, the problem is never totally ambiguous, and the degree of non-ambiguity is bounded from below by $\delta$. When $\delta = 1$, then the label $y \in \{-1,1\}$ is a deterministic function of $x$, but our results apply for all $\delta \in (0,1]$ and thus to noisy problems (with low noise). Note that problems like image classification or object recognition are well characterized by \asm{asm:separability}. Indeed, the noise in classifying an image between two disparate classes (cars/pedestrians, bikes/airplanes) is usually way smaller that $1/2$.

We will consider learning functions in a reproducing kernel Hilbert space (RKHS) $\H$ with kernel function $K: \X \times \X \to \rb$ and dot-product $\langle \cdot , \cdot \rangle_\H$. We make the following standard assumptions on~$\H$: 

\bas\label{asm:kernel-bounded}  
$\H$ is a separable Hilbert space and there exists $R > 0$, such that  for all $x \in \X$, $K(x,x) \leqslant R^2$.
\eas
For $x \in \X$, we consider the function $K_x: \X \to \rb$ defined as $K_x(x') = K(x,x')$. We have the classical reproducing property for $g \in \H$, $g(x) = \langle g, K_x \rangle_\H$~\citep{Cristianini2004,smola-book}.
We will consider other norms, beyond the RKHS norm $\| g \|_\H$, that is the $L_2$-norm (always with respect to $\rhox$), defined as 
$\| g\|_{L_2}^2 = \int_{\X} g(x)^2 d\rhox(x)$, as well as the $L_\infty$-norm  $\| \cdot \|_{L_\infty}$ on the support of $\rhox$. A key property is that \asm{asm:kernel-bounded} implies $\| g\|_{L_\infty} \leqslant R \| g\|_\H$. 


Finally, we will consider observations with standard assumptions:

\bas\label{asm:data-iid}
The observations $(x_n,y_n) \in \X \times \{-1,1\}, n \in \N^*$ are independent and identically distributed with respect to the distribution $\rho$.
\eas

\subsection{Ridge regression}
In this paper, we focus primarily on least-squares estimation to obtain estimators. We define $g_\ast$ as the minimizer over $L_2$ of 
$$ \E  ( y - g(x) )^2 = \int_{\X \times \{-1,1\}} ( y - g(x) )^2 d\rho(x,y) .$$
We always have $g_\ast(x) = \E(y|x) = \rho(1|x) - \rho(-1|x)$, but we
\emph{do not require} $g_\ast \in \H$. We also consider the ridge regression problem \citep{caponnetto2007optimal} and denote by $g_\lambda$  the unique (when $\lambda > 0$) minimizer in $\H$ of 
$$ \E  ( y - g(x) )^2  + \lambda \| g\|_\H^2  .$$
The function $g_\lambda$ always exists for $\lambda >0$ and is always an element of $\H$. 
%
%
When $\H$ is dense in $L_2$ our results depend on the $L_\infty$-error $\|g_\lambda - g_\ast\|_\infty$, which is weaker than $\|g_\lambda - g_\ast\|_\H$ which itself only exists when $g_\ast \in \H$ (which we do not assume).
When $\H$ is not dense we simply define $\Phg$ as the orthonormal projector for the $L_2$ norm on $\H$ of $g_* = \E(y|x)$ so that our bound will the depend on $\|g_\lambda - \Phg\|_\infty$. Note that $\Phg$ is the minimizer of $\E (y - g(x))^2$ with respect to $g$ in the closure of $\H$ in~$L_2$.

Moreover our  main technical  assumption is: 
\bas\label{asm:flambda-correct-sign} 
There exists $\lambda>0$ such that almost surely, 
$ \displaystyle \sign( \E(y|x) ) g_\lambda(x) \geqslant \frac{\delta}{2}$. 
\eas

In the assumption above, we could replace $\delta/2$ by any multiplicative constants in $(0,1)$ times$\delta$ (instead of $1/2$). Note that with \asm{asm:flambda-correct-sign}, $\lambda$ depends on $\delta$ and on the probability measure $\rho$, which are both fixed (respectively by \asm{asm:separability} and the problem), so that $\lambda$ is fixed too. It implies that for any estimator $\hat{g}$ such that 
$\| g_\lambda - \hat{g}\|_{L_\infty} < \delta / 2
$, the predictions from~$\hat{g}$ (obtained by taking the sign of $\hat{g}(x)$ for any $x$), are the same as the sign of the optimal prediction $\sign(\E(y|x))$. Note that a  sufficient condition is $\| g_\lambda - \hat{g}\|_{\H} < \delta/(2R)$ (which does not assume that $g_\ast \in \H$), see next subsection.

Note that more generally, for all problems for which \asm{asm:separability} is true and ridge regression (in the population case) is so that
$
\| g_\lambda - g_\ast\|_{L_\infty}
$ tends to zero as $\lambda$ tends to zero then \asm{asm:flambda-correct-sign} is satisfied, since 
$\| g_\lambda - g_\ast\|_{L_\infty}  \leqslant \delta/2$ for $\lambda$ small enough, together with \asm{asm:separability} then implies \asm{asm:flambda-correct-sign}.

In  \mysec{examples}, we provide concrete examples where  \asm{asm:flambda-correct-sign} is satisfied and we then present the SGD algorithm and our convergence results. Before we relate excess testing losses to excess testing errors.

\subsection{From testing losses to testing error}
Here we provide some results that will be useful to prove exponential rates for classification with squared loss and stochastic gradient descent. First we define the 0-1 loss defining the classification error:
$$\closs(g) = \rho(\{(x,y) : \sign(g(x)) \neq y \}),$$
where $\sign u = +1$ for $u \geq 0$ and $-1$ for $u < 0$.
In particular denote by $\closs^*$ the so-called {\em Bayes risk} $\closs^* = \closs(\condexp{y}{x})$ which is the minimum achievable classification error \citep{devroye2013probabilistic}.

A well known approach to bound the testing errors by testing losses is \emph{via transfer functions}. In particular we recall the following result \citep{devroye2013probabilistic,bartlett2006convexity}, let $g_\ast(x)$ be equal to $\condexp{y}{x}$ a.e., then
$$
\closs(g) - \closs^* \leq \phi(\|g - g_\ast\|^2_{L^2}), \qquad \forall g \in L^2(d\rhox),
$$ 
with $\phi(u) = \sqrt{u}$ (or $\phi(u) = u^\beta$, with $\beta \in [1/2,1]$, depending on some properties of $\rho$ \citep{bartlett2006convexity}. While this result does not require \asm{asm:separability} or \asm{asm:flambda-correct-sign}, it does not readily lead to exponential rates since the squared loss excess risk has minimax lower bounds that are polynomial in $n$ \citep[see][]{caponnetto2007optimal}.

Here we follow a different approach, requiring via \asm{asm:flambda-correct-sign} the existence of $g_\la$ having the same sign as $g_\ast$ and with  absolute value uniformly bounded from below. Then we can  bound the 0-1 error with respect to the distance in $\hh$ of the estimator $\widehat{g}$ from $g_\la$ as shown in the next lemma (proof in Appendix~\ref{sect:proof-A5-to-01}). This will lead to exponential rates when the distribution satisfies a margin condition  \asm{asm:separability} as we prove in the next section and in Section~\ref{sec:expo}. Note also that for the sake of completeness we recalled in Appendix \ref{sect:exp-rates-for-KRR} that exponential rates could be achieved for kernel ridge regression.

\blm[From approximately correct sign to 0-1 error]\label{lm:appr-correct-sign-to-01}
Let $q \in (0,1)$. Under \asm{asm:separability}, \asm{asm:kernel-bounded}, \asm{asm:flambda-correct-sign}, let $\widehat{g} \in \hh$ be a random function such that
$ \nor{\widehat{g} - g_\la}_\hh < \frac{\delta}{2R}, \mbox{ with probability at least }1-q .$
Then 
$$ \closs(\widehat{g}) = \closs^*, \mbox{ with probability at least }1-q,\ \textrm{and in particular}\  \expect{\closs(\widehat{g}) - \closs^*} \leq q.$$
\elm

In the next section we provide sufficient conditions and explicit settings naturally satisfying~\asm{asm:flambda-correct-sign}.

\section{Concrete Examples and Related Work}
\label{sec:examples}
In this section we illustrate specific settings that naturally satisfy \asm{asm:flambda-correct-sign}. We start by the following simple result showing that the existence of $g_\ast \in \hh$ such that $g_\ast(x) = \condexp{y}{x}$ a.e.~on the support of~$\rhox$, is sufficient to have \asm{asm:flambda-correct-sign} (proof in Appendix~\ref{sect:from-g-in-H-to-A5}). 
\bp\label{prop:gstar-in-hh-gives-gla-good}
Under \asm{asm:separability}, assume that there exists  $g_* \in \hh$
such that $g_*(x) := \condexp{y}{x}$ on the support of $\rhox$, then for any $\delta$, there exists $\la > 0$ satisfying~\asm{asm:flambda-correct-sign}, that is,  $\ \displaystyle \sign( \E(y|x) ) g_\lambda(x) \geqslant \frac{\delta}{2}$.
\ep  
We are going to use the proposition above to derive more specific settings. In particular we consider the case where the positive and negative classes are separated by a margin that is strictly positive. Let $\X \subseteq \R^d$ and denote by $\mathcal{S}$ the support of the probability $\rhox$ and by $\mathcal{S}_+ = \{x \in \X : g_*(x) > 0\}$ the part associated to the positive class, and by $\mathcal{S}_-$ the one associated with the negative class. Consider the following assumption:
\bas\label{asm:margin}
There exists $\mu > 0$ such that $\min_{x \in \mathcal{S}_+, x' \in \mathcal{S}_-} \|x - x'\| \geq \mu$.
\eas
Denote by $W^{s,2}$ the Sobolev space of order $s$ defined with respect to the $L^2$ norm, on $\R^d$ \citep[see][and Appendix~\ref{sect:A5-examples}]{adams2003sobolev}. We also introduce the following assumption:
\bas\label{asm:kernel-rich}
$\X \subseteq \R^d$ and the kernel is such that $W^{s,2} \subseteq \hh$, with $s > d/2$.
\eas
An example of kernel such that $\hh = W^{s,2}$, with $s > d/2$ is the Abel kernel $K(x,x') = e^{-\frac{1}{\sigma}\|x-x'\|}$, for $\sigma > 0$.
In the following proposition we show that if there exist two functions in $\hh$, one matching $\condexp{y}{x}$ on $\mathcal{S}_+$ and the second matching $\condexp{y}{x}$ on $\mathcal{S}_-$ and if the kernel satisfies \asm{asm:kernel-rich}, then \asm{asm:flambda-correct-sign} is satisfied. 
\bp\label{prop:2g-makes-gstar}
Under \asm{asm:separability},~\asm{asm:margin},~\asm{asm:kernel-rich}, if there exist two functions $g_+^*, g_-^* \in W^{s,2}$ such that $g_+^*(x) = \condexp{y}{x}$ on $\mathcal{S}_+$ and $g_-^*(x) = \condexp{y}{x}$ on $\mathcal{S}_-$, then \asm{asm:flambda-correct-sign} is satisfied.
\ep
Finally we are able to introduce another setting where \asm{asm:flambda-correct-sign} is naturally satisfied (the proof of the proposition above and the example below are given in Appendix~\ref{sect:A5-examples}).
\bex[Independent noise on the labels]\label{ex:independent-noise-on-labels}
Let $\rhox$ be a probability distribution on $\X \subseteq \R^d$ and let $\mathcal{S}_+, \mathcal{S}_- \subseteq \X$ be a partition of the support of $\rho_\X$ satisfying $\rhox(S_+), \rhox(S_-) > 0$ and \asm{asm:margin}. Let $n \in \N^*$. For $1 \leq i \leq n$,  $x_i$ independently sampled from $\rhox$ and the label $y_i$ defined by the law 
$$y_i = \begin{cases}
\ \ \zeta_i & \textrm{if} ~ x_i \in S_+\\
-\zeta_i & \textrm{if} ~ x_i \in S_-,
\end{cases}$$ 
with $\zeta_i$ independently distributed as $\zeta_i = -1$ with probability $p \in [0,1/2)$ and $\zeta_i = 1$ with probability $1-p$. Then \asm{asm:separability} is satisfied with $\delta = 1-2p$ and \asm{asm:flambda-correct-sign} is satisfied as soon as \asm{asm:kernel-bounded} and \asm{asm:kernel-rich} are, that is, the kernel is bounded and $\H$ is rich enough (see an example in Appendix \ref{sect:examples-for-glambda} Figure \ref{fig:example-A5}).
\eex

Finally note that the results of this section can be easily generalized from $\X = \R^d$ to any Polish space, by using a {\em separating} kernel \citep{de2014learning,rudi2014learning} instead of \asm{asm:kernel-rich}.

\section{Stochastic Gradient descent}
\label{sec:sgd}

We now consider the stochastic gradient algorithm to solve the ridge regression problem with a fixed  strictly positive regularization parameter $\lambda$. 
We consider solving the regularized problem with regularization $\|  g- g_0 \|_{\H}^2$ through stochastic approximation starting from a function $g_0 \in \H$ (typically~$0$).\footnote{Note that $g_0$ is the initialization of the recursion, and is not the limit of $g_\lambda$ when $\lambda$ tends to zero (this limit being~$\Phg$).} Denote by $F: \hh \to \R$, the functional
$$ F(g)= \E (Y - g(X) )^2 = \E( Y - \langle K_X , g \rangle  )^2,$$
where the last identity is due to the reproducing property of the RKHS $\hh$.
Note that $F$ has the following gradient   $\nabla F(g) = - 2 \E \left[  (Y - \langle K_X , g \rangle)K_X  \right]$. We consider also  $F_\lambda = F + \lambda \| \cdot - g_0\|_\H^2$, for which $\nabla F_\lambda(g) = \nabla F(g) + 2 \lambda (g - g_0)$, and we have for each pair of observation $(x_n,y_n)$ that $F_\lambda (g)  = \E \big[ F_{n,\lambda}(g)\big] = \E (\langle g, K_{x_n}\rangle-y_n)^2 + \lambda \|g-g_0\|_\H^2 $, with
 $F_{n,\lambda}(g) = (\langle g, K_{x_n}\rangle-y_n)^2 + \lambda \|g-g_0\|_\H^2$.
 
 Denoting $\Sigma = \E \big[ K_{x_n} \otimes K_{x_n} \big]$ the covariance operator defined as a linear operator from $\H$ to $\H$ \citep[see][and references therein]{fukumizu2004dimensionality}, we have the optimality conditions for $g_\lambda$ and~$\Phg$:
$$\Sigma g_\lambda - \E \left(y_n K_{x_n}\right) + \lambda ( g_\lambda - g_0) = 0,  \qquad \E \left[ \left(y_n - \Phg(x_n)\right)K_{x_n}\right] = 0,$$
see \citet{caponnetto2007optimal} or Appendix~\ref{ap:optimality} for the proof of the last identity. 
 Let $(\gamma_n)_{n \geqslant 1}$ be a positive sequence; we consider the stochastic gradient recursion\footnote{The complexity of $n$ steps of the recursion is $O(n^2)$ if using kernel functions or $O(\tau n)$ when using explicit feature representations, with $\tau$ the complexity of computing dot-products  and adding feature vectors.} in $\H$ started at $g_0$:
\eqal{\label{eq:firstSGD}
{g}_n={g}_{n-1}-\frac{\gamma_n}{2} \nabla F_{n,\lambda}({g}_{n-1}) = {g}_{n-1}-\gamma_n \left[(\langle K_{x_n},{g}_{n-1}\rangle-y_n)K_{x_n} + \lambda ({g}_{n-1} - g_0) \right].
}
We are going to consider Polyak-Ruppert averaging \citep{polyak1992acceleration}, that is $\bar{g}_n = \frac{1}{n+1} \sum_{i=0}^{n} g_i$, as well as the tail-averaging estimate $\bar{g}_n^{\textrm {tail}} = \frac{1}{\lfloor n/2 \rfloor} \sum_{i=\lfloor n/2 \rfloor}^{n} g_i$, studied by \citet{jain2016parallelizing}. For the sake of clarity, all the results in the main text are for the tail averaged estimate but note that all of them have been also proved for the full average in Appendix \ref{ap:average}.

As explained earlier (see Lemma~\ref{lm:appr-correct-sign-to-01}), we need to show the convergence of ${g}_n$ to $g_\lambda$ in $\H$-norm. We are going to consider two cases: (1) for the non-averaged recursion $(\gamma_n)$ is a decreasing sequence, with the important particular case  $\gamma_n = \gamma/n^\alpha$, for $\alpha \in [0,1]$; (2) for the averaged or tail-averaged functions $(\gamma_n)$ is a constant sequence equal to $\gamma$.
For all the proofs of this section see Appendix  \ref{sec:AppSGD}.
In the next subsection we reformulate the recursion in Eq.~(\ref{eq:firstSGD}) as a least-squares recursion converging to $g_\lambda$.

\subsection{Reformulation as noisy recursion}

We can first reformulate the SGD recursion equation in Eq.~(\ref{eq:firstSGD}) as a regular least-squares SGD recursion with noise,
with the notation $\xi_n = y_n - \Phg (x_n)$, which satisfies $\E \big[ \xi_n K_{x_n}  \big] =0$. This is the object of the following lemma (for the proof see Appendix \ref{ap:SGDreformulation}.):

\begin{lemma}
\label{le:SGDrecursion}
The SGD recursion can be rewritten as follows:
\begin{align}
\label{eq:SGDrecursion}
    {g}_n - g_\lambda =  \big[
I - \gamma_n   (  K_{x_n} \otimes K_{x_n} + \lambda I ) \big]
  ( {g}_{n-1} - g_\lambda )   + \gamma_n \varepsilon_n,
\end{align}
with the noise term $\varepsilon_k  =   \xi_k K_{x_k}   +   (\Phg(x_k)- g_\lambda(x_k)) K_{x_k} - \E \left[ (\Phg(x_k)- g_\lambda(x_k)) K_{x_k} \right] \in \H.$
\end{lemma}
We are thus in presence of a least-squares problem in the Hilbert space $\H$, to estimate a function $g_\lambda \in \H$ with a specific noise $\varepsilon_n$ in the gradient and feature vector $K_{x}$. In the next section, we will consider the generic recursion above, which will require some bounds on the noise. In our setting, we have the following almost sure bounds and the noise (see Lemma \ref{le:noise} of  Appendix \ref{sec:AppSGD}):
\begin{align*}
 \| \varepsilon_n \|_\H &\leqslant R  ( 1 + 2 \| \Phg - g_\lambda\|_{L_\infty} )  \\
\E \big[ \varepsilon_n \otimes \varepsilon_n \big]
& \preccurlyeq 2 \left(1+\|\Phg- g_\lambda\|^2_\infty\right) \Sigma,
\end{align*} 
where $\Sigma = \E \big[ K_{x_n} \otimes K_{x_n} \big]$ is the covariance operator.

\subsection{SGD for general Least-Square problems}

\input sgd.tex

\section{Exponentially Convergent SGD for Classification error}
\label{sec:expo}

\input exp.tex



\section{Conclusion}
\label{sec:conc}

In this paper, we have shown that stochastic gradient could be exponentially convergent, once some margin conditions are assumed; and even if a weaker margin condition is assumed, fast rates can be achieved (see Appendix \ref{ap:weakmargin}). This is obtained by running averaged stochastic gradient on a least-squares problem, and proving new deviation inequalities.

Our work could be extended in several natural ways: (a) our work relies on new concentration results for the least-mean-squares algorithm (i.e., SGD for square loss), it is natural to extend it to other losses, such as the logistic or hinge loss; (b) going beyond binary classification is also natural with the square loss~\citep{ciliberto2016consistent,NIPS2017_6634} or without~\citep{taskar2005learning}; (c)~in our experiments, we use regularization, but we have experimented with unregularized recursions, which do exhibit fast convergence, but for which proofs are usually harder~\citep{dieuleveut2016nonparametric}; finally, (d) in order to avoid the $O(n^2)$ complexity, extending the results of~\citet{rudi2017falkon,rudi2017generalization} would lead to a subquadratic complexity.

\acks{We acknowledge support from the European Research Council (grant SEQUOIA 724063). We would like to thank Rapha\"el Berthier for useful discussions.}

\bibliography{exp_rates}

\appendix

\input app.tex

\end{document}

%% file: additional_defs_colt.tex
\newcommand{\Nystrom}[1]{{Nystr\"om}}

\providecommand{\scal}[2]{\left\langle{#1},{#2}\right\rangle}
\providecommand{\nor}[1]{\bigl\|{#1}\bigr\|}

\providecommand{\tr}{\operatorname{Tr}}

\newcommand{\R}{\mathbb R}

\newcommand{\N}{\mathbb N}

\newcommand{\hh}{\mathcal H}

\newcommand{\Phg}{\tilde{g}_*}

\newcommand{\la}{\lambda}

\newcommand{\expect}[1]{\mathbb{E}\left[{#1}\right]}
\newcommand{\condexp}[2]{\mathbb{E}\left({#1}\middle|{#2}\right)}

\newcommand{\rhox}{{\rho_{\X}}}

\renewcommand{\L}{{L}}

\newcommand{\eqals}[1]{\begin{align*}#1\end{align*}}
\newcommand{\eqal}[1]{\begin{align}#1\end{align}}
\newcommand{\bpr}{\begin{proof}}
\newcommand{\epr}{\end{proof}}
\newcommand{\be}{\begin{equation}}
\newcommand{\ee}{\end{equation}}

\newcommand{\bd}{\begin{definition}}
\newcommand{\ed}{\end{definition}}

\newcommand{\bi}{\begin{itemize}}
\newcommand{\ei}{\end{itemize}}

\newtheorem{ass}{Assumption}
\newcommand{\ba}{\begin{ass}}
\newcommand{\ea}{\end{ass}}

\newcommand{\br}{\begin{remark}}
\newcommand{\er}{\end{remark}}

\newcommand{\bp}{\begin{proposition}}
\newcommand{\ep}{\end{proposition}}

\newcommand{\blm}{\begin{lemma}}
\newcommand{\elm}{\end{lemma}}

\newcommand{\bt}{\begin{theorem}}
\newcommand{\et}{\end{theorem}}

\newcommand{\bcor}{\begin{corollary}}
\newcommand{\ecor}{\end{corollary}}

\newcommand{\bex}{\begin{example}}
\newcommand{\eex}{\end{example}}

\newtheorem{assshort}{\bfseries (A \hspace*{.5mm} ) $\!\!\!\!\!\!\!\!\!\!$ }
\newcommand{\bas}{\begin{assshort}}
\newcommand{\eas}{\end{assshort}}

\newcommand{\closs}{{\cal R}}

\newcommand{\T}{\Sigma}
\newcommand{\Tn}{\widehat{\Sigma}}
\newcommand{\Tl}{\Sigma_\lambda}
\newcommand{\Tnl}{\widehat{\Sigma}_\lambda}

\newcommand{\BIT}{\begin{itemize}}
\newcommand{\EIT}{\end{itemize}}
\newcommand{\BNUM}{\begin{enumerate}}
\newcommand{\ENUM}{\end{enumerate}}
\newcommand{\BA}{\begin{array}}
\newcommand{\EA}{\end{array}}

\newcommand{\sign}{\mathop{ \mathrm{sign}}}
\newcommand{\idm}{I}
\newcommand{\rb}{\mathbb{R}}

\newcommand{\mysec}[1]{Section~\ref{sec:#1}}

\def \E{{\mathbb E}}
\def \P{{\mathbb P}}

\def \L{  { \Lambda} }
\def \E{{\mathbb E}}
\def \P{{\mathbb P}}
\def \H{{\mathcal H}}
\def \X{{\mathcal X}}
\def \F{{\mathcal F}}

\def \X{{\mathcal X}}

\newcommand{\asm}[1]{{\textbf{(A\ref{#1})}}}

\newtheorem{sgdassshort}{\bfseries (H \hspace*{.5mm} ) $\!\!\!\!\!\!\!\!\!\!$ }

\newcommand{\basgd}{\begin{sgdassshort}}
\newcommand{\easgd}{\end{sgdassshort}}
\newcommand{\sgdasm}[1]{{\textbf{(H\ref{#1})}}}

%% file: sgd.tex

We now consider results on (averaged) SGD for least-squares that are interesting on their own. As said before, we show  results in two different settings depending on the step-size sequence. First,  we consider $(\gamma_n)$ as a decreasing sequence, second we take $(\gamma_n)$ constant but prove the convergence of the (tail-)averaged iterates.

Since the results we need could be of interest (even for finite-dimensional models), in this section, we study the following general recursion:
\begin{align}
\label{eq:SGDabstract}
\eta_n = ( \idm - \gamma H_n) \eta_{n-1} + \gamma_n \varepsilon_n,
\end{align}
We make the following assumptions:  
\vspace{-0.2cm}
\basgd \label{asm:init}  We start at some $\eta_0 \in \H$. \easgd
\vspace{-0.65cm}
\basgd \label{asm:noise-iid}   $(H_n,\varepsilon_n)_{n \geqslant 1} $ are  i.i.d. and $H_n$ is a positive self-adjoint operator so that almost surely $H_n \succcurlyeq \lambda \idm$, and $H  := \E H_n$. \easgd
\vspace{-0.5cm}
\basgd \label{asm:noise-bound}  Noise: $ \E  \varepsilon_n   = 0$, $\| \varepsilon_n  \|_\H \leqslant c^{1/2} $ almost surely and $\E  ( \varepsilon_n   \otimes \varepsilon_n ) \preccurlyeq C$, with $C$ commuting with $H$. Note that one consequence of this assumption is $\E \|\varepsilon_n\|_\H^2 \leqslant \tr C$. \easgd
\vspace{-0.5cm}
\basgd \label{asm:weird-bound}  For all $n \geqslant 1$, $\E \Big[  H_n C H^{-1} H_n \Big] \preccurlyeq \gamma_0^{-1} C$ and $\gamma \leqslant \gamma_0$. \easgd
\vspace{-0.5cm}
\basgd \label{asm:commute}  $A$ is a positive self-adjoint operator which commutes with $H$. \easgd
Note that we will later apply the results of this section to $H_n =  K_{x_n} \otimes K_{x_n} + \lambda I$, $H = \Sigma + \lambda \idm$, $C = \Sigma$ and $A \in \{\idm, \Sigma\}$. We first consider the non-averaged SGD recursion, then the (tail-)averaged recursion. The key difference with existing bounds is the need for precise probabilistic deviation results.

For least-squares, one can always separate the impact of the initial condition $\eta_0$ and of the noise terms $\varepsilon_k$, namely $\eta_n = \eta_n^{\textrm {bias}} + \eta_n^{\textrm {variance}}$, where $\eta_n^{\textrm {bias}}$ is the recursion with no noise ($\varepsilon_k = 0$), and $\eta_{n}^{\textrm {variance}}$ is the recursion started at $\eta_0=0$. The final performance will be bounded by the sum of the two separate performances \citep[see, e.g.,][]{defossez2014constant}. Hence all of our bounds will depend on these two. See more details in Appendix \ref{sec:AppSGD}.

\subsection{Non-averaged SGD}

In this section, we prove results for the recursion defined by Eq.~\eqref{eq:SGDabstract} in the case where for $\alpha \in [0,1]$, $ \gamma_n = \gamma/n^\alpha$. These results extend the ones of \citet{gradsto} by providing deviation inequalities, but are limited to least-squares. For general loss functions and in the strongly-convex case, see also~\citet{kakade2009generalization}.

\begin{theorem}[SGD, decreasing step size: $\gamma_n = \gamma/n^\alpha$] \label{th:SGDalpha}
Assume \sgdasm{asm:init}, \sgdasm{asm:noise-iid}, \sgdasm{asm:noise-bound}, $\gamma_n = \gamma/n^\alpha$, $\gamma\lambda < 1$ and denote by $\eta_n \in \H$ the n-th iterate of the recursion in Eq. \eqref{eq:SGDabstract}. We have for $t > 0, n \geqslant 1$ and  $\alpha \in (0,1)$, $\|g_n - g_\lambda\|_\H \leqslant \exp\left(  -\frac{\gamma\lambda}{1-\alpha}\left( (n+1)^{1-\alpha} -1  \right)   \right) \|g_0 - g_\lambda\|_\H + V_n$, almost surely for~$n$ large enough \footnote {See Appendix Section \ref{sec:AppSGD} Lemma \ref{le:nalpha} for more details.}, with
$\displaystyle \P \left( V_n \geqslant t \right) \leqslant 2\exp\left( -\frac{ t^2 }{ 8 \gamma \tr C/\lambda  + \gamma c^{1/2}t }\cdot n^{\alpha}\right).$
\end{theorem}
We can make the following observations: 
\vspace{-0.1cm}
\BIT
\itemsep-3pt
\item The proof technique (see Appendix \ref{ap:SGDalpha} for the detailed proof) relies on the following scheme: we notice that $\eta_n$ can be decomposed in two terms, (a) the bias: obtained from a product of $n$ contractant operators, and (b) the variance: a sum of increments of a martingale. We treat separately the two terms. For the second one, we prove almost sure bounds on the increments and on the variance that lead to a Bernstein-type concentration result on the tail $\P (V_n \geqslant t)$. Following this proof technique, the coefficient in the latter exponential is composed of the variance bound plus the almost sure bound of the increments of martingale times $t$.
\item Note that we only presented in Theorem \ref{th:SGDalpha} the case where $\alpha \in (0,1)$. Indeed, we only focused on the case where we had exponential convergence (see the whole result in the Appendix: Proposition \ref{prop:fullalpha}). Actually, that there are three different regimes. For $\alpha=0$ (constant step-size), the algorithm is not converging, as the tail probability bound on $\P \left( V_n \geqslant t \right) $ is not dependent on $n$. For $\alpha=1$, confirming results from~\citet{gradsto}, there is no exponential forgetting of initial conditions. And for $\alpha \in (0,1)$, the forgetting of initial conditions and the tail probability are converging to zero exponentially fast, respectively,  as $\exp( - C n^{1-\alpha})$ and $\exp( - C n^{\alpha})$, for a constant $C$, hence the natural choice of $\alpha=1/2$ in our experiments.

\EIT

\subsection{Averaged and Tail-averaged SGD with constant step-size}
 
In the subsection, we take: $\forall n \geqslant 1, \ \gamma_n = \gamma$. We first start with a result on the variance term, whose proof extends the work of \citet{daft} to deviation inequalities which are sharper than the ones from \cite{newsto}.

\begin{theorem}[Convergence of the variance term in averaged SGD]
\label{th:SGDaveraged}
Assume \sgdasm{asm:init}, \sgdasm{asm:noise-iid}, \sgdasm{asm:noise-bound},$\ $ \sgdasm{asm:weird-bound}, \sgdasm{asm:commute} and consider the average of the $n+1$ first iterates of the sequence defined in Eq. \eqref{eq:SGDabstract}: $\bar{\eta}_n = \frac{1}{n+1} \sum_{i=0}^n \eta_i$. Assume $\eta_0 = 0$.
 We have for $t > 0, n \geqslant 1$:    
\begin{equation}
\displaystyle\P \left( \left\| A^{1/2} \bar{\eta}_n  \right\|_\H \geqslant t \right) \leqslant 2 \exp\left[-\frac{(n+1) t^2}{E_t}\right],
\end{equation}
where $E_t$ is defined with respect to the constants introduced in the assumptions:  
\begin{equation}
\label{eq:Et}
 E_t =   4\tr(AH^{-2}C)+\frac{2c^{1/2} \|A^{1/2}\|_{\textrm {op}}}{3\lambda}\cdot t  .
 \end{equation}
\end{theorem}
The work that remains to be done is to bound the bias term of the recursion $\bar{\eta}_n^{\textrm {bias}}$. We have done it for the full averaged sequence (see Appendix \ref{ap:SGDfullaverage} Theorem \ref{th:withbias}) but as it is quite technical and could lower a bit the clarity of the reasoning, we have decided to leave it in the Appendix. We present here another approach and consider the tail-averaged recursion, $\bar{\eta}_n^{\textrm {tail}} 
= \frac{1}{\lfloor n/2 \rfloor} \sum_{i=\lfloor n/2 \rfloor}^{n} \eta_i$ \citep[as proposed by][]{jain2016parallelizing,shamir2011SGD}.
For this, we use the simple almost sure bound
$ \| \eta_i^{\textrm {bias}} \|_\H \leqslant ( 1- \lambda \gamma )^i \|\eta_0\|_\H$, such that 
$\| \bar{\eta}_n^{\textrm {tail, bias}} \|_\H \leqslant ( 1- \lambda \gamma )^{n/2} \|\eta_0\|_\H$. For the variance term, we can simply use the result above for $n$ and $n/2$, as 
$\bar{\eta}_n^{\textrm {tail}}  = 2 \bar{\eta}_n  - \bar{\eta}_{n/2}$. This leads to:
 \clearpage
 \begin{corollary}[Convergence of tail-averaged SGD]
 \label{co:SGDtailaveraged}
Assume \sgdasm{asm:init}, \sgdasm{asm:noise-iid}, \sgdasm{asm:noise-bound}, \sgdasm{asm:weird-bound}, \sgdasm{asm:commute} and consider the tail-average of the sequence defined in Eq. \eqref{eq:SGDabstract}: $\bar{\eta}_n^{\textrm {tail}} = \frac{1}{\lfloor n/2 \rfloor} \sum_{i=\lfloor n/2 \rfloor}^{n} \eta_i$.
 We have for $t > 0, n \geqslant 1$:   
\begin{eqnarray}
\left\|A^{1/2}\bar{\eta}_n^{\textrm {tail}} \right\|_\H &\leqslant& (1-\gamma\lambda)^{n/2} \|A^{1/2}\|_{op} \|\eta_0\|_\H + L_n \quad,\ \text{with} \\
 \P(L_n \geqslant t ) &\leqslant& 4\exp\left( -  (n+1)t^2 /( 4 E_t)\right),
\end{eqnarray}
where $L_n$ is defined in the proof (see Appendix \ref{ap:SGDcorrolary}) and is the variance term of the tail-averaged recursion.
\end{corollary}
We can make the following observations on the two previous results:
\vspace{-0.1cm}
\BIT
\itemsep-3pt
\item The proof technique (see Appendix \ref{ap:SGDaverage} and \ref{ap:SGDcorrolary} for the detailed proofs) relies on concentration inequality of Bernstein type. Indeed, we notice that (in the setting of  Theorem \ref{th:SGDaveraged}) $\bar{\eta}_n$ is a sum of increments of a martingale. We prove almost sure bounds on the increments and on the variance \citep[following the proof technique of][]{daft} that lead to a Bernstein type concentration result on the tail $\P (V_n \geqslant t)$. Following the proof technique summed-up before, we see that $E_t$ is composed of the variance bound plus the almost sure bound times $t$. 
\item Remark that classically, $A$ and $C$ are proportional to $H$ for excess risk predictions. In the finite $d$-dimensional setting this leads us to the usual variance bound proportional to the dimension~$d$: $\tr (AH^{-2}C) \cong \tr \idm = d$. The result is general in the sense that we can apply it for all  matrices $A$ commuting with $H$ (this can be used to prove results in $L_2$ or in~$\H$). 
\item Finally, note that we improved the variance bound with respect to the strong convexity parameter $\lambda$ which is usually of the order $1/\lambda^2$ \citep[see][]{shamir2011SGD}, and is here $\tr (AH^{-2}C)$. Indeed, in our setting, we will apply it for $A = C = \Sigma$ and $H = \Sigma + \lambda I$, so that $\tr (AH^{-2}C)$ is upper bounded by the effective dimension $\tr (\Sigma (\Sigma + \lambda I)^{-1})$ which can be way smaller than $1/\lambda^2$ \citep[see][]{caponnetto2007optimal,dieuleveut2016nonparametric}.
\item The complete proof for the full average is written in Appendix \ref{ap:SGDfullaverage} and more precisely in Theorem \ref{th:withbias}. In this case the initial conditions are not forgotten exponentially fast though.
\EIT

%% file: exp.tex

In this section we want to show our main results, on the error made (on unseen data) by the $n$-th iterate of the regularized SGD algorithm. Hence, we go back to the original SGD recursion defined in Eq.~(\ref{eq:SGDrecursion}). Let us recall it:
\begin{equation*}
    {g}_n - g_\lambda =  \big[
I - \gamma_n   (  K_{x_n} \otimes K_{x_n} + \lambda I ) \big]
  ( {g}_{n-1} - g_\lambda )   + \gamma_n \varepsilon_n,
\end{equation*}
with the noise term $\varepsilon_k  =   \xi_k K_{x_k}   +   (\Phg(x_k)- g_\lambda(x_k)) K_{x_k} - \E \left[ (\Phg(x_k)- g_\lambda(x_k)) K_{x_k} \right] \in \H.$ Like in the previous section we are going to state two results in two different settings, the first one for SGD with decreasing step-size ($\gamma_n = \gamma / n^\alpha$) and the second one for tail averaged SGD with constant step-size. For all the proofs of this section see the Appendix (section \ref{sec:error}).

\subsection{SGD with decreasing step-size}

In this section, we focus on decreasing step-sizes $\gamma_n = \gamma/n^\alpha$ for $\alpha \in (0,1)$, which lead to exponential convergence rates. Results for $\alpha=1$ and $\alpha=0$ can be derived in a similar way (but do not lead to exponential rates).
\clearpage
\begin{theorem}
\label{th:erroralpha}
Assume \asm{asm:separability}, \asm{asm:kernel-bounded}, \asm{asm:data-iid}, \asm{asm:flambda-correct-sign} and $\gamma_n = \gamma/n^\alpha$, $\alpha \in (0,1)$ for any $n$ and $\gamma\lambda < 1$. Let $g_n$ be the n-th iterate of the recursion defined in Eq.~\eqref{eq:SGDrecursion}, as soon as $n$ satisfies the inequality $\exp\left(  -\frac{\gamma\lambda}{1-\alpha}\left( (n+1)^{1-\alpha} -1  \right)   \right) \leqslant \delta / (5R\|g_0- g_\lambda\|_\H)$, then 
\begin{align*}
\closs(g_n) = \closs^*, \mbox{ with probability at least }1-2\exp\left( -\frac{ \delta^2}{C_R}\cdot n^{\alpha}\right),
\end{align*}
with $C_R = 2 ^{\alpha + 7}\gamma R^2 \tr \Sigma  \left(1+\|\Phg- g_\lambda\|^2_\infty\right)/\lambda+8\gamma R^2 \delta( 1 + 2 \| \Phg - g_\lambda\|_\infty )/3 $,  and in particular 
$$ \expect{\closs(g_n) - \closs^*} \leqslant 2\exp\left( -\frac{ \delta^2}{C_R}\cdot n^{\alpha}\right).$$
\end{theorem}
Note that Theorem \ref{th:erroralpha} shows that with probability at least $1-2\exp\left( -\frac{ \delta^2}{C_R}\cdot n^{\alpha}\right)$, the predictions of $g_n$ are perfect. We can also make the following observations:
\vspace{-0.1cm}
\BIT
\itemsep-3pt

\item The idea of the proof (see Appendix \ref{ap:EXPalpha} for the detailed proof) is the following: we know that as soon as $\|g_n - g_\lambda\|_\H \leqslant \delta/(2R)$, the predictions of $g_n$ are perfect (Lemma \ref{lm:appr-correct-sign-to-01}). We just have to apply Theorem \ref{th:SGDalpha} for to the original SGD recursion and make sure to bound each term by $\delta/(4R)$. Similar results for non-averaged SGD could be derived beyond least-squares (e.g., hinge or logistic loss) using results from~\citet{kakade2009generalization}. 

\item Also note that the larger the $\alpha$, the smaller the bound. However, it is only valid for $n$ larger that a certain quantity depending of $\lambda \gamma$. A good trade-off is $\alpha=1/2$, for which we get an excess error  of $2\exp\left( -\frac{ \delta ^2}{C_R} n^{1/2}\right)$, which is valid as soon as 
$n \geqslant \log(10 R \|g_0 - g_\lambda\|_\H / \delta)/(4\lambda^2\gamma^2)$. Notice also that we should go for large $\gamma \lambda $ to increase the factor in the exponential and make the condition happen as soon as possible.
\item If we want to emphasize the dependence of the bound on the important parameters, we can write that: $\expect{\closs(g_n) - \closs^*} \lesssim 2\exp\left( -\lambda \delta^2n^{\alpha}/R^2\right).$
\item When the condition on $n$ is not met, then we still have the usual bound obtained by taking directly the excess loss \citep{bartlett2006convexity} but we lose exponential convergence.

\EIT

\subsection{Tail averaged SGD with constant step-size}
We now consider the tail-averaged recursion\footnote{The full averaging result corresponding to Theorem \ref{th:errortail} is proved in Appendix \ref{ap:EXPfullaverage}, Theorem \ref{th:expwithbias}.}, with the following result:

\begin{theorem}
\label{th:errortail}
Assume \asm{asm:separability}, \asm{asm:kernel-bounded}, \asm{asm:data-iid}, \asm{asm:flambda-correct-sign} and $\gamma_n = \gamma$ for any n, $\gamma\lambda < 1$ and $\gamma \leqslant \gamma_0 = (R^2 + 2\lambda)^{-1}$. Let $g_n$ be the n-th iterate of the recursion defined in Eq.~(\ref{eq:SGDrecursion}), and $\bar{g}_n^{\textrm {tail}} = \frac{1}{\lfloor n/2 \rfloor} \sum_{i=\lfloor n/2 \rfloor}^{n} g_i$, as soon as $ n \geqslant 2/(\gamma\lambda)\ln (5R \|g_0-g_\lambda\|_\H/\delta)$, then 
$$ \closs(\bar{g}_n^{\textrm {tail}}) = \closs^*, \mbox{ with probability at least }1-4\exp\left( - \delta^2 K_R (n+1)\right),$$
with $K_R^{-1} =2^9 R ^2  \left(1+\|\Phg- g_\lambda\|^2_\infty\right)\tr(\Sigma(\Sigma + \lambda \idm)^{-2})+32 \delta R^2 ( 1 + 2 \| \Phg - g_\lambda\|_\infty )/ (3\lambda),$ and in particular
$$ \expect{\closs(\bar{g}_n^{\textrm {tail}}) - \closs^*} \leqslant 4\exp\left( - \delta^2 K_R (n+1)\right).$$
\end{theorem}
\clearpage
Theorem \ref{th:errortail} shows that  with probability at least $1-4\exp\left( - \delta^2 K_R (n+1)\right)$, the predictions of $\bar{g}_n^{\textrm {tail}}$ are perfect. We can also make the following observations:
\vspace{-0.2cm}
\BIT
\itemsep-3pt

\item The idea of the proof (see Appendix \ref{ap:EXPaverage} for the detailed proof) is the following: we know that as soon as $\|\bar{g}_n^{\textrm {tail}} - g_\lambda\|_\H \leqslant \delta/(2R)$, the predictions of $\bar{g}_n^{\textrm {tail}}$ are perfect (Lemma \ref{lm:appr-correct-sign-to-01}). We just have to apply  Corollary \ref{co:SGDtailaveraged} to   the original SGD recursion, and make sure to bound each term by $\delta/(4R)$. 
\item If we want to emphasize the dependence of the bound on the important parameters, we can write that: $\expect{\closs(g_n) - \closs^*} \lesssim 2\exp\left( -\lambda^2 \delta^2n/R^4\right).$ Note that the $ \lambda^2 $ could be made much smaller with assumptions on the decrease of eigenvalues of $\Sigma$ \citep[it has been shown][that if the decay happens at speed $1/n^{\beta}$: $\tr \Sigma (\Sigma + \lambda \idm)^{-2} \leqslant \lambda^{-1}\tr \Sigma (\Sigma + \lambda \idm)^{-1}\leqslant R^2 / \lambda^{1+1/\beta}$]{caponnetto2007optimal}.
\item We want to take $\gamma \lambda$ as big as possible to satisfy quickly the condition. In comparison to the convergence rate in the case of decreasing step-sizes, the dependence on $n$ is improved as the convergence is really an exponential of $n$ (and not of some power of $n$ as in the previous result). 
\item Finally, the complete proof for the full average is contained in Appendix \ref{ap:EXPfullaverage} and more precisely in Theorem \ref{th:expwithbias}.

\EIT

%% file: app.tex
 
\clearpage

{\LARGE Organization of the Appendix}

\vspace{0.25cm}

\begin{enumerate}

\item[\ref{ap:experiments}.] {\em Experiments} \\where the experiments and their settings are explained.

\item[\ref{sec:proba}.] {\em Probabilistic lemmas} \\where concentration inequalities in Hilbert spaces used in section \ref{sec:AppSGD} are recalled.

\item[\ref{sect:proof-A5-to-01}.] {\em From $\H$ to 0-1 loss} \\where, from high probability bound for $\|\cdot\|_\H$, we derived bound for the 0-1 error.

\item[\ref{sect:exp-rates-for-KRR}.] {\em Proofs of Exponential rates for Kernel Ridge Regression} \\where exponential rates for Kernel Ridge Regression are proven (Theorem \ref{thm:exp-class-krls}).

\item[\ref{sect:examples-for-glambda}.] {\em Proofs and additional results about concrete examples} \\where additional results and croncrete examples to satisfy \asm{asm:flambda-correct-sign} are given.

\item[\ref{ap:SGDdevelopment}.] {\em Preliminaries for Stochastic Gradient Descent} \\where the SGD recursion is derived.

\item[\ref{sec:AppSGD}.] {\em Proof of stochastic gradient descent results} \\ where high probability bounds for the general SGD recursion are shown (Theorems \ref{th:SGDalpha} and \ref{th:SGDaveraged}).

\item[ \ref{sec:error}.] {\em Exponentially convergent SGD for classification error} \\where exponential convergence of test error are shown (Theorems \ref{th:erroralpha} and \ref{th:errortail}).

\item[\ref{ap:average}.] {\em Extension for the full averaged case} \\where previous results are extended for full averaged SGD (instead of tail-averaged).

\item[\ref{ap:weakmargin}.] {\em Convergence under weaker margin assumption} \\where previous results are extended in the case of a weaker margin assumption.
\end{enumerate}

\section{Experiments} \label{ap:experiments}

To illustrate our results, we consider one-dimensional synthetic examples ($\X = [0,1] $) for which our assumptions are easily satisfied. 
Indeed, we consider the following set-up that fulfils our assumptions: 
\BIT
\item \asm{asm:separability}, \asm{asm:data-iid} We consider here $X \sim U\left([0,(1-\varepsilon)/2] \cup [(1+\varepsilon)/2,1]  \right)$ and with the notations of Example \ref{ex:independent-noise-on-labels}, we take $\mathcal{S}_+ = [0,(1-\varepsilon)/2]$ and $\mathcal{S}_- = [(1+\varepsilon)/2,1]$. For $1 \leq i \leq n$,  $x_i$ independently sampled from $\rhox$ we define $y_i = 1 $ if $x_i \in \mathcal{S}_+$ and $y_i = -1 $ if $x_i \in \mathcal{S}_-$.

\item \asm{asm:kernel-bounded} We take the kernel to be the exponential  kernel $K(x,x') = \exp(-|x-x'|)$ for which the RKHS is a Sobolev space $\H = W^{s,2}$, with $s > d/2$, which is dense in $L_2$~\citep{adams2003sobolev}.

\item \asm{asm:flambda-correct-sign} With this setting we could find a closed form for $g_\lambda$ and checked that it verified \asm{asm:flambda-correct-sign}. Indeed we could solve the optimality equation satisfied by $g_\lambda$ : $$ \forall z \in [0,1], \ \int_{0}^1 K(x,z)g_\lambda(x) d\rho_X(x) + \lambda g_\lambda(z) = \int_{0}^1 K(x,z)g_\rho(x) d\rho_X(x),  $$ the solution being a linear combination of exponentials in each set : $[0,(1-\varepsilon)/2]$, $[(1-\varepsilon)/2,(1+\varepsilon)/2]$ and $[(1+\varepsilon)/2,1]$.
\EIT

\begin{figure}[ht]
    \centering
    \includegraphics[width=0.45\textwidth]{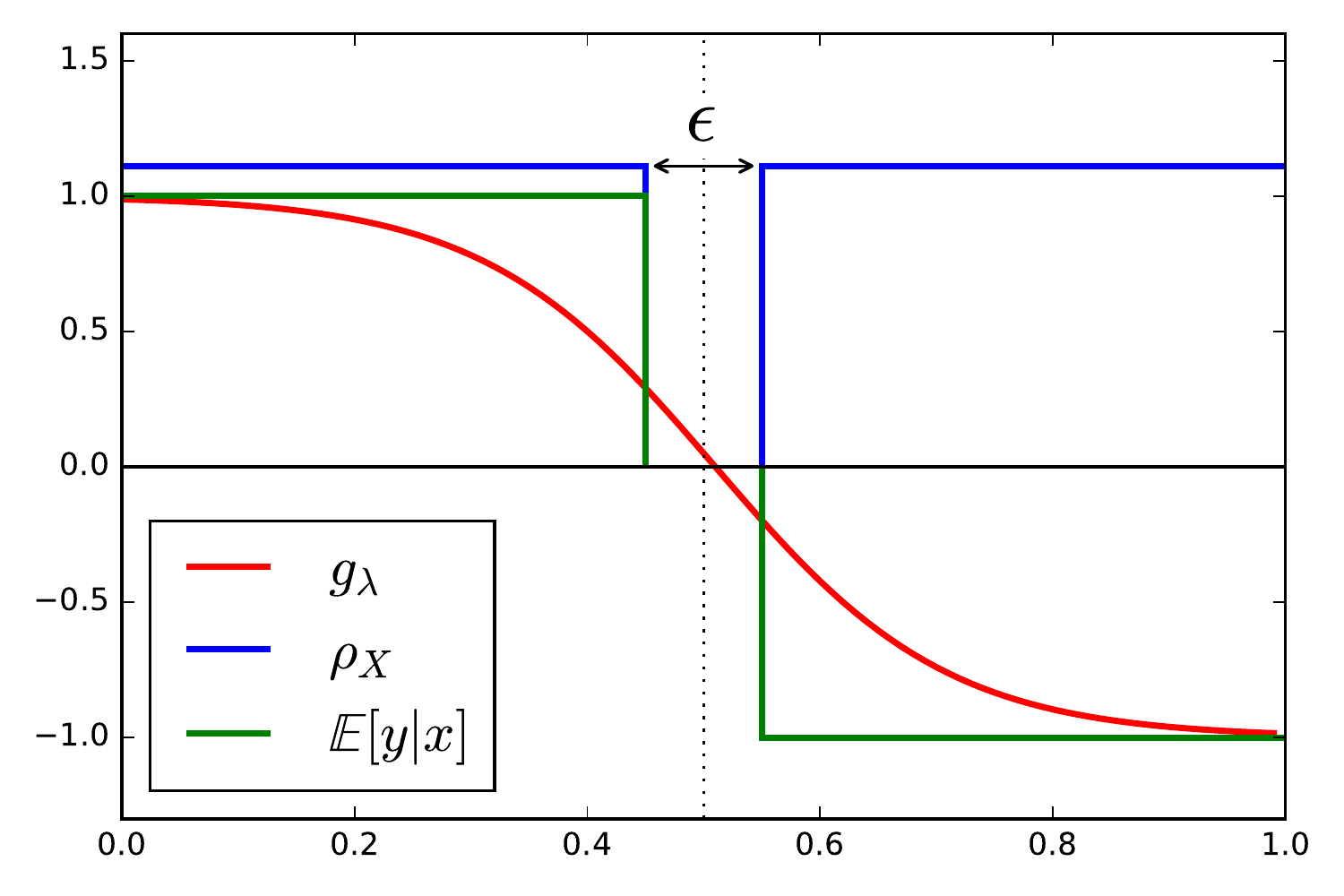}
    \caption{Representing the $\rho_\X$ density (uniform with $\varepsilon$-margin), the best estimator, i.e., $\E (x|y)$ and $g_\lambda$ used for the simulations ($\lambda = 0.01$).}
    \label{fig:densities}
\end{figure}

In the case of SGD with decreasing step size, we computed only the test error $\mathbb{E}({\cal R}(g_n)-{\cal R}^*))$. For tail averaged SGD with constant step size, we computed the test error as well as the training error, the test loss (which corresponds to the $L_2$ loss : $\int_0^1 (g_n (x) - g_\lambda(x))^2d\rho(x)$) and the training loss.
In all cases we computed the errors of the $n$-th iterate with respect to the calculated $g_\lambda$, taking $g_0 = 0$. For any $n \geqslant 1$,
$$g_n  =  {g}_{n-1} - \gamma_n  \big[ ( {g}_{n-1}(x_n)- y_n)K_{x_n}   + \lambda{g}_{n-1}  \big]. $$

\begin{figure}[ht]
\footnotesize
\stackunder[1pt]{\includegraphics[width=0.48\textwidth]{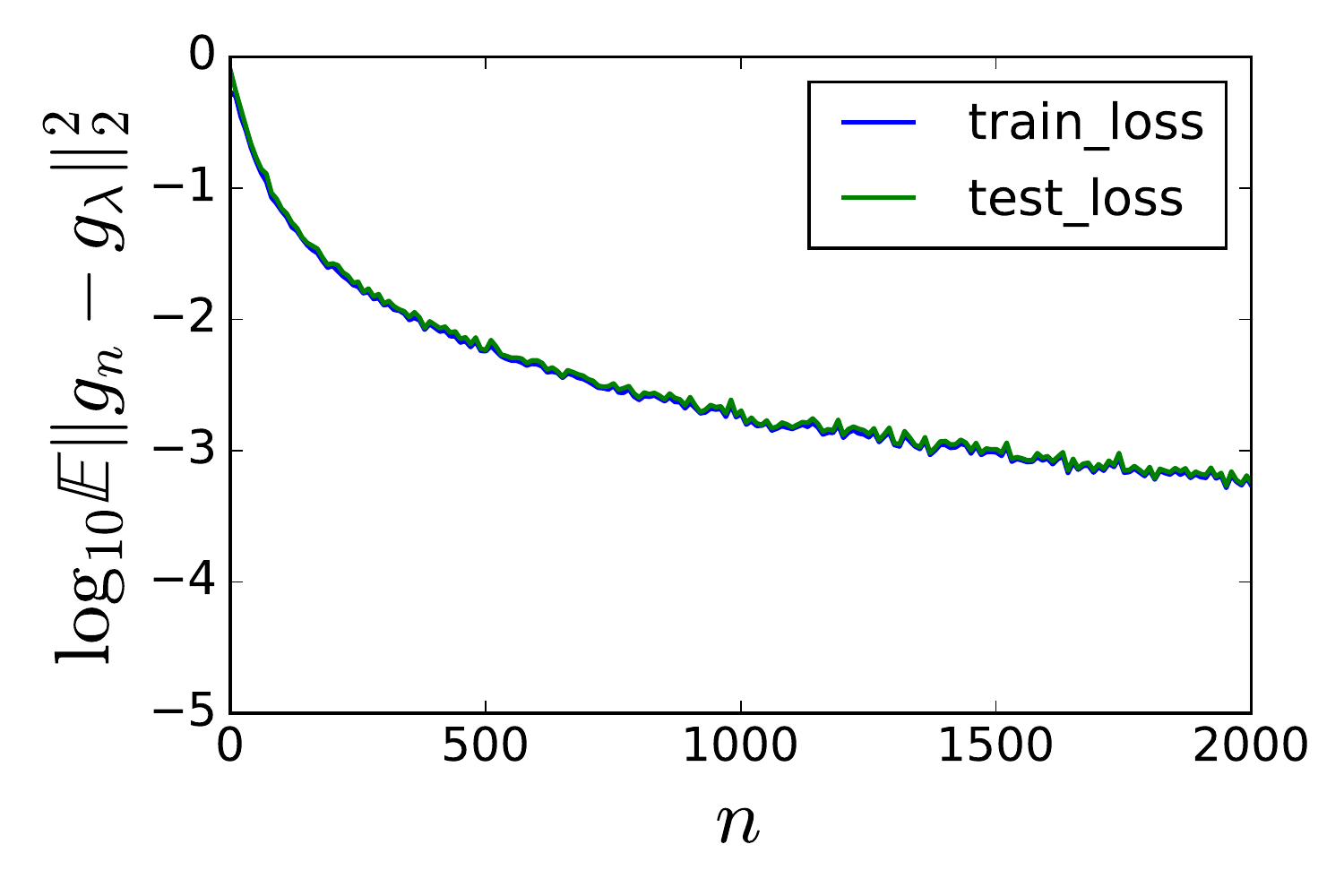}}{}%
\hspace{1cm}%
\stackunder[1pt]{\includegraphics[width=0.48\textwidth]{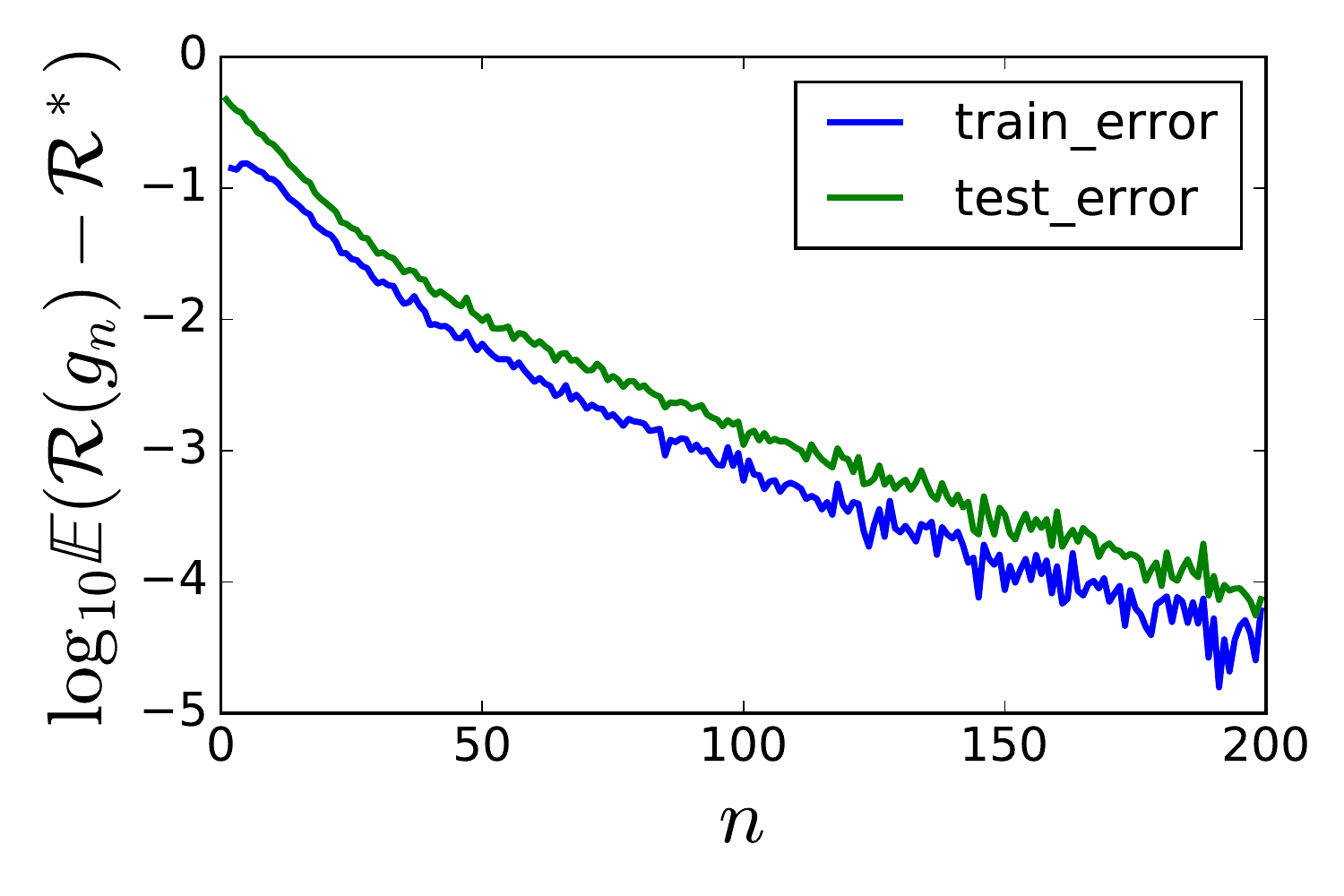}}{}%
\vspace{-0.5cm}
\caption{ \small Showing linear convergence for the $L^{01}$ errors in the case of margin of width $\varepsilon$. {\bfseries Left} figure corresponds to the test and training loss in the averaged case whereas the {\bfseries right} one corresponds to the error in the same setting. Note that the y-axis is the same while the x-axis is different of a factor 10. The fact that the error plot is a line after a certain $n$ matches our theoretical results. We took the following parameters :  $\varepsilon = 0.05$, $\gamma = 0.25$, $\lambda = 0.01$.}
\label{fig:plots}
\end{figure}

\begin{figure}[ht]
\footnotesize
\stackunder[1pt]{\includegraphics[width=0.48\textwidth]{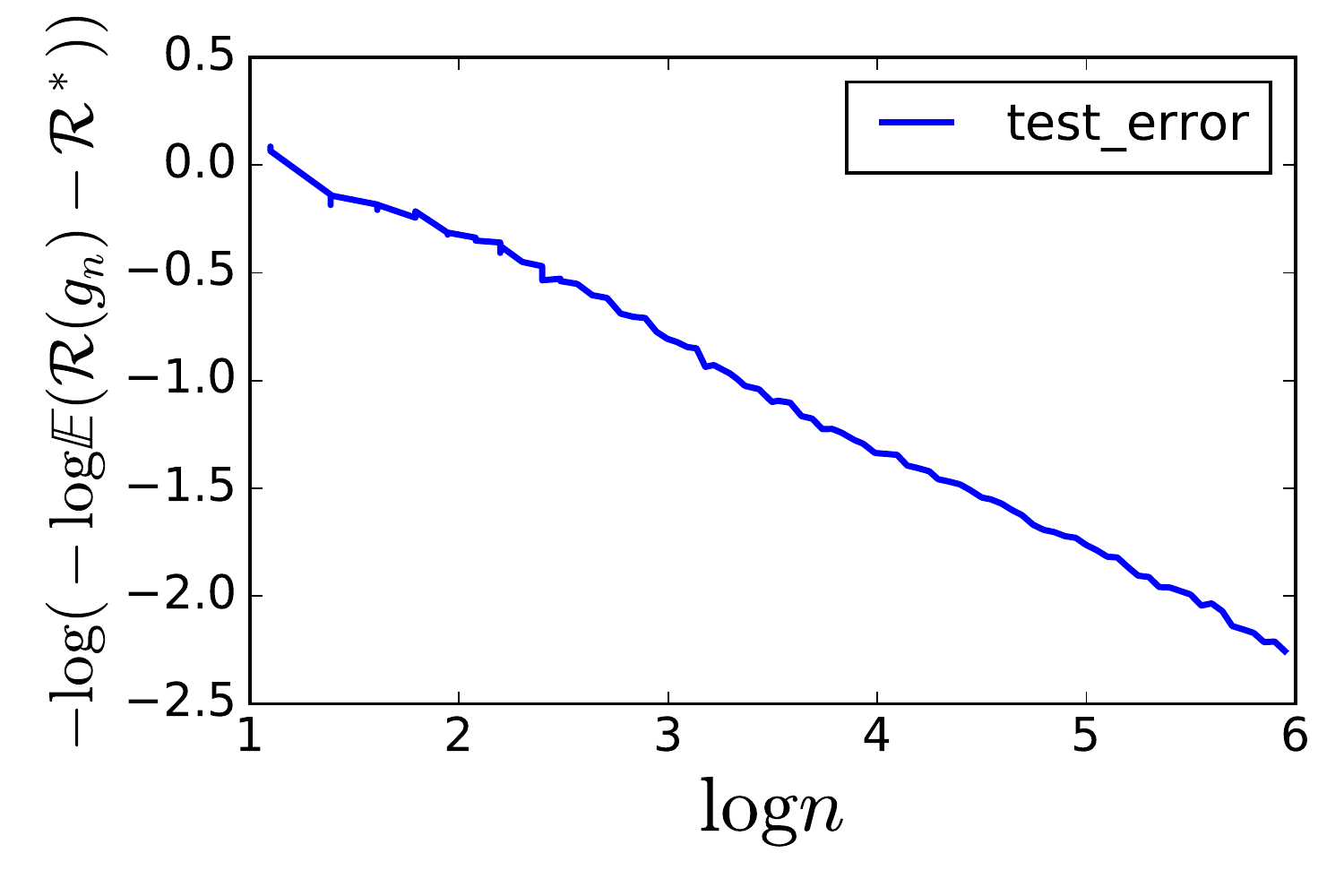}}{}%
\hspace{1cm}%
\stackunder[1pt]{\includegraphics[width=0.48\textwidth]{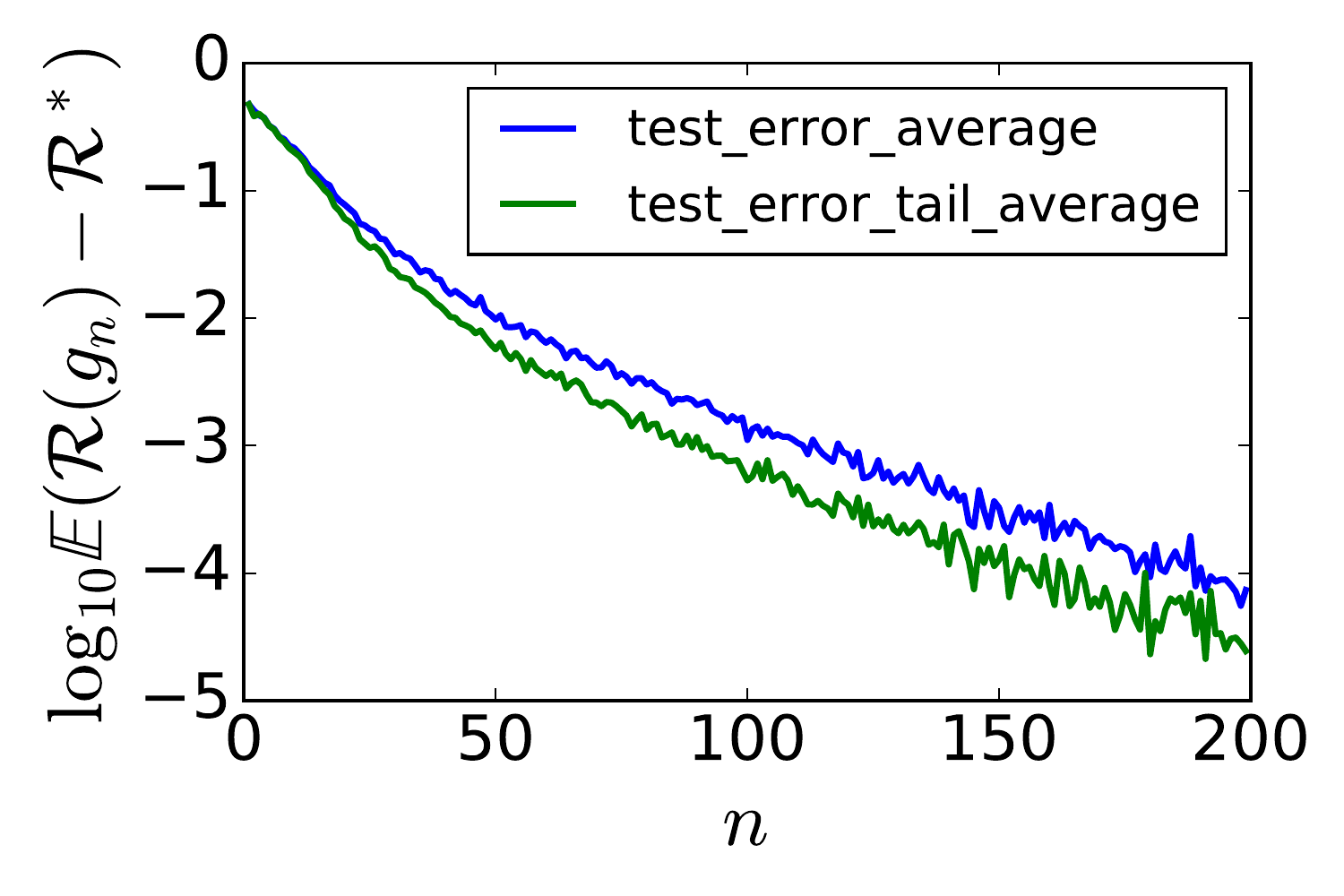}}{}
\vspace{-0.8cm}
\caption{ \small {\bfseries Left} plot shows the error in the non-averaged case for $\gamma_n = \gamma / \sqrt{n}$ and {\bfseries right} compares the test error between averaged and tail averaged case. We took the following parameters :  $\varepsilon = 0.05$, $\gamma = 0.25$, $\lambda = 0.01$.}
\label{fig:techplots}
\end{figure}

We can use representants to find the recursion on the coefficients. Indeed, if  $g_n = \sum_{i = 1}^n a^n_i K_{x_i},$ then the following recursion for the $(a_i^n)$ reads : 
\begin{eqnarray*}
\text{for } i \leqslant n-1, \ a_i^n &=& (1-\gamma_n \lambda) a_i^{n-1} \\
a_n^n &=& -\gamma_n (\sum_{i = 1}^{n-1} a_i^{n-1} K(x_n,x_i)-y_n).
\end{eqnarray*}
From $(a_i^n)$, we can also compute the coefficients of $\bar{g}_n$ and $\bar{g}_n^{\textrm {tail}}$ that we note $\bar{a}^n_i$ and $\bar{a}^{n,\textrm {tail}}_i$ respectively: $\bar{a}^n_i = \sum_{k = i}^n \frac{a_i^k}{n+1}$ and $\bar{a}^{n,\textrm {tail}}_i = \frac{1}{\lfloor n/2 \rfloor}\sum_{k = \lfloor n/2 \rfloor }^n a_i^k.$
To show our theoretical results we have decided to present the following figures: 

\BIT
\item For the exponential convergence of the averaged and tail averaged cases, we plotted the error $\log_{10}\mathbb{E}({\cal R}(g_n)-{\cal R}^*))$ as a function of $n$. With this scale and following our results it goes as a line after a certain $n$ (Figures \ref{fig:plots} and \ref{fig:techplots} right).
\item We recover the results of \citet{daft} that show convergence at speed $1/n$ for the loss (Figure \ref{fig:plots} left). We adapted the scale to compare with the error plot.
\item For Figure \ref{fig:techplots} left, we plotted $-\log (- \log (\mathbb{E}({\cal R}(g_n)-{\cal R}^*)) ) $ of the excess error with respect to the $\log$ of $n$ to show a line of slope $-1/2$. It meets our theoretical bound of the form $\exp(-K\sqrt{n})$, 
\EIT

Note that for the plots where we plotted the expected excess errors, i.e., $\mathbb{E}({\cal R}(g_n)-{\cal R}^*)$, we plotted the mean of the errors over 1000 replications until $n = 200$, whereas for the plots where we plotted the losses, i.e., a function of $\|g_n- g_*\|_2$, we plotted the mean of the loss over 100 replications until $n = 2000$.

We can make the following observations:

First remark that between plots of losses and errors (Figure \ref{fig:plots} left and right resp.), there is a factor~10 between the numbers of samples (200 for errors and 2000 for losses) and another factor~10 between errors and losses ($10^{-4}$ for errors and $10^{-3}$ for losses). That underlines well our theoretical result which is the difference between exponential rates of convergence of the excess error and $1/n$ rate of convergence of the loss.

Moreover, we see that even if the excess error with tail averaging seems a bit faster, we have linear rates too for the convergence of the excess error in the averaged case. Finally, we remark that the error on the train set is always below the one for a unknown test set (of what seems to be close to a factor 2).

 \section{Probabilistic lemmas} \label{sec:proba}

In this section we recall two fundamental results for concentration inequalities in Hilbert spaces shown in \citet{pinelis1994optimum}.

\begin{proposition}
\label{Probabilisticprop}
Let $(X_k)_{k \in \mathbb{N}}$ be a sequence of vectors of $\mathcal{H}$ adapted to a non decreasing sequence of $\sigma$-fields $(\mathcal{F}_k)$ such that $\E \left[ X_k | \mathcal{F}_{k-1} \right] = 0$, $\sup_{k \leqslant n} \| X_k \| \leqslant a_n$ and $\sum_{k = 1}^n \E \left[ \|X_k\|^2 | \mathcal{F}_{k-1} \right] \leqslant b_n^2$ for some sequences $(a_n),(b_n) \in \left(\mathbb{R}_+^*\right)^{\mathbb{N}}$. Then, for all $t  \geqslant 0$, $n \geqslant 1$,
\begin{eqnarray}
\P \left( \left\| \sum_{k = 1}^n X_k  \right\| \geqslant t \right) \leqslant 2 \exp\left( \frac{t}{a_n} - \left(\frac{t}{a_n} + \frac{b_n^2}{a_n^2}\right)\ln \left( 1+ \frac{t a_n}{b_n} \right)\right).
\end{eqnarray}  
\end{proposition}
\begin{proof}
As $\E \left[ X_k | \mathcal{F}_{k-1} \right] = 0$, the $\mathcal{F}_j$-adapted sequence $(f_j)$ defined by $f_j = \sum_{k = 1}^j X_k$ is a martingale and so is the stopped-martingale $(f_{j\wedge n})$. By applying Theorem 3.4 of \citet{pinelis1994optimum} to the martingale $(f_{j\wedge n})$, we have the result.
\end{proof}

\begin{corollary}
\label{Probabilisticcor}
Let $(X_k)_{k \in \mathbb{N}}$ be a sequence of vectors of $\mathcal{H}$ adapted to a non decreasing sequence of $\sigma$-fields $(\mathcal{F}_k)$ such that $\E \left[ X_k | \mathcal{F}_{k-1} \right] = 0$, $\sup_{k \leqslant n} \| X_k \| \leqslant a_n$ and $\sum_{k = 1}^n \E \left[ \|X_k\|^2 | \mathcal{F}_{k-1} \right] \leqslant b_n^2$ for some sequences $(a_n),(b_n) \in \left(\mathbb{R}_+^*\right)^{\mathbb{N}}$. Then, for all $t \geqslant 0$, $n \geqslant 1$,
\begin{eqnarray}
\P \left( \left\| \sum_{k = 1}^n X_k  \right\| \geqslant t \right) \leqslant 2 \exp\left(-\frac{t^2}{2\left( b_n^2 + a_n t / 3 \right)}\right).
\end{eqnarray}  
\end{corollary}

\begin{proof}
We apply \ref{Probabilisticprop} and simply notice that 
\begin{eqnarray*}
\frac{t}{a_n} - \left(\frac{t}{a_n} + \frac{b_n^2}{a_n^2}\right)\ln \left( 1+ \frac{t a_n}{b_n} \right) &=& -\frac{b_n^2}{a_n^2}\left(\left(1+\frac{a_n t}{b_n^2}\right)\ln\left(1+ \frac{a_n t}{b_n^2}\right)  -\frac{a_n t}{b_n^2}\right) \\
&=& -\frac{b_n^2}{a_n^2}\phi\left(\frac{a_n t}{b_n^2}\right),
\end{eqnarray*}
where $\phi(u) = (1+u) \ln (1+u) - u$ for $u > 0$. Moreover $\phi (u) \geqslant\displaystyle \frac{u^2}{2\left( 1+u/3 \right)}$, so that:
\begin{eqnarray*}
\frac{t}{a_n} - \left(\frac{t}{a_n} + \frac{b_n^2}{a_n^2}\right)\ln \left( 1+ \frac{t a_n}{b_n} \right) \leqslant -\frac{b_n^2}{a_n^2}\frac{(a_n t/b_n^2)^2}{2\left( 1+a_n t/3b_n^2 \right)} = -\frac{t^2}{2\left( b_n^2+a_n t/3 \right)}.
\end{eqnarray*}
\end{proof}

\section{From \texorpdfstring{$\H$}{H} to 0-1 loss}\label{sect:proof-A5-to-01}

In this section we prove Lemma~\ref{lm:appr-correct-sign-to-01}. Note that \asm{asm:flambda-correct-sign} requires the existence of $g_\la$ having the same sign of $g_\ast$ almost everywhere on the support of $\rhox$ and with absolute value uniformly bounded from below. In Lemma~\ref{lm:appr-correct-sign-to-01} we prove that we can bound the 0-1 error with respect to the distance in $\hh$ of the estimator $\widehat{g}$ form $g_\la$. 

\bpr{\bfseries of Lemma~\ref{lm:appr-correct-sign-to-01}}
Denote by $W$ the event such that $\nor{\widehat{g} - g_\la}_\hh < \delta/(2R)$. Note that for any $f \in \hh$, 
$$f(x) = \scal{f}{K_x}_\hh \leq \nor{K_x}_\hh \nor{f}_\hh \leq R \nor{f}_\hh,$$
for any $x \in \X$. So for $\widehat{g} \in W$, we have
$$ |\widehat{g}(x) - g_\la(x)| \leq R \nor{\widehat{g}-g_\la}_\hh  < \delta/2 \quad \forall x \in \X.$$

Let $x$ be in the support of $\rhox$. By \asm{asm:flambda-correct-sign} $|g_\la(x)| \geq \delta/2$ a.e.. Let $\widehat{g} \in W$ and $x \in \X$ such that $g_\la(x) > 0$, we have
$$\widehat{g}(x) = g_\la(x) - (g_\la(x) - \widehat{g}(x)) \geq g_\la(x) - |g_\la(x) - \widehat{g}(x)| > 0,$$
so $\sign(\widehat{g}(x)) = \sign(g_\la(x)) = +1$. Similarly let $\widehat{g} \in W$ and $x \in \X$ such that $g_\la(x) < 0$, we have
$$\widehat{g}(x) = g_\la(x) + ( \widehat{g}(x) - g_\la(x)) \leq g_\la(x) + |g_\la(x) - \widehat{g}(x)| < 0,$$
so $\sign(\widehat{g}(x)) = \sign(g_\la(x)) = -1$. Finally note that for any $\widehat{g} \in \hh$, by \asm{asm:flambda-correct-sign}, either $g_\la(x) > 0$ or $g_\la(x) < 0$ a.e., so $\sign(\widehat{g}(x)) = \sign(g_\la(x))$ a.e.

Now note that by \asm{asm:separability}, \asm{asm:flambda-correct-sign} we have that $\sign(g_\ast(x)) = \sign(g_\la(x))$ a.e., where $g_\ast(x):= \condexp{y}{x}$. So when $\widehat{g} \in W$, we have that $\sign(\widehat{g}(x)) = \sign(g_\la(x)) = \sign(g_\ast(x))$ a.e., so
$$ \closs(\widehat{g}) = \rho(\{(x,y): \sign(\widehat{g}(x)) \neq y\}) =  \rho(\{(x,y): \sign(g_\ast(x)) \neq y\}) = \closs^*.$$

Finally note that
$$ \expect{\closs(\widehat{g})} = \expect{\closs(\widehat{g}){\mathbf 1}_W} + \expect{\closs(\widehat{g}){\mathbf 1}_{W^c}},$$
where ${\mathbf 1}_W$ is $1$ on the set $W$ and $0$ outside, $W^c$ is the complement set of $W$. 
So, when $\widehat{g} \in W$, we have
$$ \expect{\closs(\widehat{g}){\mathbf 1}_W} = \closs^*\expect{{\mathbf 1}_W} \leq \closs^*,$$
while
$$\expect{\closs(\widehat{g}){\mathbf 1}_{W^c}} \leq \expect{{\mathbf 1}_{W^c}} \leq q.$$
\epr

\section{Exponential rates for Kernel Ridge Regression}\label{sect:exp-rates-for-KRR}

\subsection{Results}

In this section, we first specialize some results already known in literature about the consistency of kernel ridge least-squares regression (KRLS) in $\hh$-norm \citep{caponnetto2007optimal} and then we derive exponential classification learning rates.
Let $(x_i,y_i)_{i=1}^n$ be $n$ examples independently and identically distributed according to $\rho$, that is Assumption \asm{asm:data-iid}. Denote by $\Sigma, \widehat{\Sigma}$ the linear operators on $\hh$ defined by
$$ \widehat{\Sigma} = \frac{1}{n} \sum_{i=1}^n K_{x_i} \otimes K_{x_i}, \quad \Sigma = \int_\X (K_x \otimes K_x )d\rhox(x),$$
referred to as the covariance and empirical (non-centered) covariance operators \citep[see][and references therein]{fukumizu2004dimensionality}.
We recall that the KRLS estimator $\widehat{g}_\la \in \hh$, which minimizes the regularized empirical risk, is defined as follows in terms of~$\widehat{\Sigma}$,
$$ \widehat{g}_\la = (\widehat\Sigma + \la I)^{-1} \left(\frac{1}{n} \sum_{i=1}^n y_i K_{x_i}\right).$$
Moreover we recall that the population regularized estimator $g_\la$ is characterized by \citep[see][]{caponnetto2007optimal}
$$ g_\la = (\Sigma + \la I)^{-1} \left(\expect{y K_x}\right).$$
The following lemma bounds the empirical regularized estimator with respect to the population one in terms of $\la, n$ and is essentially contained in the work of \citet{caponnetto2007optimal}; here we rederive it in a subcase (see below for the proof).
\blm\label{lm:krls-analytic-variance}
Under assumption \asm{asm:kernel-bounded},~\asm{asm:data-iid} for any $\la > 0$, note $u_n = \|\frac{1}{n} \sum_{i=1}^n y_i K_{x_i} - \expect{y K_x}\|_{\hh}$ and $v_n = \|\T - \Tn\|_{\textrm {op}}$, we have:
$$\|\widehat{g}_\la - g_\la\|_{\hh} \leq \frac{u_n}{\la} + \frac{R v_n}{\la^2}.$$
\elm
By using deviation inequalities for $u_n, v_n$ in Lemma~\ref{lm:krls-analytic-variance} and then applying Lemma~\ref{lm:appr-correct-sign-to-01}, we obtain the following exponential bound for kernel ridge regression (see complete proof below):
\bt\label{thm:exp-class-krls}
Under \asm{asm:separability},\asm{asm:kernel-bounded},\asm{asm:data-iid},\asm{asm:flambda-correct-sign} we have that for any $n \in \N$, 
$$\closs(\widehat{g}_\la) - \closs^* = 0\mbox{ with probability at least }1-4 
\exp\left(-\frac{C_0 \la^4\delta^2}{R^8} n\right).$$
Moreover,
$ \displaystyle \expect{\closs(\widehat{g}_\la) - \closs^*} \leq 4 
\exp\left(-C_0 \la^4\delta^2 n/ R^8\right), $
with $C_0^{-1} :=  72(1 + \la R^2)^2$.
\et  
The result above is a refinement of Thm.~2.6 from \citet{yao2007early}. We improved the dependency in $n$ and removed the requirements that $g^* \in \hh$ or $g^* = \Sigma^{r}w$ for a $w \in L^2(d\rhox)$ and $r > 1/2$. Similar results exist for losses that are usually considered more suitable for classification, like the hinge or logistic loss and more generally losses that are non-decreasing \citep[see][]{koltchinskii2005exponential}. With respect to this latter work, our analysis uses the explicit characterization of the kernel ridge regression estimator in terms of linear operators on $\hh$ \citep[see][]{caponnetto2007optimal}. This, together with \asm{asm:flambda-correct-sign}, allows us to use analytic tools specific to reproducing kernel Hilbert spaces, leading to proofs that are comparatively simpler, with explicit constants and a clearer problem setting (consisting essentially in \asm{asm:separability},~\asm{asm:flambda-correct-sign} and no assumptions on $\condexp{y}{x}$). 

Finally note that the exponent of $\la$ could be reduced by using a refined analysis under additional regularity assumption of $\rhox$ and $\condexp{y}{x}$ \citep[as {\em source condition} and {\em intrinsic dimension} from][]{caponnetto2007optimal}, but it is beyond the scope of this paper. 

\subsection{Proofs}

Here we prove that Kernel Ridge Regression achieves exponential classification rates under assumptions~\asm{asm:separability},~\asm{asm:flambda-correct-sign}. In particular by Lemma~\ref{lm:krls-analytic-variance} we bound $\nor{\widehat{g}_\la - g_\la}_\hh$ in high probability and then we use Lemma~\ref{lm:appr-correct-sign-to-01} that gives exponential classfication rates when $\nor{\widehat{g}_\la - g_\la}_\hh$ is small enough in high probability.

\bpr{\bfseries of Lemma~\ref{lm:krls-analytic-variance}}
Denote by $\Tnl$ the operator $\widehat{\Sigma} + \la I$ and with $\Tl$ the operator $\T + \la I$. We have
\eqals{
	\widehat{g}_\la - g_\la & = \Tnl^{-1} \left(\frac{1}{n} \sum_{i=1}^n y_i K_{x_i}\right) - \Tl^{-1}(\expect{y K_x})  \\
	& = \Tnl^{-1} \left(\frac{1}{n} \sum_{i=1}^n y_i K_{x_i} - \expect{y K_x}\right) ~+~ (\Tnl^{-1}-\Tl^{-1})\expect{yK_x}.
}
For the first term, since $\nor{\Tnl^{-1}}_{\textrm {op}} \leq \la^{-1}$, we have 
\eqals{
\nor{\Tnl^{-1} \left(\frac{1}{n} \sum_{i=1}^n y_i K_{x_i} - \expect{y K_x}\right)}_\hh &\leq \nor{\Tnl^{-1} }_{\textrm {op}} \nor{\frac{1}{n} \sum_{i=1}^n y_i K_{x_i} - \expect{y K_x}}_{\hh} \\
& \leq \frac{1}{\lambda} \nor{\frac{1}{n} \sum_{i=1}^n y_i K_{x_i} - \expect{y K_x}}_{\hh}.
}
For the second term, since $\|\Tl^{-1}\|_{\textrm {op}} \leq \la^{-1}$ and $\|\expect{yK_x}\| \leq \expect{\|yK_x\|} \leq R$, we have
\eqals{
\nor{(\Tnl^{-1} - \Tl^{-1})\expect{yK_x}}_\hh &= \nor{\Tnl^{-1}(\T - \Tn)\Tl^{-1}\expect{yK_x}}_\hh \\
& \leq  \nor{\Tnl^{-1}}_{\textrm {op}}\nor{\T - \Tn}_{\textrm {op}}\nor{\Tl^{-1}}_{\textrm {op}}\nor{\expect{yK_x}}_\hh \leq \frac{R}{\la^2} \nor{\T - \Tn}_{\textrm {op}}.
}
\epr

\bpr{\bfseries of Theorem~\ref{thm:exp-class-krls}}
Let $\tau > 0$. By Lemma~2 we know that
$$\|\widehat{g}_\la - g_\la\|_{\hh} \leq \frac{u_n}{\la} + \frac{R v_n}{\la^2},$$
with $u_n = \|\frac{1}{n} \sum_{i=1}^n (y_i K_{x_i} - \expect{y K_x})\|_{\hh}$ and $v_n = \|\T - \Tn\|_{\textrm {op}}$. 
For $u_n$ we can apply Pinelis inequality \citep[Thm.~3.5][]{pinelis1994optimum}, since $(x_i, y_i)_{i=1}^n$ are sampled independently according to the probability $\rho$ and that $y_i K_{x_i} - \expect{y K_{x}}$ is zero mean. Since $$\nor{\frac{1}{n}(y_i K_{x_i} - \expect{y K_{x}})}_\hh \leq \frac{2R}{n}$$ a.e. and $\hh$ is a Hilbert space, then we apply Pinelis inequality with $b^2_* = \frac{4R^2}{n}$ and $D = 1$, obtaining
$$u_n \leq \sqrt{\frac{8R^2 \tau}{n}},$$
with probability at least $1 - 2e^{-\tau}$.
Now, denote by $\nor{\cdot}_{HS}$ the Hilbert-Schmidt norm and recall that $\nor{\cdot} \leq \nor{\cdot}_{HS}$.  To bound $v_n$ we apply again the Pinelis inequality \citep[see also][]{rosasco2010learning} considering that the space of Hilbert-Schmidt operators is again a Hilbert space and that $\Tn = \frac{1}{n} \sum_{i=1}^n K_{x_i} \otimes K_{x_i}$, that $(x_i)_{i=1}^n$ are independently sampled from $\rhox$ and that $\expect{K_{x_i} \otimes K_{x_i}} = \Sigma$. In particular we apply it with $D = 1$ and $b^2_* = \frac{4R^4}{n}$, so
$$v_n = \nor{\T - \Tn} \leq \nor{\T - \Tn}_{HS} \leq \sqrt{\frac{8 R^4 \tau}{n}},$$
with probability $1-2e^{-\tau}$.
Finally we take the intersection bound of the two events obtaining, with probability at least $1-4e^{-\tau}$,
$$\|\widehat{g}_\la - g_\la\|_{\hh} \leq \sqrt{\frac{8R^2 \tau}{\la^2 n}} + \sqrt{\frac{8R^6\tau}{\la^4 n}}.$$
By selecting $\tau = \frac{\delta^2}{9R^2(\sqrt{\frac{8R^2}{\la^2 n}} + \sqrt{\frac{8R^6}{\la^4 n}})^2}$, we obtain $\|\widehat{g}_\la - g_\la\|_{\hh} \leq \frac{\delta}{3R}$, with probability $1-4e^{-\tau}$.
Now we can apply Lemma~\ref{lm:appr-correct-sign-to-01} to have the exponential bound for the classification error.
\epr

\section{Proofs and additional results about concrete examples}
\label{sect:examples-for-glambda}

In the next subsection we prove that $g_\ast \in \hh$ is sufficient to satisfy \asm{asm:flambda-correct-sign}, while in subsection \ref{sect:A5-examples} we prove that specific settings naturally satisfy \asm{asm:flambda-correct-sign}.

\subsection{From \texorpdfstring{$g_\ast \in \hh$}{gstar} to \asm{asm:flambda-correct-sign}}\label{sect:from-g-in-H-to-A5}

Here we assume that there exists $g_\ast \in \hh$ such that $g_\ast(x) = \condexp{y}{x}$ a.e. on the support of $\rhox$. 
First we introduce $A(\la)$, that is a quantity related to the approximation error of $g_\la$ with respect to $g_\ast$ and we study its behavior when $\la \to 0$. Then we express $\nor{g_\la - g_\ast}_\hh$ in terms of $A(\la)$. Finally we prove that for any $\delta$ given by \asm{asm:separability}, there exists $\la$ such that \asm{asm:flambda-correct-sign} is satisfied.

Let $(\sigma_t, u_t)_{t \in \N}$ be an eigenbasis of $\T$ with $\sigma_1 \geq \sigma_2 \geq \dots \geq 0$, and let $\alpha_j = \scal{g_\ast}{u_j}$ we introduce the following quantity 
$$A(\la) = \sum_{t : \sigma_t \leq \la} \alpha_t^2.$$

\blm\label{lm:Ala-go-to-0}
Under \asm{asm:kernel-bounded}, $A(\la)$ is decreasing for any $\la > 0$ and
$$\lim_{\la \to 0} A(\la) = 0.$$
\elm
\bpr
Under \asm{asm:kernel-bounded} and the linearity of trace, we have that
$$\sum_{j \in N} \sigma_j = \tr(\T) = \int \tr\left(K_x \otimes K_x\right) d\rhox(x) = \int \scal{K_x}{K_x}_\hh d\rhox(x) = \int K(x,x) d\rhox(x) \leq R^2.$$
Denote by $t_\la \in \N$, the number $\min \{t \in \N ~|~ \sigma_t \leq \la \}$. Since the $(\sigma_j)_{j \in \N}$ is a non-decreasing summable sequence, then it converges to $0$, then
$$\lim_{\la \to 0}~ t_\la = \infty.$$
Finally, since $(\alpha_j^2)_{j \in \N}$ is a summable sequence
we have that 
$$ \lim_{\la \to 0} A(\la) = \lim_{\la \to 0} \sum_{t: \sigma_t \leq \la} \alpha_t^2 = \lim_{\la \to 0} \sum_{j = t_\la} \alpha_j^2 = \lim_{t \to \infty} \sum_{j = t}^\infty \alpha_j^2 = 0.$$
\epr

Here we express $\nor{g_\la - g_\ast}_{\hh}$ in terms of $\nor{g_\ast}_\hh$ and of $A(\sqrt{\la})$.
\blm\label{lm:krls-analytic-bias}
Under \asm{asm:kernel-bounded}, for any $\la > 0$ we have
$$\nor{g_\la - g_\ast}_{\hh} \leq \sqrt{\sqrt{\la}\nor{g_\ast}_\hh^2 + {A}(\sqrt{\la})}.$$
\elm
\bpr
Denote by $\Tl$ the operator $\T + \la I$.  Note that since $g_\ast \in \hh$, then 
$$\expect{yK_x} = \expect{g_\ast(x) K_x} = \expect{(K_x \otimes K_x)g_\ast} = \expect{K_x \otimes K_x}g_\ast = \Sigma g_\ast,$$
then $g_\la = \Tl^{-1}\expect{yK_x} = \Tl^{-1}\T g_\ast$. So we have
\eqals{
	\nor{g_\la - g_\ast}_\hh & = \nor{\Tl^{-1}\T g_\ast - g_\ast}_\hh = \nor{(\Tl^{-1}\T - I)g_\ast}_\hh = \la \nor{\Tl^{-1} g_\ast}_\hh.
}
Moreover
$$
\la \nor{(\T + \la I)^{-1} g_\ast}_\hh \leq \sqrt{\la} \nor{(\T + \la I)^{-1/2}}~\sqrt{\la} \nor{(\T + \la I)^{-1/2} g_\ast}_\hh \leq \sqrt{\la} \nor{(\T + \la I)^{-1/2} g_\ast}_\hh.$$
Now we express $\sqrt{\la}\nor{(\T + \la I)^{-1/2} g_\ast}_\hh$ in terms of $A(\la)$. We have that
\eqals{
	\la\nor{(\T + \la I)^{-1/2} g_\ast}_\hh^2 = \la \scal{g_\ast}{(\T + \la I)^{-1} g_\ast} 
	&= \la \scal{g_\ast}{\left(\sum_{j \in \N}(\sigma_j + \la)^{-1} u_j \otimes u_j\right)g_\ast}  \\
	&= \sum_{j\in \N} \frac{\la \alpha_j^2}{\sigma_j + \la}.
}
Now divide the series in two parts
$$\sum_{j\in \N} \frac{\la \alpha_j^2}{\sigma_j + \la} = S_1(\la) + S_2(\la), \quad S_1(\la) = \sum_{j: \sigma_j \geq \sqrt{\la}} \frac{\la \alpha_j^2}{\sigma_j + \la}, ~~ S_2(\la) = \sum_{j: \sigma_j < \sqrt{\la}} \frac{\la \alpha_j^2}{\sigma_j + \la}.$$
For each term in $S_1$, since $j$ is selected such that $\sigma_j \geq \sqrt{\la}$ we have that 
$\la(\sigma_j + \la)^{-1} \leq \la(\sqrt{\la} + \la)^{-1} \leq \la/\sqrt{\la} \leq \sqrt{\la},$
so
$$S_1(\la) \leq \sqrt{\la} \sum_{j: \sigma_j \geq \sqrt{\la}} \alpha_j^2 \leq \sqrt{\la} \sum_{j \in \N} \alpha_j^2 = \sqrt{\la} \nor{g_\ast}^2.$$
For $S_2$, we have that $\la(\sigma_j + \la)^{-1} \leq 1$, so
$$S_2(\la) \leq \sum_{j: \sigma_j < \sqrt{\la}}  \alpha_j^2 = A(\sqrt{\la}).$$
\epr

\bpr{\bfseries of Proposition~\ref{prop:gstar-in-hh-gives-gla-good}}
By Lemma~\ref{lm:krls-analytic-bias} we have that 
$$\nor{g_\la - g_\ast}_{\hh} \leq \sqrt{\sqrt{\la}\nor{g_\ast}_\hh^2 + {A}(\sqrt{\la})}.$$
Now note that the r.h.s. is non-decreasing in $\la$, and is $0$ when $\la \to 0$, due to Lemma~\ref{lm:Ala-go-to-0}. Then there exists $\la$ such that $\nor{g_\la - g_\ast}_{\hh} < \frac{\delta}{2R}$.

Since $|f(x)| \leq R\nor{f}_\hh$ for any $f \in \hh$ when the kernel satisfies \asm{asm:kernel-bounded} and moreover \asm{asm:separability} holds, we have that for any $x \in \X$ such that $g_\ast(x) > 0$ we have
\eqals{
g_\la(x) = g_\ast(x) - (g_\ast(x) - g_\la(x)) \geq g_\ast(x) - |g_\ast(x) - g_\la(x)| \geq \delta - R\nor{g_\la - g_\ast} \geq \delta/2,
}
so $\sign(g_\ast(x)) = \sign(g_\la(x)) = +1$ and $\sign(g_\ast(x))g_\la(x) \geq \delta/2$. Analogously for any $x \in \X$ such that $g_\ast(x) < 0$ we have
\eqals{
	g_\la(x) = g_\ast(x) + (g_\la(x) - g_\ast(x)) \leq g_\ast(x) + |g_\ast(x) - g_\la(x)| \leq -\delta + R\nor{g_\la - g_\ast} \leq -\delta/2,
}
so $\sign(g_\ast(x)) = \sign(g_\la(x)) = -1$ and $\sign(g_\ast(x))g_\la(x) \geq \delta/2$. Note finally that $g_\ast(x) = 0$ on a zero measure set by \asm{asm:flambda-correct-sign}.
\epr

\subsection{Examples}\label{sect:A5-examples}

In this subsection we first introduce some notation and basic results about Sobolev spaces, then we prove Prop.~\ref{prop:2g-makes-gstar} and Example~\ref{ex:independent-noise-on-labels}.

In what follows denote by $A_{t}$ the $t$-fattening of a set $A \subseteq \R^d$, that is $A_t = \bigcup_{x \in P} B_t(x)$ where $B_t(x)$ is the open ball of ray $t$ centered in $x$. 
We denote by $W^{s,2}(\R^d)$ the Sobolev space endowed with norm 
$$\nor{f}_{W^{s,2}} = \left\{f \in \L^1(\R^d) \cap \L^2(\R^d) ~\middle|~ \int_{\R^d} {\cal F}(f)(\omega)^2 (1 + \|\omega\|^2)^{s/2} d\omega < \infty \right\}.$$
Finally we define the function $\phi_{s,t}:\X\to\R$, that will be used in the proofs as follows
$$\phi_{s, t}(x) ~~=~~ q_{d,\delta} ~t^{-d} ~ 1_{\{0\}_t}(x) ~ (1 - \|x/t\|^2)^{s - d/2},$$
with $q_{d,s} = \pi^{-d/2}\Gamma(1+s)/\Gamma(1+s-d/2)$ and $t > 0, s \geq d/2$.
Note that $\phi_{s, t}(x)$ is supported on $\{0\}_{\epsilon/2}$, satisfies
$$ \int_{\R^d} \phi_{s, t}(y) dy = 1$$
and it is continuous and belongs to $W^{s,2}(\R^d)$.

\bp\label{prop:exists-gPN}
Let $P, N$ two compact subsets of $\R^d$ with Hausdorff distance at least $\epsilon > 0$. There exists $g_{P, N} \in W^{s,2}$ such that 
$$ g_{P,N}(x) = 1,~~ \forall ~ x \in P,\qquad q_{P,N}(x) = 0,~~ \forall ~ x \in N.$$
In particular $g_{P, N} = 1_{P_{\epsilon/2}} * \phi_{s,\epsilon/2}$.
\ep
\bpr
Denote by $v_{\epsilon,s}$ the function $(1 - \|2x/\epsilon\|^2)^{s - d/2}$. We have
\eqals{
	g_{P,N}(x) &~~=~~ q_{d,s} (\epsilon/2)^{-d} \int_{\R^d} ~1_{P_{\epsilon/2}}(x-y) ~1_{\{0\}_{\epsilon/2}}(y) ~v_{\epsilon,s}(y) ~ d y \\
	&~~=~~ q_{d,s} (\epsilon/2)^{-d} \int_{\{0\}_{\epsilon/2}} ~1_{P_{\epsilon/2}}(x-y) ~ v_{\epsilon,s}(y) ~ d y \\
	&~~=~~ q_{d,s} (\epsilon/2)^{-d} \int_{\{x\}_{\epsilon/2}} ~1_{P_{\epsilon/2}}(y) ~ v_{\epsilon,s}(y-x) ~ d y 
}
Now when $x \in P$, then $\{x\}_{\epsilon/2} \subseteq P_{\epsilon/2}$, so 
\eqals{
	g_{P,N}(x) 	&~~=~~ q_{d,s} (\epsilon/2)^{-d} \int_{\{x\}_{\epsilon/2}} ~1_{P_{\epsilon/2}}(y) ~ v_{\epsilon,s}(y-x) ~ d y  \\
	&~~=~~ q_{d,s} (\epsilon/2)^{-d} \int_{\{x\}_{\epsilon/2}} v_{\epsilon,s}(y-x) d y ~~=~~ q_{d,s} \epsilon^{-d} \int_{\{0\}_{\epsilon/2}}v_{\epsilon,s}(y) d y \\
	&~~=~~ q_{d,s} (\epsilon/2)^{-d} \int_{\R^d} 1_{\{0\}_{\epsilon/2}}(y)v_{\epsilon,s}(y) d y ~~ = ~~ \int_{\R^d} \phi_{s,\epsilon/2}(y) dy ~~ = ~~ 1.
}
Conversely, when $x \in N$, then $\{x\}_{\epsilon/2} \cap P_{\epsilon/2} = \emptyset$, so
\eqals{
	g_{P,N}(x) 	&~~=~~ q_{d,s} (\epsilon/2)^{-d} \int_{\{x\}_{\epsilon/2}} ~1_{P_{\epsilon/2}}(y) ~ v_{\epsilon,s}(y-x) ~ d y  ~~ = ~~ 0.
}
Now we prove that $g_{P,N} \in W^{s,2}$. First note that $P_{\epsilon/2}$ is compact whenever $P$ is compact. This implies that $1_{P_{\epsilon/2}}$ is in $L^2(\R^d)$.
Since $g_\delta$ is the convolution of an $L^2(\R^d)$ function and a $W^{s,2}$, then it belongs to $W^{s,2}$.
\epr 

\begin{figure}[t]
  \centering
  \includegraphics[width=0.5\textwidth]{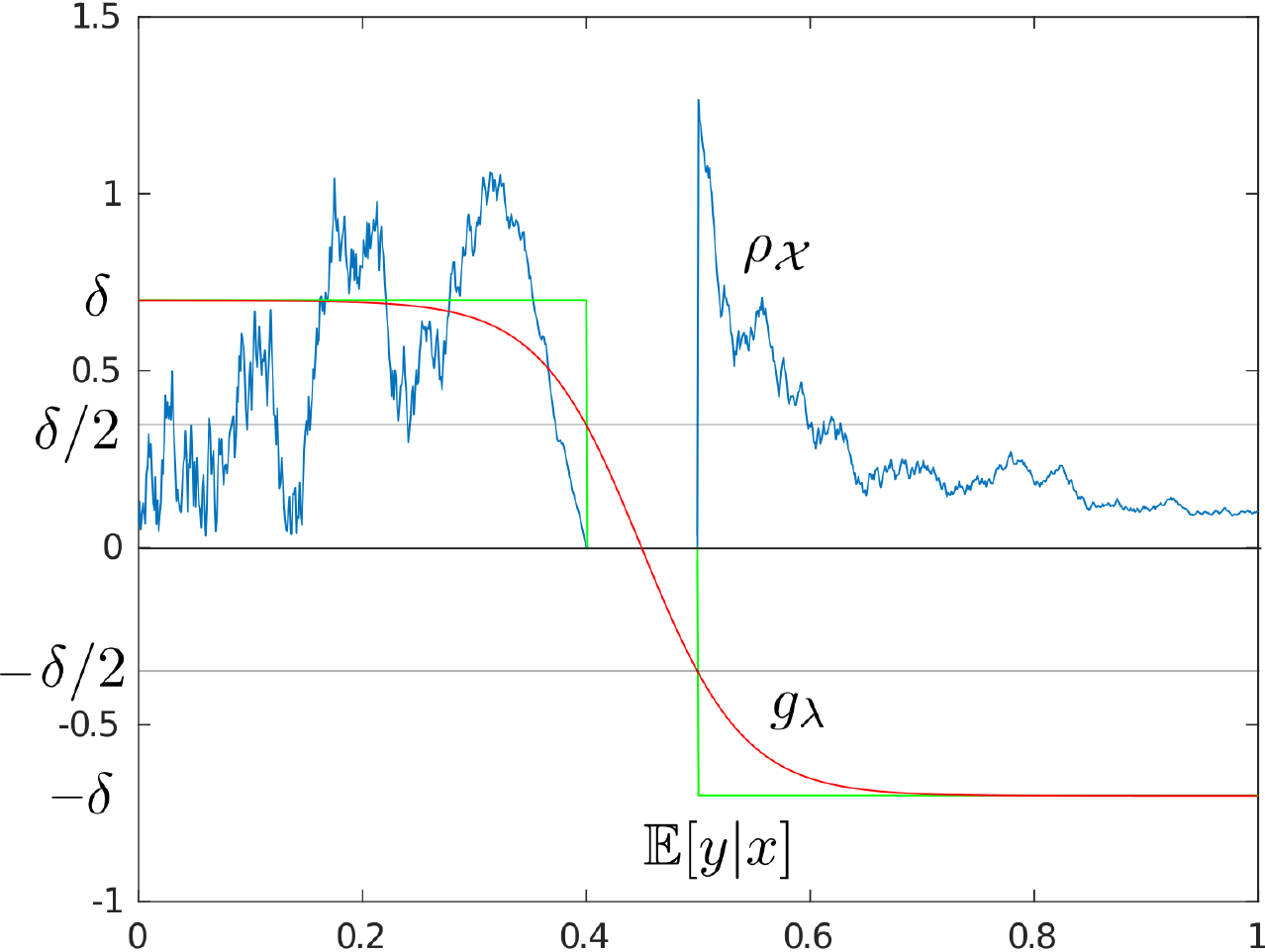}
  \caption{Pictorial representation of a model in $1$D satisfying Example~\ref{ex:independent-noise-on-labels}, ($p = 0.15$). Blue: $\rhox$, green: $\condexp{y}{x}$, red: $g_\la$.}
  \label{fig:example-A5}
  %
\end{figure}

\bpr{\bfseries of Proposition~\ref{prop:2g-makes-gstar}}
Since we are under \asm{asm:margin}, we can apply Prop.~\ref{prop:exists-gPN} that prove the existence two functions $q_{\mathcal{S}_+, \mathcal{S}_-}, q_{\mathcal{S}_-, \mathcal{S}_+} \in W^{s,2}$ with the property to be respectively equal to $1$ on $\mathcal{S}_+$, $0$ on $\mathcal{S}_-$, and $1$ on $\mathcal{S}_-$, $0$ on $\mathcal{S}_+$. 
Since $W^{s,2}$ is a Banach algebra \citep[see][]{adams2003sobolev}, then $g h \in W^{s,2}$ for any $g, h \in W^{s,2}$. So in particular 
$$g_\ast ~=~ g_+^*  q_{\mathcal{S}_+,\mathcal{S}_-} ~-~ g_-^*  q_{\mathcal{S}_-, \mathcal{S}_+},$$
belongs to $W^{s,2}$ (and so to $\hh$) and is equal to $\condexp{y}{x}$ a.e. on the support of $\rhox$ by definition. Finally, \asm{asm:flambda-correct-sign} is satisfied, by Prop.~\ref{prop:gstar-in-hh-gives-gla-good}.
\epr

\bpr{\bfseries of Example~\ref{ex:independent-noise-on-labels}}
By definition of $y$, we have that
$$\condexp{y}{x} = (1-2p)g(x), \quad g(x) = {\mathbf 1}_{\mathcal{S}_+} - {\mathbf 1}_{\mathcal{S}_-}.$$
In particular note that \asm{asm:separability} is satisfied with $\delta = 1-2p > 0$ since $p \in [0,1/2)$. Moreover note that  $\condexp{y}{x}$ is constant $\delta$ on $\mathcal{S}_+$ and $-\delta$ on $\mathcal{S}_-$. Note now that there exists two functions in $W^{s,2} \subseteq \hh$ (due to \asm{asm:kernel-rich}) that are, respectively $\delta$ on $\mathcal{S}_+$ and $-\delta$ on $\mathcal{S}_-$. They are exactly $g^*_+ := \delta q_{\mathcal{S}_+, \mathcal{S}_-}$ and $g^\ast_- = -\delta q_{\mathcal{S}_-, \mathcal{S}_+}$, from Prop.~\ref{prop:exists-gPN}. So we can apply Prop.~\ref{prop:2g-makes-gstar}, that given $g^\ast_+, g^*_-$ guarantees that \asm{asm:flambda-correct-sign} is satisfied. See an example in Figure \ref{fig:example-A5}.
\epr

\section{Preliminaries for Stochastic Gradient Descent}
\label{ap:SGDdevelopment}

In this section we show two preliminary results on stochastic gradient descent.

\subsection{Proof of the optimality condition on \texorpdfstring{$g_*$}{g*}}
\label{ap:optimality}

In this subsection we prove the optimality condition on $g_*$: $$\E \left[ \left(y_n - \Phg(x_n)\right)K_{x_n}\right] = 0.$$ Let us recall that as $\H$ is not necessarily dense in $L_2$, we have defined $\Phg$ as the orthonormal projector for the $L_2$ norm on $\H$ of $g_* = \E(y|x)$ which is the minimizer over all $g \in L_2$ of $\E (y - g(x) )^2$. Let $\F$ be the linear space $ \bar{\H}^{L_2}$ equipped with the $L_2$ norm, remark that $\Phg$ verifies $\Phg = \underset{g \in \F}{\text{argmin}} \| g-g_*  \|^2_{L_2}$ and that $g_* - \Phg = \mathcal{P}_{\H^\perp} (g_*) \in \F^\perp$.  

\begin{align*}
\E \left[ \left(y_n - \Phg(x_n)\right)K_{x_n}\right] &= \E \left[ \left(y_n - \E(y_n|x_n) + \E(y_n|x_n) - \Phg(x_n)\right)K_{x_n}\right] \\
&= \E \left[ \left(y_n - \E(y_n|x_n)\right)K_{x_n}\right] + \E \left[ \left(g_*(x_n) - \Phg(x_n)\right)K_{x_n}\right] \\
&= \E \left[\mathcal{P}_{\H^\perp} (g_*)(x_n)K_{x_n}\right] \\
&=0,
\end{align*}
where the last equality is true because we have $<\mathcal{P}_{\H^\perp} (g_*) , K(\cdot,z) >_{L_2} = 0$ and,
\begin{align*}
\|\E \left[\mathcal{P}_{\H^\perp} (g_*)(x_n)K_{x_n}\right]\|^2_\H &= \left\| \int_x \mathcal{P}_{\H^\perp} (g_*)(x)K_x d \rho(x) \right\|^2_\H  \\
&= \int_z \mathcal{P}_{\H^\perp} (g_*)(z) \left(\underbrace{\int_x \mathcal{P}_{\H^\perp} (g_*)(x)  K(x,z)  d \rho(x)}_{ = 0 }\right) d \rho(z) = 0.
\end{align*}

\subsection{Proof of Lemma~\ref{le:SGDrecursion}: reformulation of SGD as noisy recursion}
\label{ap:SGDreformulation}

Let $n \geqslant 1$ and $g_0 \in \H$, we start form the SGD recursion defined by \eqref{eq:firstSGD}:
\begin{align*}
{g}_n 
 & = {g}_{n-1} - \gamma_n  \big[ ( \langle K_{x_n} ,  {g}_{n-1}\rangle - y_n)K_{x_n}   + \lambda ({g}_{n-1} - g_0) \big] \\
 & = {g}_{n-1} - \gamma_n \big[ K_{x_n} \otimes K_{x_n} {g}_{n-1} - y_n K_{x_n}  + \lambda ({g}_{n-1} - g_0) \big] \\
 & = {g}_{n-1} - \gamma_n \big[ K_{x_n} \otimes K_{x_n} {g}_{n-1} - \Phg(x_n) K_{x_n} - \xi_n K_{x_n} + \lambda ({g}_{n-1} - g_0) \big] ,
 \end{align*}
 leading to (using the optimality conditions for $g_\lambda$ and $g_\ast$):
 \begin{align*}
{g}_n - g_\lambda 
  & =  {g}_{n-1} - g_\lambda - \gamma_n \big[ K_{x_n} \otimes K_{x_n} ( {g}_{n-1} - g_\lambda) + \lambda ({g}_{n-1} - g_0)  \\
   &   \hspace*{1.77cm} + (K_{x_n} \otimes K_{x_n}) g_\lambda - \Phg(x_n) K_{x_n} \big] + \gamma_n \xi_n K_{x_n} \\
 & = {g}_{n-1} - g_\lambda
 - \gamma_n \big[ K_{x_n} \otimes K_{x_n} ( {g}_{n-1} - g_\lambda)  + \lambda ({g}_{n-1} - g_0) \\
 &   \hspace*{1.77cm} 
 + (K_{x_n} \otimes K_{x_n} - \Sigma) g_\lambda + \Sigma g_\lambda  - \Phg(x_n) K_{x_n} \big] + \gamma_n \xi_n K_{x_n}\\
  & =  {g}_{n-1} - g_\lambda
 - \gamma_n \big[ K_{x_n} \otimes K_{x_n} ( {g}_{n-1} - g_\lambda)  + \lambda {g}_{n-1} 
 + (K_{x_n} \otimes K_{x_n} - \Sigma) g_\lambda \\
&   \hspace*{1.77cm} -\lambda g_\lambda + \E \left[ \Phg(x_n) K_{x_n} \right] - \Phg(x_n) K_{x_n} \big] + \gamma_n \xi_n K_{x_n}\\
 & =  {g}_{n-1} - g_\lambda
 - \gamma_n \big[ (  K_{x_n} \otimes K_{x_n} + \lambda I)( {g}_{n-1} - g_\lambda)   + (K_{x_n} \otimes K_{x_n} - \Sigma) g_\lambda \\
 &  \hspace*{1.77cm} +   \E \left[ \Phg(x_n) K_{x_n} \right] - \Phg(x_n) K_{x_n}   \big] + \gamma_n \xi_n K_{x_n}\\
& =  \big[
I - \gamma_n   (  K_{x_n} \otimes K_{x_n} + \lambda I ) \big]
  ( {g}_{n-1} - g_\lambda ) \\
&    \hspace*{1.77cm}   + \gamma_n \left[  \xi_n K_{x_n} +  ( \Sigma- K_{x_n} \otimes K_{x_n}) g_\lambda +   \Phg(x_n) K_{x_n} - \E \left[ \Phg(x_n) K_{x_n} \right] \right] \\ 
  & =  \big[
I - \gamma_n   (  K_{x_n} \otimes K_{x_n} + \lambda I ) \big]
  ( {g}_{n-1} - g_\lambda )\\
&    \hspace*{1.77cm}     + \gamma_n \left[  \xi_n K_{x_n}  - (K_{x_n} \otimes K_{x_n}) g_\lambda +   \Phg(x_n) K_{x_n} + \Sigma g_\lambda - \E \left[ \Phg(x_n) K_{x_n} \right] \right] \\ 
    & =  \big[
I - \gamma_n   (  K_{x_n} \otimes K_{x_n} + \lambda I ) \big]
  ( {g}_{n-1} - g_\lambda )  \\
&    \hspace*{1.77cm}   + \gamma_n \left[  \xi_n K_{x_n}   +   (\Phg(x_n)- g_\lambda(x_n)) K_{x_n} - \E \left[ (\Phg(x_n)- g_\lambda(x_n)) K_{x_n} \right] \right].
\end{align*}

\section{Proof of stochastic gradient descent results} \label{sec:AppSGD}

Let us recall for the Appendix the SGD recursion defined in Eq.~\eqref{eq:SGDabstract}: 
\begin{equation*}
   \eta_n = ( \idm - \gamma H_n) \eta_{n-1} + \gamma_n \varepsilon_n,
\end{equation*}
for which we assume \sgdasm{asm:init}, \sgdasm{asm:noise-iid}, \sgdasm{asm:noise-bound},\sgdasm{asm:weird-bound}, \sgdasm{asm:commute}.

\paragraph{Notations.}

We define the following notations, which will be useful during all the proofs of the section: 

\BIT
\item the following contractant operators:  for $i \geqslant k $, $$M(i,k) = (\idm - \gamma H_i ) \cdots ( \idm - \gamma H_k), \text{ and } M(i,i+1) = \idm, $$ 
\item the following sequences $Z_k = M(n,k+1) \varepsilon_k$ and $W_n = \sum_{k=1}^n \gamma_k  Z_k$. 
\EIT

then,
\begin{equation}
 \eta_n =
M(n,n) \eta_{n-1} + \gamma_n \varepsilon_n 
\end{equation}
\begin{equation}
\label{eq:decomposition}
 \eta_n = 
M(n,1)  \eta_0 + \sum_{k=1}^n \gamma_k M(n,k+1) \varepsilon_k,
\end{equation}

Note that in all this section, when there is no ambiguity, we will use $\|\cdot\|$ instead of $\|\cdot\|_\H$. 

\subsection{Non-averaged SGD - Proof of Theorem \ref{th:SGDalpha}}
\label{ap:SGDalpha}

In this section, we define the two following sequences: $\displaystyle \alpha_n = \prod_{i=1}^n ( 1-\gamma_i \lambda)$, \\ $\displaystyle \beta_n = \sum_{k=1}^n \gamma_k^2 \prod_{i=k+1}^n (1-\gamma_i \lambda)^2$ and $\zeta_n = \displaystyle \sup_{ k \leqslant n}\gamma_k \prod_{i=k+1}^n \left(1-\gamma_i\lambda\right)$.   

We can decompose $\eta_n$ in two terms: 
\begin{eqnarray}
 \eta_n = \underbrace{M(n,1) \eta_0}_{\textbf{Biais term}} + \underbrace{W_n}_{\textbf{Noise term}},
\end{eqnarray}
\BIT
\item The biais term represents the speed at which we forget initial conditions. It is the product of $n$ contracting operators $$\|  M(n,1)  \eta_0 \| \leqslant \prod_{i=1}^n ( 1-\gamma_i \lambda) \| \eta_0\| = \alpha_n \|\eta_0\|.$$
\item The noise term $W_n$ which is a martingale. We are going to show by using a concentration inequality that the probability of the event $\{ \| W_n \| \geq t\}$ goes to zero exponentially fast.
\EIT

\subsubsection{General result for all \texorpdfstring{$(\gamma_n)$}{gamma}}

As $W_n = \sum_{k=1}^n \gamma_k Z_k$, we want to apply Corollary \ref{Probabilisticcor} of section \ref{sec:proba} to $(\gamma_kZ_k)_{k \in \mathbb{N}}$ that is why we need the following lemma:

\begin{lemma}
\label{le:boundsalpha}
We have the following bounds: 
\begin{eqnarray}
\sup_{ k \leqslant n} \| \gamma_k Z_k \| &\leqslant& c^{1/2} \zeta_n, \ and \\
\sum_{k=1}^n \mathbb{E}\left[ \|\gamma_k Z_k\|^2 | \mathcal{F}_{k-1} \right] &\leqslant& \tr C \beta_n,
\end{eqnarray}
where $c$ and $C$ are defined by \sgdasm{asm:noise-bound}.
\end{lemma}

\begin{proof}
First, $ \|\gamma_k Z_k \| = \gamma_k \left\| M(n,k+1) \varepsilon_k \right\| \leq \gamma_k\left\| M(n,k+1) \right\|_{\text{op}} \left\| \varepsilon_k \right\| \leq \gamma_k\displaystyle \frac{\alpha_n}{\alpha_k} \left\| \varepsilon_k \right\| \leqslant \zeta_n c^{1/2} $. 

Second, 
\begin{eqnarray*}
\sum_{k=1}^n \mathbb{E}\left[ \|\gamma_k Z_k\|^2 | \mathcal{F}_{k-1} \right] 
&\leqslant& \sum_{k=1}^n \displaystyle \frac{\alpha_n^2}{\alpha_k^2}\ \gamma_k^2\  \E \left\| \varepsilon_k  \right\|^2 \\
&\leqslant& \sum_{k=1}^n \displaystyle \frac{\alpha_n^2}{\alpha_k^2}\ \gamma_k^2\  \tr C.
\end{eqnarray*}

Hence,
\begin{eqnarray*}
\sum_{k=1}^n \mathbb{E}\left[ \|\gamma_k Z_k\|^2 | \mathcal{F}_{k-1} \right]  &\leqslant& \sum_{k=1}^n \gamma_k^2 \prod_{i=k+1}^n (1-\gamma_i \lambda)^2 \tr C 
\\
&=&  \tr C \beta_n.      
\end{eqnarray*}
\end{proof}
\begin{proposition}
\label{prop:alphabetazeta}
We have the following inequality: for $t > 0, n \geqslant 1$,
\begin{eqnarray}
 \|\eta_n \| &\leqslant& \alpha_n \|\eta_0\| + V_n, \quad \text{with}\\
\P \left( V_n \geqslant t \right) &\leqslant& 2\exp \left( - \frac{t^2}{2(\tr C \beta_n + c^{1/2} \zeta_n t / 3)}  \right).
\end{eqnarray}
\end{proposition}
\begin{proof}
We just need to apply Lemma \ref{le:boundsalpha} and Corollary \ref{Probabilisticcor} to the martingale $W_n$ and $V_n = \|W_n\|$ for all $n$.
\end{proof}

\subsubsection{Result for \texorpdfstring{$\gamma_n = \gamma/n^\alpha$}{gamma} }

We now derive estimates of $\alpha_n, \beta_n$ and $\zeta_n$ to have explicit bound for the previous result in the case where $\gamma_n = \displaystyle \frac{\gamma}{n^\alpha}$ for $\alpha \in [0,1]$. Some of the estimations are taken from \citet{gradsto}.

\begin{lemma}
\label{le:nalpha}
In the interesting particular case where $\gamma_n = \displaystyle \frac{\gamma}{n^\alpha}$ for $\alpha \in [0,1]$:
\BIT
\item for $\alpha = 1,\ i.e \ \gamma_n=\displaystyle \frac{\gamma}{n}$, then $\zeta_n = \displaystyle \frac{\gamma}{1-\gamma\lambda}\alpha_n$, and we have the following estimations for $\gamma\lambda < 1/2$: \\(i)\ ~$\alpha_n \leqslant \displaystyle \frac{1}{n^{\gamma\lambda}}$, (ii)\ ~$\beta_n \leqslant \displaystyle \frac{2(1-\gamma\lambda)}{1-2\gamma\lambda} \frac{4^{\gamma\lambda}\gamma^2}{n^{2\gamma\lambda}}$, (iii)\ ~$ \zeta_n \leqslant \displaystyle \frac{\gamma}{(1-\lambda\gamma)n^{\gamma\lambda}}$.
\item for $\alpha = 0,\ i.e \ \gamma_n=\displaystyle \gamma$, then $\zeta_n = \gamma$, and we have the following: \\(i)\ ~$\alpha_n = (1-\gamma\lambda)^n$, (ii)\ ~$\beta_n \leqslant \displaystyle \frac{\gamma}{\lambda}$, (iii)\ ~$ \zeta_n = \gamma$.
\item for $\alpha \in \left]0,1\right[,\ \zeta_n=\max\left\{ \gamma_n, \displaystyle \frac{\gamma}{1-\gamma\lambda}\alpha_n \right\}$, and we have the following estimations: 

(i)\ ~$\alpha_n \leqslant \displaystyle \exp\left(  -\frac{\gamma\lambda}{1-\alpha}\left( (n+1)^{1-\alpha} -1  \right)   \right)$, 

(ii) Denoting $ L_\alpha = \frac{2\lambda\gamma}{1-\alpha} 2^{1-\alpha}\left( 1-\left(\frac{3}{4}\right)^{1-\alpha} \right)$, we distinguish three cases:  
\BIT
\item $\alpha >1/2$, \ ~$\beta_n \leqslant \gamma^2\frac{2\alpha}{2\alpha-1} \exp\left( -L_\alpha n^{1-\alpha}\right) +  \frac{2^\alpha\gamma}{\lambda n^\alpha}$,
\item $\alpha = 1/2$, \ ~$\beta_n \leqslant \gamma^2\ln (3n) \exp\left( -L_\alpha n^{1-\alpha}\right) +  \frac{2^\alpha\gamma}{\lambda n^\alpha}$,
\item $\alpha  < 1/2$, \ ~$\beta_n \leqslant \gamma^2\frac{n^{1-2\alpha}}{1- 2\alpha} \exp\left( -L_\alpha n^{1-\alpha}\right) +  \frac{2^\alpha\gamma}{\lambda n^\alpha}$.
\EIT
(iii)\ ~$ \zeta_n \leqslant \max \left\{\frac{\gamma}{1-\gamma\lambda}\exp\left(  -\frac{\gamma\lambda}{1-\alpha}\left( (n+1)^{1-\alpha} -1  \right)   \right), \frac{\gamma}{n^\alpha} \right\}.$
\EIT

Note that in this case for $n$ large enough we have the following estimations: 

(i)\ ~$\alpha_n \leqslant \displaystyle \exp\left(  -\frac{\gamma\lambda}{2^{1-\alpha}(1-\alpha)}n^{1-\alpha}    \right)$, (ii)\ ~$\beta_n \leqslant \displaystyle \frac{2^{\alpha+1}\gamma}{\lambda n^\alpha}$, (iii)\ ~$ \zeta_n \leqslant \displaystyle \frac{\gamma}{n^{\alpha}}$.
\end{lemma}

\begin{proof}
First we show for $\alpha \in [0,1]$ the equality for $\zeta_n$.
Denote $a_k = \gamma_k \prod_{i=k+1}^n (1-\gamma_i\lambda)$, we want to find $\zeta_n = \sup_{k \leqslant n} a_k$. We show for $\gamma_n = \displaystyle \frac{\gamma}{n^\alpha}$ that $(a_k)_{k\geqslant 1 }$ decreases then increases so that $\zeta_n = \max\{ a_1 , a_n \}$.
Let $k \leqslant n-1$, 
\begin{eqnarray*}
\frac{a_{k+1}}{a_k}&=&\frac{\gamma_{k+1}}{\gamma_k}\frac{1}{(1-\gamma_{k+1}\lambda)}\\
&=&\frac{1}{\frac{\gamma_{k}}{\gamma_{k+1}}-\gamma_{k}\lambda}\\
\end{eqnarray*}
Hence, $\displaystyle \frac{a_{k}}{a_{k+1}} - 1 = \frac{\gamma_{k}}{\gamma_{k+1}}-\gamma_{k}\lambda-1$. Take $\alpha \in \left]0,1\right[$, in this case where $\gamma_n = \displaystyle \frac{\gamma}{n^\alpha}$, 

$$ \frac{a_{k}}{a_{k+1}} - 1 = \left(1+\frac{1}{k}\right)^\alpha - \frac{\gamma\lambda}{k^\alpha}-1 .$$

A rapid study of the function $\displaystyle f_\alpha(x) = \left(1+\frac{1}{x}\right)^\alpha - \frac{\gamma\lambda}{x^\alpha}-1$ in $\mathbb{R}_+^\star$ shows that it decreases until $x_\star = \left( \gamma \lambda\right)^{\frac{1}{(\alpha-1)}}-1$ then increases. This concludes the proof for $\alpha \in \left]0,1\right[$. By a direct calculation for $\alpha = 1$, $\displaystyle \frac{a_{k}}{a_{k+1}} - 1 = \frac{1-\gamma\lambda}{k} \geqslant 0$ thus $a_k$ is non increasing and $\zeta_n =a_1 = \displaystyle \frac{\gamma}{1-\gamma\lambda}\alpha_n$. Similarly, for $\alpha=0$, $\displaystyle \frac{a_{k}}{a_{k+1}} - 1 = \gamma\lambda < 0$ thus $a_k$ is increasing and $\zeta_n =a_n = \gamma_n$.

We show now the different estimations we have for $\alpha_n$, $\beta_n$ and $\zeta_n$ for the three cases above.
\BIT
\item for $\alpha = 1$, 
\begin{eqnarray*}
\ln \alpha_n &=& \sum_{i=1}^n \ln \left( 1 - \frac{\gamma \lambda}{i}\right) \leqslant -\gamma \lambda \sum_{i=1}^n \frac{1}{i} \leqslant -\gamma\lambda \ln n \\
\alpha_n &\leqslant& \frac{1}{n^{\gamma \lambda}}.
\end{eqnarray*}
Then, 
\begin{eqnarray*}
\beta_n &=& \gamma^2 \sum_{k=1}^n \frac{1}{k^2}\prod_{i=k+1}^n \left( 1-\frac{\gamma\lambda}{i} \right)^2 \\
\beta_n &\leqslant& \gamma^2 \sum_{k=1}^n \frac{1}{k^2}\exp\left( -2 \gamma\lambda \sum_{i=k+1}^n \frac{1}{i}  \right) \\
&\leqslant& \gamma^2 \sum_{k=1}^n \frac{1}{k^2}\exp\left( -2 \gamma\lambda \ln\left( \frac{n+1}{k+1} \right)  \right) \\
&\leqslant& \gamma^2 \sum_{k=1}^n \frac{1}{k^2}\left( \frac{k+1}{n+1} \right)^{2\gamma\lambda}  \\
&\leqslant& 4^{\gamma\lambda}\gamma^2 \sum_{k=1}^n \frac{1}{k^2}\left( \frac{k}{n} \right)^{2\gamma\lambda}  \\
&\leqslant& \frac{4^{\gamma\lambda}\gamma^2}{n^{2\gamma\lambda}} \sum_{k=1}^n k^{2\gamma\lambda-2},
\end{eqnarray*}
Moreover for $\displaystyle \gamma\lambda < \frac{1}{2},\  \sum_{k=1}^n k^{2\gamma\lambda-2} \leqslant 1 - \frac{1}{2\gamma\lambda-1} = \frac{2(1-\gamma\lambda)}{1-2\gamma\lambda}$, hence,

\begin{eqnarray*}
\beta_n &\leqslant&  \frac{2(1-\gamma\lambda)}{1-2\gamma\lambda} \frac{4^{\gamma\lambda}\gamma^2}{n^{2\gamma\lambda}}
\end{eqnarray*}

Finally,
\begin{eqnarray*}
\zeta_n = \frac{\gamma}{1-\gamma\lambda} \alpha_n \leqslant \frac{\gamma}{1-\gamma\lambda} \frac{1}{n^{\gamma\lambda}}.
\end{eqnarray*}

\item for $\alpha = 0$, 
\begin{eqnarray*}
\alpha_n = \prod_{i=1}^n \left( 1 - \gamma \lambda\right) = (1-\gamma\lambda)^n.
\end{eqnarray*}
Then, 
\begin{eqnarray*}
\beta_n = \gamma^2 \sum_{k=1}^n \prod_{i=k+1}^n \left( 1-\gamma\lambda \right)^2 = \gamma^2 \sum_{k=1}^n \left( 1- \gamma\lambda \right)^{2(n-k)} \leqslant \frac{1}{1-(1-\lambda\gamma)^2} \leqslant \frac{\gamma}{\lambda}.
\end{eqnarray*}
Finally,
\begin{eqnarray*}
\zeta_n = \gamma_n = \gamma.
\end{eqnarray*}

\item for $\alpha \in ]0,1[$, 
\begin{eqnarray*}
\ln \alpha_n &=& \sum_{i=1}^n \ln \left( 1 - \frac{\gamma \lambda}{i^\alpha}\right) \leqslant -\gamma \lambda \sum_{i=1}^n \frac{1}{i^\alpha} \leqslant -\gamma\lambda \frac{(n+1)^{1-\alpha}-1}{1-\alpha} \\
\alpha_n &\leqslant& \exp\left( -\frac{\gamma\lambda}{1-\alpha} \left((n+1)^{1-\alpha}-1\right) \right).
\end{eqnarray*}
To have an estimation on $\beta_n$, we are going to split it into two sums. Let $m \in \llbracket 1,n \rrbracket$,
\begin{align*}
\beta_n &= \sum_{k=1}^n \gamma_k^2\prod_{i=k+1}^n \left( 1-\gamma_i\lambda \right)^2 = \sum_{k=1}^{m} \gamma_k^2\prod_{i=k+1}^n \left( 1-\gamma_i\lambda \right)^2 +  \sum_{k=m+1}^n \gamma_k^2\prod_{i=k+1}^n \left( 1-\gamma_i\lambda \right)^2\\
\beta_n &\leqslant \sum_{k=1}^m  \gamma_k^2 \exp\left( -2 \lambda \sum_{i=m+1}^n \gamma_i  \right) +  \frac{\gamma_m}{\lambda}\sum_{k=m+1}^n \prod_{i=k+1}^n \left( 1-\gamma_i\lambda \right)^2 \lambda\gamma_k\\
 &\leqslant \sum_{k=1}^n  \gamma_k^2 \exp\left( -2 \lambda \sum_{i=m+1}^n \gamma_i  \right) \\
 & \hspace{2.98cm}+  \frac{\gamma_m}{\lambda}\sum_{k=m+1}^n \left[ \prod_{i=k+1}^n \left( 1-\gamma_i\lambda \right)^2 - \prod_{i=k+1}^n \left( 1-\gamma_i\lambda \right)^2 (1-\gamma_k\lambda)  \right]\\
 &\leqslant \sum_{k=1}^n  \gamma_k^2 \exp\left( -2 \lambda \sum_{i=m+1}^n \gamma_i  \right) +  \frac{\gamma_m}{\lambda}\sum_{k=m+1}^n \left[ \prod_{i=k+1}^n \left( 1-\gamma_i\lambda \right)^2 - \prod_{i=k}^n \left( 1-\gamma_i\lambda \right)^2 \right]\\
 &\leqslant \sum_{k=1}^n  \gamma_k^2 \exp\left( -2 \lambda \sum_{i=m+1}^n \gamma_i  \right) +  \frac{\gamma_m}{\lambda} \left(1- \prod_{i=m+1}^n \left( 1-\gamma_i\lambda \right)^2\right) \\
 &\leqslant \sum_{k=1}^n  \gamma_k^2 \exp\left( -2 \lambda \sum_{i=m+1}^n \gamma_i  \right) +  \frac{\gamma_m}{\lambda}.
\end{align*}
By taking $\gamma_n = \displaystyle \frac{\gamma}{n^\alpha}$ and $ m = \displaystyle \lfloor\frac{n}{2}\rfloor $, we get: 
\begin{eqnarray*}
\beta_n &\leqslant& \gamma^2\sum_{k=1}^n  \frac{1}{k^{2\alpha}} \exp\left( -2 \lambda\gamma \sum_{i=\lfloor\frac{n}{2}\rfloor+1}^n \frac{1}{i^{\alpha}}  \right) +  \frac{2^\alpha\gamma}{\lambda n^\alpha} \\
 &\leqslant& \gamma^2\sum_{k=1}^n  \frac{1}{k^{2\alpha}} \exp\left( -\frac{2 \lambda\gamma}{1-\alpha} \left( \left(n+1\right)^{1-\alpha}-\left(\frac{n}{2}+1\right)^{1-\alpha} \right) \right) +  \frac{2^\alpha\gamma}{\lambda n^\alpha} \\
  &\leqslant& \gamma^2\sum_{k=1}^n  \frac{1}{k^{2\alpha}} \exp\left( -\frac{2 \lambda\gamma}{1-\alpha} n^{1-\alpha} \left( \left(1+\frac{1}{n}\right)^{1-\alpha}-\left(\frac{1}{2}+\frac{1}{n}\right)^{1-\alpha} \right) \right) +  \frac{2^\alpha\gamma}{\lambda n^\alpha} \\
  &\leqslant& \gamma^2\sum_{k=1}^n  \frac{1}{k^{2\alpha}} \exp\left( -\frac{2 \lambda\gamma}{1-\alpha} n^{1-\alpha} 2^{1-\alpha}\left( 1-\left(\frac{3}{4}\right)^{1-\alpha} \right)\right) +  \frac{2^\alpha\gamma}{\lambda n^\alpha} .
\end{eqnarray*}
Calling $S_n^\alpha = \sum_{k=1}^n \frac{1}{k^{2\alpha}}$ and noting that: for $\alpha > 1/2,\ S_n^\alpha \leqslant \frac{2\alpha}{2\alpha - 1}$, $\alpha = 1/2,\ S_n^\alpha \leqslant \ln(3n)$ and $\alpha < 1/2,\ S_n^\alpha \leqslant \frac{n^{1-2\alpha}}{1 - 2\alpha }$ we have the expected result.

Finally,
\begin{eqnarray*}
\zeta_n \leqslant \max \left\{\frac{\gamma}{1-\gamma\lambda}\exp\left(  -\frac{\gamma\lambda}{1-\alpha}\left( (n+1)^{1-\alpha} -1  \right)   \right), \frac{\gamma}{n^\alpha} \right\}.
\end{eqnarray*}
\EIT

\end{proof}

With this estimations we can easily show the Theorem \ref{th:SGDalpha}. In the following we recall the main result of this Theorem and give an extension for $\alpha = 0$ and $\alpha = 1$ that cannot be found in the main text.

\begin{proposition}[SGD, decreasing step size: $\gamma_n = \gamma/n^\alpha$]
\label{prop:fullalpha}
Assume \sgdasm{asm:init}, \sgdasm{asm:noise-iid}, \sgdasm{asm:noise-bound}, $\gamma_n = \gamma/n^\alpha$, $\gamma\lambda < 1$ and denote by $\eta_n \in \H$ the n-th iterate of the recursion in Eq. \eqref{eq:SGDabstract}. We have for $t > 0, n \geqslant 1$,
\BIT
\item for $\alpha = 1$ and $\gamma\lambda < 1/2$, $\displaystyle \ \|g_n - g_\lambda\|_\H \leqslant \frac{\|g_0 - g_\lambda\|_\H}{n^{\gamma\lambda}}  +V_n,$ almost surely, with
 \begin{align*}
\P \left(V_n  \geqslant t\right) \leqslant 2\exp\left(-\frac{t^2}{4^{3/2}(\tr C) \gamma^2/((1-2\gamma\lambda) n ^{\gamma\lambda}) +4tc^{1/2}\gamma/3  }\cdot n^{\gamma\lambda}\right);
\end{align*}
\item for $\alpha = 0$, $\displaystyle \ \|g_n - g_\lambda\|_\H \leqslant (1-\gamma\lambda)^n \|g_0 - g_\lambda\|_\H + V_n$, almost surely, with
$$\P \left( V_n \geqslant t \right) \leqslant 
2\exp\left( -\frac{t^2}{2\gamma(\tr C  / \lambda + t c^{1/2} / 3 )}\right);$$

\item for $\alpha \in (0,1)$, $\|g_n - g_\lambda\|_\H \leqslant \exp\left(  -\frac{\gamma\lambda}{1-\alpha}\left( (n+1)^{1-\alpha} -1  \right)   \right) \|g_0 - g_\lambda\|_\H + V_n$, almost surely for $n$ large enough \footnote {See Appendix Section \ref{sec:AppSGD} Lemma \ref{le:nalpha} for more details.}, with
$$\P \left( V_n \geqslant t \right) \leqslant 2\exp\left( -\frac{ t^2 }{ \gamma (2 ^{\alpha + 2} \tr C/\lambda  + 2 c^{1/2}t  / 3)}\cdot n^{\alpha}\right).$$
\EIT
\end{proposition}

\begin{proof}[Proof of Theorem~\ref{th:SGDalpha}]
We apply Proposition \ref{prop:alphabetazeta}, and the bound found on $\alpha_n$, $\beta_n$ and $\zeta_n$ in Lemma \ref{le:nalpha} to get the results.
\end{proof}

\subsection{Averaged SGD for the variance term \texorpdfstring{($\eta_0 = 0$)}{} - Proof of Theorem \ref{th:SGDaveraged}}
\label{ap:SGDaverage}

We consider the same recursion but with $\gamma_n = \gamma$:
$$
\eta_n = ( \idm - \gamma H_n) \eta_{n-1} + \gamma \varepsilon_n,
$$
started at  $\eta_0=0$ and with assumptions \sgdasm{asm:init}, \sgdasm{asm:noise-iid}, \sgdasm{asm:noise-bound},\sgdasm{asm:weird-bound}, \sgdasm{asm:commute}.

However, in this section, we consider the averaged: 
$$
\bar{\eta}_n = \frac{1}{n+1} \sum_{i=0}^n \eta_i.
$$

Thus, we get
$$
\bar{\eta}_n = \frac{1}{n+1} \sum_{i=0}^n \gamma \sum_{k=1}^i M(i,k+1) \varepsilon_k
= \frac{\gamma}{n+1} 
 \sum_{k=1}^n \Big( \sum_{i=k}^n M(i,k+1) \Big) \varepsilon_k
 = \frac{\gamma}{n+1}  \sum_{k=1}^n \bar{Z}_k.
 $$

Our the goal is to bound $\displaystyle\P \left( \left\| \bar{\eta}_n  \right\| \geqslant t \right) $ using Propostion \ref{Probabilisticprop} that is going to lead us to some Bernstein concentration inquality. Calling, as above, $\displaystyle \bar{Z}_k =  \sum_{i=k}^n M(i,k+1) \varepsilon_k$, and as $\E \left[ \bar{Z}_k | \mathcal{F}_{k-1} \right] = 0$ we just need to bound, $\sup_{k \leqslant n} \| \bar{Z}_k \|$ and $\sum_{k = 1}^n \E \left[ \|\bar{Z}_k\|^2 | \mathcal{F}_{k-1} \right] $. For a more general result, we consider in the following lemma $\displaystyle (A^{1/2}\bar{Z}_k)_k$.

\begin{lemma}
\label{le:averagebounds}
Assuming \sgdasm{asm:init}, \sgdasm{asm:noise-iid}, \sgdasm{asm:noise-bound},\sgdasm{asm:weird-bound}, \sgdasm{asm:commute}, we have the following bounds for $\displaystyle \bar{Z}_k =  \sum_{i=k}^n M(i,k+1) \varepsilon_k$: 
\begin{align}
\sup_{k \leqslant n} \| A^{1/2} \bar{Z}_k \| &\leqslant \frac{c^{1/2} \|A\|_{op}^{1/2} }{\gamma \lambda} \\
\sum_{k = 1}^n \E \left[ \|A^{1/2} \bar{Z}_k\|^2 | \mathcal{F}_{k-1} \right] &\leqslant n \frac{1}{\gamma^2}\frac{1}{1-\gamma/2\gamma_0} \tr \left(A H^{-2}\cdot C \right).
\end{align}

\end{lemma}

\begin{proof}
First $\| A^{1/2} \bar{Z}_k \| \leqslant \|A\|_{op}^{1/2} \| \bar{Z}_k \|$ and we have, almost surely, $\| \varepsilon_k  \| \leqslant c^{1/2} $ and $ H_n \succcurlyeq \lambda \idm $, thus for all $k$, as $\gamma\lambda \leqslant 1$, $\idm - \gamma H_k \preccurlyeq (1-\gamma\lambda) \idm$. Hence,  $\| M(i,k+1) \|_{\textrm {op}} \leqslant ( 1- \gamma \lambda)^{i-k}$ and,
$$ \| \bar{Z}_k \| \leqslant \|\varepsilon_k\| \sum_{i=k}^n \| M(i,k+1) \|_{\textrm {op}}
 \leqslant
c^{1/2} \sum_{i=k}^n ( 1- \gamma \lambda)^{i-k} 
 \leqslant
\frac{c^{1/2} }{\gamma \lambda}
 $$
Second, we need an upper bound on $\displaystyle \E \left[ \|A^{1/2}\bar{Z}_k\|^2 | \mathcal{F}_{k-1} \right]$, we are going to find it in two steps: 
\BIT
\item \textbf{Step 1:} we first show that the upper bound depends of the trace of some operator involving~$H^{-1}$. \begin{eqnarray*}
\E \left[ \|A^{1/2} \bar{Z}_k\|^2 | \mathcal{F}_{k-1} \right]
& \leqslant & 2 \sum_{i=k}^n \tr\left( A \left(\gamma H\right)^{-1}\E \left[    M(i,k+1) C {M(i,k+1)}^*\right] \right),
\end{eqnarray*} 
\item \textbf{Step 2:} we then upperbound this sum to a telescopic one involving $H^{-2}$ to finally show: \begin{eqnarray*}
\E \left[ \|A^{1/2} \bar{Z}_k\|^2 | \mathcal{F}_{k-1} \right]
& \leqslant & \frac{1}{\gamma^2}\frac{1}{1-\gamma/2\gamma_0} \tr \left(AH^{-2} C \right).
\end{eqnarray*} 
\EIT

\paragraph{Step 1:} We write,

\begin{eqnarray*}
\E \left[ \|A^{1/2} \bar{Z}_k\|^2 | \mathcal{F}_{k-1} \right] & =& \E \left[ \sum_{k\leqslant i, j \leqslant n} \left\langle A^{1/2} M(i,k+1)\varepsilon_k, A^{1/2} M(j,k+1)\varepsilon_k\right\rangle | \mathcal{F}_{k-1} \right] \\
& =& \E \left[ \sum_{k\leqslant i, j \leqslant n} \left\langle M(i,k+1)\varepsilon_k, A M(j,k+1)\varepsilon_k\right\rangle | \mathcal{F}_{k-1} \right] \\
& =& \sum_{k\leqslant i, j \leqslant n} \E \left[ \tr \left( M(i,k+1)^* A M(j,k+1) \cdot \varepsilon_k\otimes \varepsilon_k\right) \right] \\
& =& \sum_{k\leqslant i, j \leqslant n} \tr\left( \E \left[  M(i,k+1)^* A M(j,k+1)\right] \cdot \E \left[ \varepsilon_k\otimes \varepsilon_k \right]\right).
\end{eqnarray*}
We have $\E \left[ \varepsilon_k\otimes \varepsilon_k \right] \preccurlyeq C $ so that as every operators are positive semi-definite, 
\begin{eqnarray*}
\E \left[ \|A^{1/2} \bar{Z}_k\|^2 | \mathcal{F}_{k-1} \right] & \leqslant & \sum_{k\leqslant i, j \leqslant n} \tr\left( \E \left[  M(i,k+1)^* A M(j,k+1)\right] \cdot C\right).
\end{eqnarray*}
We now bound the last expression by dividing it into two terms, noting $M(i,k) = M^i_k$ for more compact notations (only until the end of the proof),
\begin{align*}
\sum_{k\leqslant i, j \leqslant n} \tr\left( \E \left[  {M^i_{k+1}}^* A M^j_{k+1}\right] \cdot C\right) & = \sum_{i=k}^n \tr\left( \E \left[  {M^i_{k+1}}^* A M^i_{k+1}\right] \cdot C\right) \\
& \hspace{2cm}+ 2 \sum_{k\leqslant i < j \leqslant n} \tr\left( \E \left[  {M^i_{k+1}}^* A M^j_{k+1}\right] \cdot C\right).
\end{align*}
Moreover, 
\begin{eqnarray*}
 & &  \sum_{k\leqslant i < j \leqslant n} \tr\left( \E \left[  {M^i_{k+1}}^* A M^j_{k+1}\right] \cdot C\right) 
 \\ & = &  \sum_{k\leqslant i < j \leqslant n} \tr\left( \E \left[  {M^i_{k+1}}^* A \left( \idm - \gamma H \right)^{j-i} M^i_{k+1}\right] \cdot C\right) \\
& = &  \sum_{i=k}^n \tr\left( \E \left[  {M^i_{k+1}}^* A \sum_{j=i+1}^n \left( \idm - \gamma H \right)^{j-i} M^i_{k+1}\right] \cdot C\right) \\
& = &  \sum_{i=k}^n \tr\left( \E \left[  {M^i_{k+1}}^* A \left[ \left( \idm - \gamma H \right) \left( \idm - \left( \idm - \gamma H \right)^{n-i} \right) \left(\gamma H \right)^{-1} \right] M^i_{k+1}\right] \cdot C \right) \\
& \leqslant  &  \sum_{i=k}^n \tr\left( \E \left[  {M^i_{k+1}}^* A \left[ \left(\gamma H\right)^{-1} - \idm \right] M^i_{k+1}\right] \cdot C \right)\\
& \leqslant  &  \sum_{i=k}^n \tr\left( \E \left[  {M^i_{k+1}}^* A \left(\gamma H\right)^{-1} M^i_{k+1}\right] \cdot C \right)-\sum_{i=k}^n \tr\left( \E \left[  {M^i_{k+1}}^* A M^i_{k+1}\right] \cdot C \right).
\end{eqnarray*}
Hence, 

\eqals{
\sum_{k\leqslant i, j \leqslant n} \tr & \left( \E \left[  {M^i_{k+1}}^* A M^j_{k+1}\right] \cdot C\right)  \\
& = \sum_{i=k}^n \tr\left( \E \left[  {M^i_{k+1}}^* A M^i_{k+1}\right] \cdot C\right)
+ 2 \sum_{k\leqslant i < j \leqslant n} \tr\left( \E \left[  {M^i_{k+1}}^* A M^j_{k+1}\right] \cdot C\right) \\
& \leqslant  2 \sum_{i=k}^n \tr\left( \E \left[  {M^i_{k+1}}^* A \left(\gamma H\right)^{-1} M^i_{k+1}\right] \cdot C \right)- \sum_{i=k}^n \tr\left( \E \left[  {M^i_{k+1}}^* A M^i_{k+1}\right] \cdot C \right) \\
& \leqslant 2 \sum_{i=k}^n \tr\left( \E \left[  {M^i_{k+1}}^* A \left(\gamma H\right)^{-1} M^i_{k+1}\right] \cdot C \right)\\
& \leqslant 2 \sum_{i=k}^n \tr\left(A\left(\gamma H\right)^{-1}  \E \left[   M^i_{k+1} C {M^i_{k+1}}^*  \right] \right)
}

This concludes step 1.

\paragraph{Step 2: } Let us now try to bound $\displaystyle \sum_{i=k}^n \tr\left( A \left(\gamma H\right)^{-1}\E \left[    M^i_{k+1} C {M^i_{k+1}}^*\right] \right)$. We will do so by bounding it by a telescopic sum. Indeed,
\eqals{
&\E \left[  {M^{i+1}_{k+1}}C\left(\gamma H\right)^{-1}{M^{i+1}_{k+1}}^*\right]  = \E \left[  {M^{i}_{k+1}}\left(\idm - \gamma H_{i+1}\right)C\left(\gamma H\right)^{-1}\left(\idm - \gamma H_{i+1}\right){M^{i}_{k+1}}^*\right] \\
& = \E \left[  {M^{i}_{k+1}}\E \left[ C\left(\gamma H\right)^{-1} - C H^{-1} H_{i+1}- H_{i+1} C H^{-1} + \gamma H_{i+1}C H^{-1}H_{i+1} \right]{M^{i}_{k+1}}^*\right] \\
& = \E \left[  {M^{i}_{k+1}}C\left(\gamma H\right)^{-1}{M^{i}_{k+1}}^*\right]-2\E \left[  {M^{i}_{k+1}}C{M^{i}_{k+1}}^*\right]  + \gamma\E \left[  {M^{i}_{k+1}}\E \left[ H_{i+1}CH^{-1}H_{i+1} \right]{M^{i}_{k+1}}^*\right],
}
such that, by multiplying the previous equality by $A \left(\gamma H\right)^{-1}$ and taking the trace we have, 
\begin{align*}
\tr\left( A \left(\gamma H\right)^{-1}\E \left[  {M^{i+1}_{k+1}}C\left(\gamma H\right)^{-1}{M^{i+1}_{k+1}}^*\right]\right) &=  \tr\left(A \left(\gamma H\right)^{-1}\E \left[  {M^{i}_{k+1}}C\left(\gamma H\right)^{-1}{M^{i}_{k+1}}^*\right]\right)\\
 &-2\tr\left(A \left(\gamma H\right)^{-1}\E \left[  {M^{i}_{k+1}}C{M^{i}_{k+1}}^*\right]\right) \\
&+ \gamma \tr\left(A \left(\gamma H\right)^{-1}\E \left[  {M^{i}_{k+1}}\E \left[ H_{i+1}CH^{-1}H_{i+1} \right]{M^{i}_{k+1}}^*\right]\right),
\end{align*}
And as $\E \Big[  H_k C H^{-1} H_k \Big] \preccurlyeq \gamma_0^{-1} C$ we have, 
$$\gamma\tr\left(A \left(\gamma H\right)^{-1} \E \left[  {M^{i}_{k+1}}\E \left[ H_{i+1}CH^{-1}H_{i+1} \right]{M^{i}_{k+1}}^*\right]\right) \leqslant \gamma / \gamma_0 \tr\left(A \left(\gamma H\right)^{-1} \E \left[  {M^{i}_{k+1}}C{M^{i}_{k+1}}^*\right]\right), $$
thus,
\begin{eqnarray*}
\tr\left(A \left(\gamma H\right)^{-1} \E \left[ {M^{i+1}_{k+1}}C\left(\gamma H\right)^{-1}{M^{i+1}_{k+1}}^*\right]\right) &\leqslant&  \tr\left(A \left(\gamma H\right)^{-1}\E \left[  {M^{i}_{k+1}}C\left(\gamma H\right)^{-1}{M^{i}_{k+1}}^*\right]\right) \\
&-&2\tr\left(A \left(\gamma H\right)^{-1}\E \left[  {M^{i}_{k+1}}C{M^{i}_{k+1}}^*\right]\right) \\
& + & \gamma / \gamma_0 \tr\left(A \left(\gamma H\right)^{-1} \E \left[  {M^{i}_{k+1}}C{M^{i}_{k+1}}^*\right]\right)
\end{eqnarray*}
\begin{align*}
&\tr\left(A \left(\gamma H\right)^{-1} \E \left[  {M^{i}_{k+1}}C{M^{i}_{k+1}}^*\right]\right)\\ 
 &\leqslant \frac{1}{2-\frac{\gamma}{\gamma_0}} \left(\tr\left(A \left(\gamma H\right)^{-1}\E \left[  {M^{i}_{k+1}}C\left(\gamma H\right)^{-1}{M^{i}_{k+1}}^*\right]\right) - \tr\left(A \left(\gamma H\right)^{-1} \E \left[ {M^{i+1}_{k+1}}C\left(\gamma H\right)^{-1}{M^{i+1}_{k+1}}^*\right]\right)\right).
\end{align*}
If we take all the calculations from the beginning,
\begin{eqnarray*}
\E \left[ \|A^{1/2} \bar{Z}_k\|^2 | \mathcal{F}_{k-1} \right] & \leqslant & \sum_{k\leqslant i, j \leqslant n} \tr\left( \E \left[  {M^i_{k+1}}^* A M^j_{k+1}\right] \cdot C\right) \\
& \leqslant & 2 \sum_{i=k}^n \tr\left(A \left(\gamma H\right)^{-1} \E \left[  {M^{i}_{k+1}}C{M^{i}_{k+1}}^*\right]\right)\\
& \leqslant & \frac{2}{2-\gamma/\gamma_0} \sum_{i=k}^n \tr\left(A \left(\gamma H\right)^{-1}\E \left[  {M^{i}_{k+1}}C\left(\gamma H\right)^{-1}{M^{i}_{k+1}}^*\right]\right) \\
& &   \hspace*{1.77cm} -\tr\left(A \left(\gamma H\right)^{-1} \E \left[ {M^{i+1}_{k+1}}C\left(\gamma H\right)^{-1}{M^{i+1}_{k+1}}^*\right]\right)\\
& \leqslant & \frac{2}{2-\gamma/\gamma_0} \tr\left(A \left(\gamma H\right)^{-1}\E \left[  {M^{k}_{k+1}}C\left(\gamma H\right)^{-1}{M^{k}_{k+1}}^*\right]\right)\\
& \leqslant & \frac{1}{\gamma^2}\frac{1}{1-\gamma/2\gamma_0} \tr \left(A H^{-2}\cdot C \right),
\end{eqnarray*}
which concludes the proof if we sum this inequality from $1$ to $n$.

\end{proof}

We can now prove Theorem~\ref{th:SGDaveraged}:

\begin{proof}[Proof of Theorem~\ref{th:SGDaveraged}]
We apply Corollary \ref{Probabilisticcor} to the sequence $\left(\displaystyle\frac{\gamma}{n+1} A^{1/2}Z_k\right)_{k \leqslant n}$ thanks to Lemma \ref{le:averagebounds}. We have:
\begin{eqnarray*}
\sup_{k \leqslant n} \| \frac{\gamma}{n+1} A^{1/2}Z_k \| &\leqslant& \frac{c^{1/2} \|A^{1/2}\| }{(n+1) \lambda} \\
\sum_{k = 1}^n \E \left[ \|\frac{\gamma}{n+1} A^{1/2} Z_k\|^2 | \mathcal{F}_{k-1} \right] &\leqslant& \frac{1}{n+1}\frac{1}{1-\gamma/2\gamma_0} \tr \left(AH^{-2}\cdot C \right),
\end{eqnarray*}
so that, 
\begin{align*}
\P \left( \left\| A^{1/2}\bar{\eta}_n  \right\| \geqslant t \right) &= \P \left( \left\| \sum_{k = 1}^n \displaystyle\frac{\gamma}{n+1}A^{1/2}Z_k  \right\| \geqslant t \right) \leqslant 2 \exp\left(-\frac{t^2}{2\left( \frac{\tr \left(AH^{-2}\cdot C \right)}{(n+1)(1-\gamma/2\gamma_0)}   + \frac{c^{1/2} \|A^{1/2}\| t}{3 \lambda(n+1)}  \right)}\right)\\ 
\P \left( \left\| A^{1/2}\bar{\eta}_n  \right\| \geqslant t \right) &\leqslant 2 \exp\left(-\frac{(n+1)t^2}{ \frac{2\tr \left(AH^{-2}\cdot C \right)}{(1-\gamma/2\gamma_0)}   + \frac{2 \|A^{1/2}\|c^{1/2} t }{3 \lambda } }\right).
\end{align*}
\end{proof}

\subsection{Tail-averaged SGD - Proof of Corollary~\ref{co:SGDtailaveraged}}
\label{ap:SGDcorrolary}

We now prove the result for tail-averaging that allow us to relax the assumption that $\eta_0 = 0$. The proof relies on the fact that the bias term can easily be bounded as $\| \bar{\eta}_n^{\textrm {tail, bias}} \|_\H \leqslant ( 1- \lambda \gamma )^{n/2} \|\eta_0\|_\H$. For the variance term, we can simply use the Theorem \ref{th:SGDaveraged} for $n$ and $n/2$, as 
$\bar{\eta}_n^{\textrm {tail}}  = 2 \bar{\eta}_n  - \bar{\eta}_{n/2}$.

\begin{proof} [Proof of Corollary~\ref{co:SGDtailaveraged}] 

Let $n \geqslant 1$ and $n$ an even number for the sake of clarity (the case where $n$ is an odd number can be solved similarly), 
\begin{eqnarray*}
A^{1/2}\bar{\eta}_n^{\textrm {tail}} &=& \frac{1}{ n / 2} \sum_{k= n/2}^{n}A^{1/2} \eta_k \\
&=& \frac{1}{n/2 } \sum_{k= n/2 }^{n} A^{1/2}M(k,1)\eta_0 +\frac{1}{ n/2 } \sum_{k= n/2 }^{n} A^{1/2}W_k \\
&=& \frac{1}{ n/2 } \sum_{k= n/2 }^{n}A^{1/2}M(k,1)\eta_0 +2 A^{1/2}\overline{W}_{n} -  A^{1/2}\overline{W}_{ n/2 }.
\end{eqnarray*}
Hence,
\begin{eqnarray*}
\left\| A^{1/2}\bar{\eta}_n^{\textrm {tail}} \right\| &\leqslant & \left\| \frac{1}{n/2} \sum_{k= n/2 }^{n} A^{1/2} M(k,1)\eta_0 \right\|+2 \left\| A^{1/2}\overline{W}_{n} \right\|+ \left\| A^{1/2} \overline{W}_{ n/2 } \right\|\\
&\leqslant & \frac{1}{ n/2 } \sum_{k= n/2 }^{n} \left\| A^{1/2}M(k,1)\right\|_{op} \left\| \eta_0 \right\|+2 \left\| A^{1/2}\overline{W}_{n} \right\|+ \left\| A^{1/2} \overline{W}_{ n/2 } \right\|,
\end{eqnarray*}
Let $L_n = 2 \left\| A^{1/2}\overline{W}_{n} \right\|+ \left\|A^{1/2}  \overline{W}_{ n/2} \right\|$,
\begin{eqnarray*}
\left\| A^{1/2}\bar{\eta}_n^{\textrm {tail}} \right\| &\leqslant & \frac{1}{ n/2 } \sum_{k= n/2 }^{n} \|A^{1/2}\|_{op}(1-\gamma\lambda)^{k} \left\| \eta_0 \right\|+L_n\\
\left\| A^{1/2} \bar{\eta}_n^{\textrm {tail}} \right\| &\leqslant &  (1-\gamma\lambda)^{n/2} \|A^{1/2}\|_{op} \left\| \eta_0 \right\|+L_n,
\end{eqnarray*}
And finally for $t \geqslant 0$,
\begin{eqnarray*}
\P(L_n \geqslant t ) &=& \P(2 \left\| A^{1/2}\overline{W}_{n} \right\|+ \left\| A^{1/2} \overline{W}_{n/2} \right\| \geqslant t ) \\
&\leqslant& \P\left(2 \left\| A^{1/2}\overline{W}_{n} \right\| \geqslant t\right)+\P\left( \left\| A^{1/2} \overline{W}_{n/2} \right\| \geqslant t \right) \\
&\leqslant& 2\left[\exp\left(-\frac{(n+1) (t/2)^2}{E_{t/2}} \right) + \exp\left(-\frac{(n/2+1)t^2}{E_t} \right)\right] .
\end{eqnarray*}
Let us remark that $ E_{t/2} \leqslant E_t $. Hence,
\begin{eqnarray*}
\P(L_n \geqslant t ) &\leqslant& 2\left[\exp\left(-\frac{(n+1) t^2}{4E_t} \right) + \exp\left(-\frac{(n+1) t^2}{2E_t} \right)\right] \\
&\leqslant& 4 \exp\left(-\frac{(n+1) t^2}{4E_t} \right). 
\end{eqnarray*}

\end{proof}

\section{Exponentially convergent SGD for classification error}\label{sec:error}

In this section we prove the results for the error in the case of SGD. Let us recall the recursion:
\begin{equation*}
    {g}_n - g_\lambda =  \big[
I - \gamma_n   (  K_{x_n} \otimes K_{x_n} + \lambda I ) \big]
  ( {g}_{n-1} - g_\lambda )   + \gamma_n \varepsilon_n,
\end{equation*}
with the noise term $\varepsilon_k  =   \xi_k K_{x_k}   +   (\Phg(x_k)- g_\lambda(x_k)) K_{x_k} - \E \left[ (\Phg(x_k)- g_\lambda(x_k)) K_{x_k} \right] \in \H.$
This is the same recursion as in Eq~\eqref{eq:SGDabstract}:

\begin{equation*}
\eta_n = ( \idm - \gamma H_n) \eta_{n-1} + \gamma_n \varepsilon_n,
\end{equation*}
with $H_n =  K_{x_n} \otimes K_{x_n} + \lambda I$ and $\eta_n = g_n - g_\lambda$.
First we begin by showing that for this recursion and assuming \asm{asm:kernel-bounded}, \asm{asm:data-iid}, we can show \sgdasm{asm:init}, \sgdasm{asm:noise-iid}, \sgdasm{asm:noise-bound},\sgdasm{asm:weird-bound}.

\begin{lemma}[Showing \sgdasm{asm:init}, \sgdasm{asm:noise-iid}, \sgdasm{asm:noise-bound},\sgdasm{asm:weird-bound} for SGD recursion.]
\label{le:noise}
Let us assume \asm{asm:kernel-bounded}, \asm{asm:data-iid},
\BIT
\item \sgdasm{asm:init}  We start at some $g_0 - g_\lambda \in \H$.
\item \sgdasm{asm:noise-iid} $(H_n,\varepsilon_n)$ i.i.d. and $H_n$ is a positive self-adjoint operator so that almost surely $H_n \succcurlyeq \lambda \idm$, with $H  = \E H_n = \Sigma + \lambda \idm$.
\item \sgdasm{asm:noise-bound} We have the two following bounds on the noise:
\begin{eqnarray*}
 \| \varepsilon_n \| &\leqslant& R  ( 1 + 2 \| \Phg - g_\lambda\|_{L_\infty} ) = c^{1/2} \\
\E \varepsilon_n \otimes \varepsilon_n
& \preccurlyeq & 2 \left(1+\|\Phg- g_\lambda\|^2_\infty\right) \Sigma = C\\
\E \|\varepsilon_n\|^2 &\leqslant& 2 \left(1+\|\Phg- g_\lambda\|^2_\infty\right) \tr \Sigma = \tr C.
\end{eqnarray*} 
\item \sgdasm{asm:weird-bound} We have:
\begin{eqnarray*}
\E \Big[ H_k C H^{-1} H_k \Big]
&  \preccurlyeq 
& \left(R^2 + 2\lambda\right) C = \gamma_0^{-1} C  \ .
\end{eqnarray*}
\EIT 
\end{lemma}

\begin{proof}
\sgdasm{asm:init}, \sgdasm{asm:noise-iid} are obviously satisfied.

Let us show \sgdasm{asm:noise-bound}:

\begin{eqnarray*}
\|\varepsilon_n\| &=& \| \xi_n K_{x_n}   +   (\Phg(x_n)- g_\lambda(x_n)) K_{x_n} - \E \left[ (\Phg(x_n)- g_\lambda(x_n)) K_{x_n} \right] \| \\
&\leqslant& (|\xi_n| + |\Phg(x_n)- g_\lambda(x_n)|) \|K_{x_n} \|+  \E \left[ |\Phg(x_n)- g_\lambda(x_n)| \|K_{x_n}\| \right]\\
&\leqslant& (1+\|\Phg- g_\lambda\|_\infty) R +  \|\Phg- g_\lambda\|_\infty R  \\
&=& R(1+2 \|\Phg- g_\lambda\|_\infty) 
\end{eqnarray*}
We have \footnote{We use the following inequality: for all $a$ and $b \in \H$, $(a+b) \otimes (a+b) \preccurlyeq 2 a \otimes a + 2 b \otimes b$. Indeed, for all $x \in \H$, $\langle x, (a+b) \otimes (a+b) x\rangle  =  (\langle a +  b , x\rangle)^2 = (\langle a,x\rangle +  \langle b , x\rangle)^2 \leqslant 2 \langle a,x \rangle^2 + 2 \langle b , x \rangle^2 = 2 \langle x, (a \otimes a) x \rangle + 2 \langle x, (b \otimes b) x \rangle $.}:
\begin{align*}
\varepsilon_n \otimes \varepsilon_n  &\preccurlyeq  2 \xi_n K_{x_n} \otimes \xi_n K_{x_n} +  2\left((\Phg(x_n)- g_\lambda(x_n)) K_{x_n} - \E \left[ (\Phg(x_n)- g_\lambda(x_n)) K_{x_n} \right]\right)   \\ 
& \otimes \left((\Phg(x_n)- g_\lambda(x_n)) K_{x_n} - \E \left[ (\Phg(x_n)- g_\lambda(x_n)) K_{x_n} \right]\right)
\end{align*}
Moreover, $\E [\xi_n K_{x_n} \otimes \xi_n K_{x_n}] = \E [\xi_n^2 K_{x_n} \otimes K_{x_n}] \preccurlyeq \Sigma ,$
And,
\begin{align*}
& \E[((\Phg(x_n)- g_\lambda(x_n) )K_{x_n} - \E \left[ (\Phg(x_n)- g_\lambda(x_n) K_{x_n} \right]) \\
& \hspace{3.66cm}\otimes ((\Phg(x_n)- g_\lambda(x_n)) K_{x_n} - \E \left[ (\Phg(x_n)- g_\lambda(x_n)) K_{x_n} \right])] \\
&= \E \left[ (\Phg(x_n)- g_\lambda(x_n))^2(x_n) K_{x_n} \otimes K_{x_n} \right] - \E \left[ (\Phg(x_n)- g_\lambda(x_n)) K_{x_n} \right] \\
& \hspace{9cm} \otimes \E \left[ (\Phg(x_n)- g_\lambda(x_n)) K_{x_n} \right] \\
&\preccurlyeq \E \left[ (\Phg(x_n)- g_\lambda(x_n))^2(x_n) K_{x_n} \otimes K_{x_n} \right] \\
&\preccurlyeq \|\Phg- g_\lambda\|^2_\infty \Sigma.
\end{align*}
So that, 
$$  
\E \varepsilon_n \otimes \varepsilon_n \preccurlyeq 2 \left(1+\|\Phg- g_\lambda\|^2_\infty\right) \Sigma
$$
Finally $\E \varepsilon_n \otimes \varepsilon_n \preccurlyeq 2 \left(1+\|\Phg- g_\lambda\|^2_\infty\right) \Sigma$, we have $\tr \E \varepsilon_n \otimes \varepsilon_n \leqslant 2 \left(1+\|\Phg- g_\lambda\|^2_\infty\right) \tr \Sigma$, thus $$\tr \E \varepsilon_n \otimes \varepsilon_n = \E \tr \varepsilon_n \otimes \varepsilon_n = \E \|\varepsilon_n\|^2 \leqslant 2 \left(1+\|\Phg- g_\lambda\|^2_\infty\right) \tr \Sigma.$$
To conclude the proof of this lemma, let us show \sgdasm{asm:weird-bound}.
We have: 
\begin{align*}
\E \Big[ ( K_{x_k} \otimes K_{x_k} + \lambda \idm )  \Sigma ( \Sigma + \lambda \idm)^{-1} ( K_{x_k} \otimes K_{x_k} + \lambda \idm ) \Big] &= \E \Big[  K_{x_k} \otimes K_{x_k}  \Sigma ( \Sigma + \lambda \idm)^{-1} K_{x_k} \otimes K_{x_k}  \Big] \\ &+ \lambda  \Sigma \Sigma ( \Sigma + \lambda \idm)^{-1}+\lambda \Sigma 
\end{align*}
Moreover, $\lambda  \Sigma \Sigma ( \Sigma + \lambda \idm)^{-1} = \lambda  \Sigma (\Sigma + \lambda \idm - \lambda \idm) ( \Sigma + \lambda \idm)^{-1} = \lambda \Sigma - \lambda^2 \Sigma( \Sigma + \lambda \idm)^{-1} \preccurlyeq \lambda \Sigma,  $ and similarly, $\E \Big[  K_{x_k} \otimes K_{x_k}  \Sigma ( \Sigma + \lambda \idm)^{-1} K_{x_k} \otimes K_{x_k}  \Big] = \E \Big[  (K_{x_k} \otimes K_{x_k} )^2 \Big] - \lambda \E \Big[  K_{x_k} \otimes K_{x_k} ( \Sigma + \lambda \idm)^{-1} K_{x_k} \otimes K_{x_k}  \Big] \preccurlyeq R^2 \Sigma$.

Finally we obtain $ \E \Big[ ( K_{x_k} \otimes K_{x_k} + \lambda \idm )  \Sigma ( \Sigma + \lambda \idm)^{-1} ( K_{x_k} \otimes K_{x_k} + \lambda \idm ) \Big] \preccurlyeq R^2 \Sigma + \lambda \Sigma + \lambda \Sigma = (R^2 + 2 \lambda) \Sigma.$ 

\end{proof}

\subsection{SGD with decreasing step-size: proof of Theorem~\ref{th:erroralpha}}
\label{ap:EXPalpha}

\begin{proof}[Proof of Theorem~\ref{th:erroralpha} ]

Let us apply Theorem~\ref{th:SGDalpha} to $g_n - g_\lambda$. We assume \asm{asm:kernel-bounded}, \asm{asm:data-iid} and $A = \idm$, such that \asm{asm:kernel-bounded}, \asm{asm:data-iid}, we can show that \sgdasm{asm:init}, \sgdasm{asm:noise-iid}, \sgdasm{asm:noise-bound},\sgdasm{asm:weird-bound}, \sgdasm{asm:commute} are verified (Lemma~\ref{le:noise}). Let $\delta$ correspond to the one of \asm{asm:flambda-correct-sign}.
 We have for $t = \delta/(4R), n \geqslant 1$:   
\begin{align*}
\|g_n - g_\lambda\|_\H &\leqslant \exp\left(  -\frac{\gamma\lambda}{1-\alpha}\left( (n+1)^{1-\alpha} -1  \right)   \right) \|g_0 - g_\lambda\|_\H + \|W_n\|_\H,\ \text{a.s, with}\\
\P \left( \|W_n\|_\H \geqslant \delta/(4R) \right) &\leqslant 2\exp\left( -\frac{\delta^2}{C_R } n^{\alpha}\right), \quad C_R = \displaystyle  \gamma(2 ^{\alpha + 6} R^2 \tr C/\lambda+ 8 R c^{1/2} \delta /3).
\end{align*}
Then if $n$ is such that $\exp\left(  -\frac{\gamma\lambda}{1-\alpha}\left( (n+1)^{1-\alpha} -1  \right)   \right) \leqslant \displaystyle \frac{\delta} {5R\|g_0- g_\lambda\|_\H}$, 
\begin{eqnarray*}
\left\|g_n - g_\lambda \right\|_\H &\leqslant& \frac{\delta}{5R} + \frac{\delta}{4R},\ \text{ with probability } 1-2\exp\left( -\frac{\delta^2}{C_R } n^{\alpha}\right), \\
\left\|g_n - g_\lambda \right\|_\H &<& \frac{\delta}{2R},\ \text{ with probability } 1-2\exp\left( -\frac{\delta^2}{C_R } n^{\alpha}\right).
\end{eqnarray*}

Now assume \asm{asm:separability}, \asm{asm:flambda-correct-sign}, we simply apply Lemma \ref{lm:appr-correct-sign-to-01} to $g_n$ with $q = 2\exp\left( -\frac{\delta^2}{C_R } n^{\alpha}\right)$ And 
\begin{align*}
C_R &= \gamma(2 ^{\alpha + 6} R^2 \tr C/\lambda  + 8 R c^{1/2}\delta /3) \\ 
C_R &= = \gamma\left(\frac{2 ^{\alpha + 7} R^2 \tr \Sigma  \left(1+\|\Phg- g_\lambda\|^2_\infty\right)}{\lambda}+\frac{8R^2 \delta( 1 + 2 \| \Phg - g_\lambda\|_\infty )}{3}\right).
\end{align*}

\end{proof}

\subsection{Tail averaged SGD with constant step-size: proof of Theorem~\ref{th:errortail} }
\label{ap:EXPaverage}

\begin{proof}[Proof of Theorem~\ref{th:errortail} ]

Let us apply Corollary~\ref{co:SGDtailaveraged} to $g_n - g_\lambda$. We assume \asm{asm:kernel-bounded}, \asm{asm:data-iid} and $A = \idm$, such that \sgdasm{asm:init}, \sgdasm{asm:noise-iid}, \sgdasm{asm:noise-bound},\sgdasm{asm:weird-bound}, \sgdasm{asm:commute} are verified (Lemma~\ref{le:noise}). Let $\delta$ correspond to the one of \asm{asm:flambda-correct-sign}.
 We have for $t = \delta/(4R), n \geqslant 1$:   
\begin{eqnarray*}
\left\|\bar{g}_n^{\textrm {tail}} - g_\lambda \right\|_\H &\leqslant& (1-\gamma\lambda)^{n/2}  \|g_0- g_\lambda\|_\H + L_n \quad,\text{with} \\
 \P(L_n \geqslant t ) &\leqslant& 4\exp\left( - (n+1)t^2/(4E_t)\right).
\end{eqnarray*}
Then as soon as $(1-\gamma\lambda)^{n/2} \leqslant \displaystyle \frac{\delta} {5R\|g_0- g_\lambda\|_\H}$, 
\begin{eqnarray*}
\left\|\bar{g}_n^{\textrm {tail}} - g_\lambda \right\|_\H &\leqslant& \frac{\delta}{5R} + \frac{\delta}{4R},\ \text{ with probability } 1-4\exp\left( - (n+1)\delta^2/(64 R^2E_{\delta/(4R)})\right), \\
\left\|\bar{g}_n^{\textrm {tail}} - g_\lambda \right\|_\H &<& \frac{\delta}{2R},\ \text{ with probability } 1-4\exp\left( - (n+1)\delta^2/(64 R^2E_{\delta/(4R)})\right).
\end{eqnarray*}

Now assume \asm{asm:separability}, \asm{asm:flambda-correct-sign}, we simply apply Lemma \ref{lm:appr-correct-sign-to-01} to $\bar{g}_n^{\textrm {tail}}$ with $q = 4\exp\left( - (n+1)\delta^2/K_R)\right)$. And 
\begin{align*}
K_R = 64 R^2E_{\delta/(4R)} &=  64 R^2 \left(4\tr(H^{-2}C)+\frac{2c^{1/2}}{3\lambda}\cdot \frac{\delta}{4R}\right)  \\
 &= 512 R ^2  \left(1+\|\Phg- g_\lambda\|^2_\infty\right)\tr((\Sigma + \lambda \idm)^{-2}\Sigma)+\frac{32 \delta R^2 ( 1 + 2 \| \Phg - g_\lambda\|_\infty )}{3\lambda} .
\end{align*}

\end{proof}

\section{Extension of Corollary \ref{co:SGDtailaveraged} and Theorem \ref{th:errortail} for the full averaged case.}
\label{ap:average}

\subsection{Extension of Corollary \ref{co:SGDtailaveraged} for the full averaged case.}
\label{ap:SGDfullaverage}

Let us recall the SGD abstract recursion defined in Eq. \eqref{eq:SGDabstract} that we are going to further apply with $\eta_n = g_n - g_\lambda$, $H_n = K_{x_n} \otimes K_{x_n} + \lambda \idm$ and $H = \Sigma + \lambda \idm$: 
\begin{align*}
   \eta_n &= ( \idm - \gamma H_n) \eta_{n-1} + \gamma_n \varepsilon_n, \\
 \eta_n &= 
\underbrace{M(n,1)  \eta_0}_{\eta_n^{\textrm {bias}}} + \underbrace{\sum_{k=1}^n \gamma_k M(n,k+1) \varepsilon_k}_{\eta_n^{\textrm {variance}}}.
\end{align*}
\paragraph{Notations.} The second term, $\eta_n^{\textrm {variance}}$, is treated by Theorem \ref{th:SGDaveraged} of the article. Now consider that $\eta_0 \neq 0$ and let us bound the initial condition term i.e., $\eta_n^{\textrm {bias}} = M(n,1)  \eta_0$. Let us define also an auxiliary sequence $(u_n)$ that follows the same recursion as $\eta_n^{\textrm {bias}}$ but with $H$:
\begin{align*}
\eta_n^{\textrm {bias}} & = (\idm - \gamma H_n)\eta_{n-1}^{\textrm {bias}} \\
u_n & = (\idm - \gamma H)u_{n-1}, \qquad u_0 = \eta_{0}^{\textrm {bias}} = \eta_0.
\end{align*}
We define $w_n = \eta_n^{\textrm {bias}} - u_n$ and as always we consider the first $n$ average of each of these sequences that we are going to denote $\bar{w}_n$, $\bar{\eta}^{\textrm {bias}}_n$ and $\bar{u}_n$ respectively. 

Note $\tilde{\varepsilon}_n = (H-H_n) \eta_{n-1}^{\textrm {bias}}$ and $\tilde{H}_n = H$, then $w_n$ follows the recursion : $w_0 = 0$, and
\begin{align} 
\label{eq:auxiliarysequence}
w_n = (\idm - \gamma \tilde{H}_n)w_{n-1} + \gamma \tilde{\varepsilon}_n .
\end{align}

Thus, $w_n$ follows the same recursion as Eq.\eqref{eq:SGDabstract} with $( \tilde{H}_n, \tilde{\varepsilon}_n)$. We thus have the following corollary:

\begin{corollary}
Assume that the sequence $(w_n)$ defined in Eq. \eqref{eq:auxiliarysequence} verifies \sgdasm{asm:init}, \sgdasm{asm:noise-iid}, \sgdasm{asm:noise-bound}, \sgdasm{asm:weird-bound} and \sgdasm{asm:commute} with $(\tilde{H}_n,\tilde{\varepsilon}_n)$, then for $t > 0, n \geqslant 1$:    
\begin{equation*}
\displaystyle\P \left( \left\| A^{1/2} \bar{w}_n  \right\|_\H \geqslant t \right) \leqslant 2 \exp\left[-\frac{(n+1) t^2}{\tilde{E}_t}\right],
\end{equation*}
where $\tilde{E}_t$ is defined with respect to the constants introduced in the assumptions (with a tilde):  
\begin{equation*}
 \tilde{E}_t =   4\tr(AH^{-2}\tilde{C})+\frac{2\tilde{c}^{1/2} \|A^{1/2}\|_{\textrm {op}}}{3\lambda}\cdot t  .
 \end{equation*}
\end{corollary}

\begin{proof}
Apply Theorem \ref{th:SGDaveraged} to the sequence $(w_n)$ defined in Eq. \eqref{eq:auxiliarysequence}.
\end{proof}

Now, we can decompose $\eta_n$ in three terms: $\eta_n = \eta_n^{\textrm {bias}} + \eta_n^{\textrm {variance}} = w_n + u_n +  \eta_n^{\textrm {variance}}$. We can thus state the following general result:

\begin{theorem}
\label{th:withbias}
Assume \sgdasm{asm:init}, \sgdasm{asm:noise-iid}, \sgdasm{asm:noise-bound}, \sgdasm{asm:weird-bound}, \sgdasm{asm:commute} for both $(H_n,\varepsilon_n)$ and $(\tilde{H}_n,\tilde{\varepsilon}_n)$, and consider the average of the sequence defined in Eq. \eqref{eq:SGDabstract}. We have for $t > 0, n \geqslant 1$:  
\begin{eqnarray}
\left\|A^{1/2}\bar{\eta}_n \right\|_\H &\leqslant& \frac{ \left\| A^{1/2} \right\|\left\| \eta_0\right\|_\H}{(n+1)\gamma\lambda} + L_n \quad,\text{with} \\
 \P(L_n \geqslant t ) &\leqslant&  4\exp\left(-\frac{(n+1) t^2}{\max(E_t,\tilde{E}_t)} \right).
\end{eqnarray}

\end{theorem}

\begin{proof}[Proof of Theorem \ref{th:withbias}]
As $\bar{\eta}_n = \bar{\eta}_n^{\textrm {bias}} + \bar{\eta}_n^{\textrm {variance}} = \bar{w}_n + \bar{u}_n +  \bar{\eta}_n^{\textrm {variance}}$, we are going to bound $\bar{u}_n$, then the sum $\bar{w}_n + \bar{\eta}_n^{\textrm {variance}}$.

First, $\displaystyle \|\bar{u}_n\| = \left\| \frac{1}{n+1} \sum_{k=0}^n u_k\right\|\leqslant \frac{1}{n+1} \sum_{k=0}^n \left\| u_k\right\|\leqslant \frac{1}{n+1} \sum_{k=0}^n (1-\gamma\lambda)^k\left\| \eta_0\right\|\leqslant \frac{\left\| \eta_0\right\|}{(n+1)\gamma\lambda} .$

Thus, we have:
\begin{eqnarray*}
\left\| A^{1/2}\bar{\eta}_n \right\| &\leqslant & \frac{ \left\| A^{1/2} \right\|\left\| \eta_0\right\|}{(n+1)\gamma\lambda} + \left\|A^{1/2}\bar{w}_n\right\| + \left\|A^{1/2}\bar{\eta}_n^{\textrm {variance}}\right\|,
\end{eqnarray*} 
Let $L_n = \left\|A^{1/2}\bar{w}_n\right\| + \left\|A^{1/2}\bar{\eta}_n^{\textrm {variance}}\right\|$, for $t \geqslant 0$,
\begin{eqnarray*}
\P(L_n \geqslant t ) &=& \P(\left\|A^{1/2}\bar{w}_n\right\| + \left\|A^{1/2}\bar{\eta}_n^{\textrm {variance}}\right\| \geqslant t ) \\
&\leqslant& \P\left(\left\|A^{1/2}\bar{w}_n\right\| \geqslant t\right)+\P\left( \left\|A^{1/2}\bar{\eta}_n^{\textrm {variance}}\right\| \geqslant t \right) \\
&\leqslant& 2\left[\exp\left[-\frac{(n+1) t^2}{\tilde{E}_t}\right] + \exp\left[-\frac{(n+1) t^2}{E_t}\right]\right] .
\end{eqnarray*}
Hence,
\begin{eqnarray*}
\P(L_n \geqslant t ) &\leqslant& 4 \exp\left(-\frac{(n+1) t^2}{\max(E_t,\tilde{E}_t)} \right). 
\end{eqnarray*}

\end{proof}

\subsection{Extension of Theorem \ref{th:errortail} for the full averaged case.}
\label{ap:EXPfullaverage}

Same situation here, we want to apply full averaged SGD instead of the tail-averaged technique. 

\begin{theorem}
\label{th:expwithbias}
Assume \asm{asm:separability}, \asm{asm:kernel-bounded}, \asm{asm:data-iid}, \asm{asm:flambda-correct-sign} and $\gamma_n = \gamma$ for any n, $\gamma\lambda < 1$ and $\gamma \leqslant \gamma_0 = (R^2 + \lambda)^{-1}$. Let $\bar{g}_n$ be the average of the first n iterate of the SGD recursion defined in Eq.~(\ref{eq:SGDrecursion}), as soon as: $\displaystyle n \geqslant \frac{5R\|g_0- g_\lambda\|_\H}{\lambda\gamma \delta} $, then 

$$ \mathcal{R}(\bar{g}_n^{\textrm {tail}}) = \mathcal{R}^*, \mbox{ with probability at least }1-4\exp\left( - \delta^2 K_R (n+1)\right),$$
and in particular
$$ \E{\mathcal{R}(\bar{g}_n^{\textrm {tail}}) - \mathcal{R}^*} \leqslant 4\exp\left( - \delta^2 K_R (n+1)\right),$$ with \begin{eqnarray*}
K_R^{-1}  = \max \left.
  \begin{cases}
    \displaystyle 128 R^2 \left(1+\|\Phg- g_\lambda\|^2_\infty\right)\tr((\Sigma + \lambda \idm)^{-2}\Sigma)+\frac{8 R^2 ( 1 + 2 \| \Phg - g_\lambda\|_\infty )}{3\lambda}\\
    \displaystyle 64 R^4 \|g_0- g_\lambda\|_\H \tr((\Sigma + \lambda \idm)^{-2}\Sigma)+\frac{16 R^4  \| g_0 - g_\lambda\|_\H}{3\lambda}.
  \end{cases}
  \right\}
\end{eqnarray*}
\end{theorem}

\begin{proof}[Proof of Theorem \ref{th:expwithbias}]

We want to apply Theorem \ref{th:withbias} to the SGD recursion. We thus want to check that assumptions \sgdasm{asm:init}, \sgdasm{asm:noise-iid}, \sgdasm{asm:noise-bound}, \sgdasm{asm:weird-bound}, \sgdasm{asm:commute} are verified for both $(H_n,\varepsilon_n)$ and $(\tilde{H}_n,\tilde{\varepsilon}_n)$. For the recursion involving $(H_n,\varepsilon_n)$, this corresponds to Lemma \ref{le:noise}. For the recursion involving $(\tilde{H}_n = H,\tilde{\varepsilon}_n = (H-H_n)M(n-1,1)(g_0-g_\lambda)$, this corresponds to the following lemma:

\begin{lemma}[Showing \sgdasm{asm:init}, \sgdasm{asm:noise-iid}, \sgdasm{asm:noise-bound}, \sgdasm{asm:weird-bound} for the auxiliary recursion.]
\label{le:noiseauxiliary}
Let us assume \asm{asm:kernel-bounded}, \asm{asm:data-iid},
\BIT
\item \sgdasm{asm:init}  We start at some $g_0 - g_\lambda \in \H$.
\item \sgdasm{asm:noise-iid}   $(\tilde{H}_n,\tilde{\varepsilon}_n)$ i.i.d. and $\tilde{H}_n$ is a positive self-adjoint operator so that almost surely $\tilde{H}_n \succcurlyeq \lambda \idm$, with $H  = \E \tilde{H}_n = \Sigma + \lambda \idm$.
\item \sgdasm{asm:noise-bound} We have the two following bounds on the noise:
\begin{eqnarray*}
 \| \tilde{\varepsilon}_n \| &\leqslant& 2 R^2 \|g_0-g_\lambda\|_\H = \tilde{c}^{1/2} \\
\E \tilde{\varepsilon}_n \otimes \tilde{\varepsilon}_n
& \preccurlyeq & R^2 \|g_0 - g_\lambda\|_\H \Sigma = \tilde{C}\\
\E \|\tilde{\varepsilon}_n\|^2 &\leqslant& R^2 \|g_0 - g_\lambda\|_\H \tr \Sigma = \tr \tilde{C}.
\end{eqnarray*} 
\item \sgdasm{asm:weird-bound} We have:
\begin{eqnarray*}
\E \Big[ \tilde{H}_k \tilde{C} H^{-1} \tilde{H}_k \Big] 
&  \preccurlyeq 
& \left(R^2 + \lambda\right) \tilde{C} = \tilde{\gamma_0}^{-1} \tilde{C}  \ .
\end{eqnarray*}
\EIT 
\end{lemma}

\begin{proof}

\sgdasm{asm:init}, \sgdasm{asm:noise-iid} are obviously satisfied.

Let us show \sgdasm{asm:noise-bound}:
For the first one:
\begin{align*}
\|\tilde{\varepsilon}_n\| &= \left\|(H-H_n)M(n-1,1)(g_0-g_\lambda)\right\| \\
&\leqslant \left\|(\Sigma-K_{x_n}\otimes K_{x_n})\right\|\left\|M(n-1,1)\right\|\left\|g_0-g_\lambda\right\| \\
&\leqslant 2 R^2 \|g_0-g_\lambda\|_\H.
\end{align*}
\begin{align*}
\|\tilde{\varepsilon}_n\| &= \left\|(H-H_n)M(n-1,1)(g_0-g_\lambda)\right\| \\
&\leqslant \left\|(\Sigma-K_{x_n}\otimes K_{x_n})\right\|\left\|M(n-1,1)\right\|\left\|g_0-g_\lambda\right\| \\
&\leqslant 2 R^2 \|g_0-g_\lambda\|_\H.
\end{align*}
And for the second inquality:
\begin{align*}
\E \left[\tilde{\varepsilon}_n \otimes \tilde{\varepsilon}_n | \F_{n-1}\right]&= \E \left[\left(\Sigma - K_{x_n}\otimes K_{x_n}\right)\eta_n^{\textrm {bias}}\otimes \eta_n^{\textrm {bias}} \left(\Sigma - K_{x_n}\otimes K_{x_n}\right)| \F_{n-1}\right] \\
&= \Sigma \eta_n^{\textrm {bias}} \otimes \eta_n^{\textrm {bias}} \Sigma - 2 \Sigma \eta_n^{\textrm {bias}}\otimes \eta_n^{\textrm {bias}} \Sigma + \E\left[ K_{x_n}\otimes K_{x_n} \eta_n^{\textrm {bias}}\otimes \eta_n^{\textrm {bias}}  K_{x_n}\otimes K_{x_n}    \right] \\
&= - \Sigma \eta_n^{\textrm {bias}}\otimes \eta_n^{\textrm {bias}} \Sigma +   \E\left[ \langle K_{x_n} , \eta_n^{\textrm {bias}} \rangle^2 K_{x_n}\otimes K_{x_n}   \right] \\
&\preccurlyeq R^2 \|g_0 - g_\lambda\|_\H \Sigma.
\end{align*}

Finally, we have for \sgdasm{asm:weird-bound} :
\begin{eqnarray*}
\E \Big[ \tilde{H}_k \tilde{C} H^{-1} \tilde{H}_k \Big] = H \tilde{C} = R^2 \|g_0 - g_\lambda\|_\H (\Sigma^2 + \lambda \Sigma)  &\preccurlyeq& R^2 \|g_0 - g_\lambda\|_\H (\|\Sigma\|_{\textrm {op}} + \lambda ) \Sigma  \\
 &\preccurlyeq&
\left(R^2 + \lambda\right) \tilde{C} = \tilde{\gamma_0}^{-1} \tilde{C} .
\end{eqnarray*}

\end{proof}

Let us apply now Theorem~\ref{th:withbias} to $g_n - g_\lambda$. We assume \asm{asm:kernel-bounded}, \asm{asm:data-iid} and $A = \idm$, such that \sgdasm{asm:init}, \sgdasm{asm:noise-iid}, \sgdasm{asm:noise-bound}, \sgdasm{asm:weird-bound}, \sgdasm{asm:commute} are verified for both problems ($(H_n,\varepsilon_n)$ and $(\tilde{H}_n,\tilde{\varepsilon}_n)$) (Lemma~\ref{le:noise},\ref{le:noiseauxiliary}). Let $\delta$ correspond to the one of Assumption \ref{asm:flambda-correct-sign}.
 We have for $t = \delta/(4R), n \geqslant 1$:   
\begin{eqnarray*}
\left\|\bar{g}_n - g_\lambda \right\|_\H &\leqslant& \frac{ \left\| g_0 - g_\lambda\right\|_\H}{(n+1)\gamma\lambda} + L_n \quad,\text{with} \\
 \P(L_n \geqslant t ) &\leqslant&  4\exp\left(-\frac{(n+1) t^2}{\max(E_t,\tilde{E}_t)} \right).
\end{eqnarray*}
Then as soon as $\displaystyle\frac{1}{(n+1)\lambda\gamma} \leqslant  \frac{\delta} {5R\|g_0- g_\lambda\|_\H}$, 
\begin{eqnarray*}
\left\|\bar{g}_n - g_\lambda \right\|_\H &\leqslant& \frac{\delta}{5R} + \frac{\delta}{4R},\ \text{ with probability } 1-4\exp\left(-\frac{(n+1) \delta^2}{16 R^2\max(E_{\delta/4R},\tilde{E}_{\delta/4R})} \right), \\
\left\|\bar{g}_n - g_\lambda \right\|_\H &<& \frac{\delta}{2R},\ \text{ with probability } 1-4\exp\left(-\frac{(n+1) \delta^2}{16 R^2\max(E_{\delta/4R},\tilde{E}_{\delta/4R})} \right).
\end{eqnarray*}

Now assume \asm{asm:separability}, \asm{asm:flambda-correct-sign}, we now only have to apply Lemma \ref{lm:appr-correct-sign-to-01} to the estimator $\bar{g}_n$ with the probability $\displaystyle q = 4\exp\left(-\frac{(n+1) \delta^2}{16 R^2\max(E_{\delta/4R},\tilde{E}_{\delta/4R})} \right)$. And,
\begin{align*}
K_R^{-1} &= 16 R^2\max(E_{\delta/4R},\tilde{E}_{\delta/4R}) \\
 &= \max
  \begin{cases}
    \displaystyle 128 R^2 \left(1+\|\Phg- g_\lambda\|^2_\infty\right)\tr((\Sigma + \lambda \idm)^{-2}\Sigma)+\frac{8 R^2 ( 1 + 2 \| \Phg - g_\lambda\|_\infty )}{3\lambda}\\
    \displaystyle 64 R^4 \|g_0- g_\lambda\|_\H \tr((\Sigma + \lambda \idm)^{-2}\Sigma)+\frac{16 R^4  \| g_0 - g_\lambda\|_\H}{3\lambda}.
  \end{cases}
\end{align*}

\end{proof}

\section{Convergence rate under weaker margin assumption}
\label{ap:weakmargin}

We make the following assumptions: 

\bas \label{asm:wseparability} $\forall \delta > 0, \ \P \left( | g_* | \leqslant 2\delta \right) \leqslant \delta^\alpha.$ \eas

\bas \label{asm:sflambda-correct-sign} There exists \footnote{This assumption is verified for the following source condition $\exists g \in \H, r>0$ s.t. $\P_\H(g) = \Sigma^r g_*$. 
If the additionnal assumption \asm{asm:eigenvalues} is verified then \asm{asm:sflambda-correct-sign} is verified with $\gamma = \frac{r-1/2}{2r+1/\beta}$ \citep{caponnetto2007optimal}.} 
$\gamma > 0$ such that $\forall \lambda > 0, $ $\|g_* - g_\lambda\|_\infty \leqslant \lambda^\gamma$.  \eas

\bas \label{asm:eigenvalues} The eigenvalues of $\Sigma$ decrease as $1/n^\beta$ for $\beta > 1$. \eas
Note that \asm{asm:wseparability} is weaker than \asm{asm:separability} and to balance this we need a stronger condition on $g_\lambda$ than \asm{asm:flambda-correct-sign} which is \asm{asm:sflambda-correct-sign}. \asm{asm:eigenvalues} is just a technical assumption needed to give explicit rate. The following Corollary corresponds to Theorem \ref{th:errortail} with the new assumptions. Note that it could also be shown for the full average sequence $\bar{g}_n$.
 
\begin{corollary}[Explicit onvergence rate under weaker margin condition]
\label{co:weakmargin}
Assume \asm{asm:kernel-bounded}, \asm{asm:data-iid}, \- \asm{asm:wseparability}, \asm{asm:sflambda-correct-sign} and \asm{asm:eigenvalues}. Let $\gamma_n = \gamma$ for any n, $\gamma\lambda < 1$ and $\gamma \leqslant \gamma_0 = (R^2 + 2\lambda)^{-1}$. Let $\bar{g}_n^{\textrm {tail}}$ be the n-th iterate of the recursion defined in Eq.~(\ref{eq:SGDrecursion}), and $\bar{g}_n^{\textrm {tail}} = \frac{1}{\lfloor n/2 \rfloor} \sum_{i=\lfloor n/2 \rfloor}^{n} g_i$, as soon as $\displaystyle n \geqslant \frac{2}{\gamma\lambda}\ln (\frac{5R \|g_0-g_\lambda\|_\H}{\delta})$, then 
$$ \E \left[ R(\bar{g}_n^{\textrm {tail}}) - R^* \right] \leqslant \frac{C_{\alpha,\beta}}{n^{\alpha\cdot q_{\gamma,\beta} }}.$$ 
\end{corollary}

\begin{proof}
The proof technique follows the one of \citet{audibert2007fast}. 

Let $\delta,\lambda > 0$, such that $\|g_* - g_\lambda\|_\infty \leqslant \delta$. Remark that $\forall j \in \mathbb{N}$, 

$\P \left({\textrm {sign} (g_*(X)) g_\lambda(X) \leqslant 2^{j} \delta } \right) \leqslant \P \left({ |g_\lambda(X)| \leqslant 2^{j} \delta } \right) \leqslant \P \left({ |g_*(X)| \leqslant 2^{j+1} \delta } \right) \leqslant 2^{\alpha j} \delta^{\alpha}.$

Note $A_0 = \{ x \in \X |\ {\textrm {sign} (g_*) g_\lambda \leqslant \delta }  \}$ and for $j \geqslant 1 $, $A_j = \{ x \in \X |\ 2^{j-1} \delta < {\textrm {sign} (g_*) g_\lambda \leqslant 2^{j} \delta }  \} $. Then,

\begin{align*}
\E \left[ R(\bar{g}_n^{\textrm {tail}}) - R^* \right] &= \sum_{j \in \mathbb{N} } \E \left[ \left(R(\bar{g}_n^{\textrm {tail}}) - R^*\right) \mathbf{1}_{A_j} \right] \\
&= \E \left[ \left(R(\bar{g}_n^{\textrm {tail}}) - R^*\right) \mathbf{1}_{{\textrm {sign} (g_*) g_\lambda \leqslant \delta }} \right] + \sum_{j \geqslant 1 }   \E \left[ \left(R(\bar{g}_n^{\textrm {tail}}) - R^*\right) \mathbf{1}_{A_j} \right] \\
&\leqslant \P \left({\textrm {sign} (g_*(X)) g_\lambda(X) \leqslant \delta } \right) + \sum_{j \geqslant 1 }   \E \left[ \left(R(\bar{g}_n^{\textrm {tail}}) - R^*\right) \mathbf{1}_{2^{j-1} \delta < {\textrm {sign} (g_*(X)) g_\lambda(X) \leqslant 2^{j} \delta }} \right] \\
&\leqslant \delta^\alpha + \sum_{j \geqslant 1 }   \E_X \left[\E_{x_1,\dots,x_n} \left[ \underbrace{\left(R(\bar{g}_n^{\textrm {tail}}) - R^*\right) \mathbf{1}_{2^{j-1} \delta < {\textrm {sign} (g_*(X)) g_\lambda(X)}} }_{\textrm {Theorem\ \ref{th:errortail}}} | x_1,\dots,x_n \right] \right. \\
& \hspace{9.23cm} \left. \cdot \mathbf{1}_{ {\textrm {sign} (g_*(X)) g_\lambda(X) \leqslant 2^{j} \delta }} \right]  \\
&\leqslant \delta^\alpha + 4 \sum_{j \geqslant 1 }   \P \left({\textrm {sign} (g_*(X)) g_\lambda(X) \leqslant 2^{j} \delta } \right) \exp\left( - (2^j\delta)^2 K_R(\delta) (n+1)\right)  \\
&\leqslant \delta^\alpha + 4 \delta^\alpha \sum_{j \geqslant 1 }  2^{\alpha j } \exp\left( - (2^j\delta)^2 K_R(\delta) (n+1)\right),
\end{align*}
and $\displaystyle K_R(\delta)^{-1} = 2^9 R ^2  \left(1+\|\Phg- g_\lambda\|^2_\infty\right)\tr(\Sigma(\Sigma + \lambda \idm)^{-2})+\frac{32 \delta R^2 ( 1 + 2 \| \Phg - g_\lambda\|_\infty )}{3\lambda} .$ Let us now choose $\delta $ as a function of $n$ to cancel the dependence on $n$ in the exponential term. In the following, as we assumed \asm{asm:sflambda-correct-sign}, we chose $\lambda = \delta^{1/\gamma}$  such that $\|g_* - g_\lambda\|_\infty \leqslant \lambda^\gamma = \delta$. Second, \asm{asm:eigenvalues} implies \citep[see][]{caponnetto2007optimal} that $\displaystyle \tr(\Sigma(\Sigma + \lambda \idm)^{-2}) \leqslant \frac{\beta}{(\beta-1) \lambda^{1+1/\beta}}.$ For $\delta$ small enough, we have:
\begin{align*}
K_R(\delta)^{-1} &\leqslant  2^{10}  \frac{\beta R ^2}{(\beta-1) \delta^{\frac{1+1/\beta}{\gamma}}} +32 \delta^{(\gamma-1)/\gamma} R^2 \\
K_R(\delta)^{-1} &\leqslant  2^{11}  \frac{\beta R ^2}{(\beta-1)}\cdot \delta^{-(\beta+1)/\beta\gamma}
\end{align*}
Hence, if we take $\delta^2 \delta^{(\beta+1)/\beta\gamma} = 1/n$, i.e., $\delta = n^{-\gamma/(2\gamma + 1+ 1/\beta) }$, we have:
\begin{align*}
\E \left[ R(\bar{g}_n^{\textrm {tail}}) - R^* \right] &\leqslant \frac{1+ \sum_{j \geqslant 1 }  2^{\alpha j + 2} \exp\left( - 4^{j} (\beta-1)/(2^{11}\beta R^2)\right)}{n^{\alpha\gamma/(2\gamma + 1+ 1/\beta) }}.
\end{align*}
As the sum converges, we have proved the result.

\end{proof}